\documentclass{article} % For LaTeX2e
\usepackage{iclr2020_conference,times}

\usepackage{hyperref}
\usepackage{url}

\usepackage[utf8]{inputenc} % allow utf-8 input
\usepackage[T1]{fontenc}    % use 8-bit T1 fonts
\usepackage{hyperref}       % hyperlinks
\usepackage{url}            % simple URL typesetting
\usepackage{booktabs}       % professional-quality tables
\usepackage{amsfonts}       % blackboard math symbols
\usepackage{nicefrac}       % compact symbols for 1/2, etc.
\usepackage{microtype}      % microtypography

\usepackage{amsmath,amssymb,amsthm,mathrsfs,amsfonts,dsfont}
\usepackage{mathtools}
\usepackage{algorithm,algorithmic}
\usepackage{amssymb, multirow, paralist, color}
\newtheorem{thm}{Theorem}
\newtheorem{prop}{Proposition}
\newtheorem{lem}{Lemma}

\newtheorem{ass}{Assumption}
\usepackage{enumitem}
\usepackage{booktabs}
\usepackage{amsmath}
\usepackage{bbm}
\usepackage{url}
\iclrfinalcopy

\def \X {\mathcal{X}}

\def \R {\mathbb{R}}

\def \w {\mathbf{w}}
\def \v {\mathbf{v}}

\def \x {\mathbf{x}}

\def \E {\mathbb{E}}

\def \x {\mathbf{x}}

\def \a {\mathbf{a}}

\def \b {\mathbf{b}}

\def \z {\mathbf{z}}
\def \s {\mathbf{s}}

\def \u {\mathbf{u}}

\def \g {\mathbf{g}}

\def \P {\mathcal{P}}

\def \E {\mathbb{E}}

\def \x {\mathbf{x}}
\def \g {\mathbf{g}}

\def \z {\mathbf{z}}
\def \u {\mathbf{u}}

\def \w {\mathbf{w}}

\def \R {\mathbb{R}}

\def \v {\mathbf{v}}

\def \ph {\widehat{p}}

\def \a {\mathbf{a}}
\def \b {\mathbf{b}}

\def \s {\mathbf{s}}

\def \X {\mathcal{X}}
\def \Y {\mathcal{Y}}

\def \P {\mathbb{P}}

\title{Stochastic AUC Maximization with\\
	Deep Neural Networks}

\author{Mingrui Liu\\
Department of Computer Science\\
The University of Iowa\\
Iowa City, IA, 52242, USA \\
\texttt{mingrui-liu@uiowa.edu}
\And
Zhuoning Yuan\\
Department of Computer Science\\
The University of Iowa\\
Iowa City, IA, 52242, USA \\
\texttt{zhuoning-yuan@uiowa.edu}
\And
Yiming Ying\\
Department of Mathematics and Statistics\\
SUNY at Albany\\
Albany, NY, 12222, USA \\
\texttt{yying@albany.edu}
\And
Tianbao Yang\\
Department of Computer Science\\
The University of Iowa\\
Iowa City, IA, 52242, USA\\
\texttt{tianbao-yang@uiowa.edu}
}

\begin{document}

\maketitle

\begin{abstract}
	%\vspace*{-0.3in}
	
	Stochastic AUC maximization has garnered an increasing interest due to better fit to imbalanced data classification. However, existing  works are limited to stochastic AUC maximization with a linear predictive model, which restricts its predictive power when dealing with extremely complex data. In this paper, we consider stochastic AUC maximization problem with a deep neural network as the predictive model. Building on the saddle point reformulation of a surrogated loss of AUC, the problem can be cast into a {\it non-convex concave} min-max problem. 
	The main contribution made in this paper is to make stochastic AUC maximization more practical for deep neural networks and big data with theoretical insights as well. In particular, we propose to explore Polyak-\L{}ojasiewicz (PL) condition that has been proved and observed in deep learning, which enables us to develop new stochastic algorithms with even faster convergence rate  and more practical step size scheme. An AdaGrad-style algorithm  is also analyzed under the PL condition with adaptive convergence rate. 
	%we design two algorithms Proximal Primal-Dual SG (PPD-SG) and Proximal Primal-Dual Adagrad (PPD-Adagrad), and establish corresponding complexity guarantees, where PPD-Adagrad enjoys adaptive complexity, i.e., it has lower complexity than PPD-SG when the growth of cumulative stochastic gradient is slow. 
	Our experimental results demonstrate the effectiveness of the proposed algorithms.
	%\vspace*{-0.53in}
\end{abstract}
%\vspace*{-0.2in}

\setlength{\abovedisplayskip}{3pt}
\setlength{\belowdisplayskip}{0pt}
\section{Introduction}
%\vspace*{-0.3in}

Deep learning has been witnessed with tremendous success for various tasks, including computer vision~\citep{krizhevsky2012imagenet,simonyan2014very,he2016deep,ren2015faster}, speech recognition~\citep{hinton2012deep,mohamed2012acoustic,graves2013generating}, natural language processing~\citep{bahdanau2014neural, sutskever2014sequence,devlin2018bert}, etc. From an optimization perspective, all of them are solving an empirical risk minimization problem in which the objective function is a surrogate loss of  the prediction error made by a deep neural network in comparison with the ground-truth label. 
For example, for image classification task, the objective function is often chosen as the cross entropy between the probability distribution calculated by forward propagation of a convolutional neural network and the vector encoding true label information~\citep{krizhevsky2012imagenet, simonyan2014very, he2016deep}, where the cross entropy is a surrogate loss of the misclassification rate. However, when the data is imbalanced, this formulation is not reasonable since the data coming from minor class have little effect in this case and the model is almost determined by the data from the majority class. 
%and 

To address this issue, AUC maximization has been proposed as a new learning paradigm~\citep{zhao2011online}.  Statistically, AUC (short for  Area Under the ROC curve)~is defined as the probability that the prediction score of a positive example is higher than that of a negative example~\citep{hanley1982meaning,hanley1983method} . Compared with misclassification rate and its corresponding surrogate loss, AUC is more suitable for imbalanced data setting~\citep{elkan2001foundations}. Several online or stochastic algorithms for AUC maximization have been developed based on a convex surrogate loss~\citep{zhao2011online,gao2013one,ying2016stochastic,liu2018fast,natole2018stochastic}. However, all of these works only consider learning a linear predictive model. This naturally motivates the following question:

\textbf{How to design stochastic algorithms with provable guarantees to solve the AUC maximization problem with a deep neural network as the predictive model?} 

In this paper, we make some efforts to answer this question. We design two algorithms with state-of-the-art complexities for this problem. Based on a surrogated loss of AUC and inspired by the min-max reformulation in~\citep{ying2016stochastic}, we cast the problem into a non-convex concave min-max stochastic optimization problem, where it is nonconvex in the primal variable and concave in the dual variable. This allows us to leverage the inexact proximal point algorithmic framework proposed in~\citep{rafique2018non} to solve stochastic AUC maximization with a deep neural network.  %To solve this problem, we first introduce an algorithm in~\cite{rafique2018non} which can be directly adapted to our setting. When the function is weakly convex in terms of the primal variable, the algorithm is guaranteed to converge to first-order stationary point. 
However, their algorithms  are limited for stochastic AUC maximization with a deep neural network due to three reasons. First, their algorithms are general and do not utilize the underlying favorable  property of the the objective function induced by an overparameterized deep neural network, which prevents them from designing better algorithms with faster convergence. Second, 
these algorithms use a polynomially decaying step size scheme instead of the widely used geometrically decaying step size scheme in deep neural network training. Third, the algorithm in~\citep{rafique2018non} with the best attainable complexity only applies to the finite-sum setting, which needs to go through all data at the end of each stage and are not applicable to the pure stochastic setting. 
%their algorithm with best attainable complexity is restricted to the finite sum setting, which is not applicable to the general stochastic AUC optimization. Second, they do not leverage the property of the neural network (e.g., PL condition) which usually holds in practice, which prevents them from getting better algorithms with faster convergence. 

%However, this algorithm is general and does not utilize the potential property of the the objective function. In addition, the algorithms in~\cite{rafique2018non} with best attainable complexity only applies for the finite-sum setting, which needs to go through all data at the end of each stage and is not applicable to the pure stochastic setting. At last, these algorithms use polynomially decaying stepsize scheme instead of the widely used geometrically decaying learning rate scheme in deep neural network training, which is not practical. 

To address these limitations, we propose to leverage the Polyak-\L{}ojasiewicz (PL) condition of the objective function for AUC maximization with a deep neural network. The PL condition (or its equivalent condition)  has been proved for a class of linear and non-linear neural networks~\citep{hardt2016identity,charles2017stability,zhou2017characterization}. It is the key to recent developments that prove that (stochastic) gradient descent can find a global minimum for an  overparameterized deep neural network~\citep{allen2018convergence,du2018gradient}. It is also observed in practice for learning deep neural networks~\citep{li2017convergence,kleinberg2018alternative}. From an optimization perspective, the PL condition has been considered extensively for designing faster optimization algorithms in the literature~\citep{karimi2016linear,reddi2016stochastic,lei2017non}. However, there still remains {\bf a big gap} between existing algorithms that focus on solving a minimization problem  and the considered min-max problem of AUC maximization.  It is a non-trivial task to leverage the PL condition of a non-convex minimization objective for developing faster primal-dual stochastic algorithms to solve its equivalent  non-convex concave min-max problem. 
% we utilize the Polyak-\L{}ojasiewicz (PL) condition which holds when using a class of neural networks as the scoring function, through which we design algorithms in spirit of the inexact proximal point method~\cite{rockafellar1976monotone,rafique2018non,davis2017proximally} with better complexity guarantees. 
The main theoretical contributions in this paper are to solve this issue. Our contributions are:
\begin{itemize}[leftmargin=*]
	%\item We reformulate the stochastic AUC maximization problem with square loss to a one-sided nonconvex min-max problem, and show that the objective satisfies PL condition when using a class of neural networks as the scoring function.
	\item We propose a stochastic algorithm named  Proximal Primal-Dual Stochastic Gradient (PPD-SG) for solving a min-max formulation of AUC maximization  under the PL condition of the surrogated AUC objective with a deep neural network. We establish a convergence rate in the order of $O(1/\epsilon)$, which is faster than that achieved by simply applying the result in~\citep{rafique2018non} to the considered problem under the PL condition, i.e., $O(1/\epsilon^3)$ and $O(n/\epsilon)$ with $n$ being the size of training set.
	%In addition, it is shown that PPD-Adagrad has adaptive iteration complexity, i.e. it has lower complexity than PPD-SG when the growth of cumulative stochastic gradient is slow. 
	%The iteration complexity of PPD-SG matches that of the algorithms in~\cite{karimi2016linear} for solving nonconvex minimization problem under PL condition. 
	
	\item In addition, we propose an AdaGrad-style primal-dual algorithm named Proximal Primal-Dual Adagrad (PPD-Adagrad), and show that it enjoys better adaptive complexity when the growth of cumulative stochastic gradient is slow. This is the first time an adaptive convergence of a stochastic AdaGrad-style algorithm is established for solving non-convex concave min-max problems. 
	
	\item We evaluate the proposed algorithms on several large-scale benchmark datasets. The experimental results show that our algorithms have superior performance than other baselines.
\end{itemize}
To the best of our knowledge, this is the first work incorporating PL condition into stochastic AUC maximization with a deep neural network as the predictive model, and more generally into solving a non-convex concave  min-max problem. Our results achieve  the state-of-the-art iteration complexity for non-convex concave min-max problems.

\section{Related Work} %{\color{red} If necessary, we  need to cut the related work for more space. If So, Trying to keep all paragraphs but shorten the discussion for individual work.}
{\bf Stochastic AUC Maximization.} Stochastic AUC maximization in the classical online setting is challenging due to its pairwise nature. There are several studies trying to update the model each time based on a new sampled/received training data. Instead of storing all examples in the memory, ~\citet{zhao2011online} employ reservoir sampling technique to maintain representative samples in a buffer, based on which their algorithms update the model. To get optimal regret bound, their buffer size needs to be $O(\sqrt{n})$, where $n$ is the number of received training examples. ~\citet{gao2013one} design a new algorithm which is not buffer-based. Instead, their algorithm needs to maintain the first-order and second-order statistics of the received data to compute the stochastic gradient, which is prohibitive for high dimensional data.  Based on a novel saddle-point reformulation of a surrogate loss of AUC proposed by~\citep{ying2016stochastic}, there are several studies~\citep{ying2016stochastic,liu2018fast,natole2018stochastic} trying to design stochastic primal-dual  algorithms.~\citet{ying2016stochastic} employ the classical  primal-dual stochastic gradient~\citep{nemirovski2009robust} and obtain $\widetilde{O}(1/\sqrt{t})$ convergence rate. \citet{natole2018stochastic} add a strongly convex regularizer, invoke composite mirror descent~\citep{duchi2010composite} and achieve $\widetilde{O}(1/t)$ convergence rate.~\citet{liu2018fast} leverage the structure of the formulation, design a multi-stage algorithm and achieve $\widetilde{O}(1/t)$ convergence rate without strong convexity assumptions. However, all of them only consider learning a linear model, which results in a convex objective function.

{\bf Non-Convex Min-max Optimization.} Stochastic optimization of non-convex min-max problems have received increasing interests recently~\citep{rafique2018non,lin2018solving,sanjabi2018solving,lu2019hybrid,jin2019minmax}. %convex-concave min-max optimization is closely related to variational inequality with monotone operators. Both stochastic mirror descent~\cite{nemirovski2009robust} and stochastic mirror-prox~\cite{juditsky2011solving} are guaranteed to enjoy $O(1/\sqrt{t})$ convergence rate in terms of the duality gap, where $t$ is the total number of iterations. 
When the objective function is weakly convex in the primal variable and is concave in the dual variable,~\citet{rafique2018non} design a proximal guided algorithm in spirit of the inexact proximal point method~\citep{rockafellar1976monotone}, which solves a sequence of convex-concave subproblems constructed by adding a quadratic proximal term in the primal variable with a periodically updated reference point. Due to the potential non-smoothness of objective function, they show the convergence to a nearly-stationary point~for the equivalent minimization problem. In the same vein as~\citep{rafique2018non},~\cite{lu2019hybrid} design an algorithm by adopting the block alternating minimization/maximization strategy and show the convergence in terms of the proximal gradient. When the objective is weakly convex and weakly concave, \citet{lin2018solving} propose a proximal algorithm which solves a strongly monotone variational inequality in each epoch and establish its convergence to stationary point. \citet{sanjabi2018solving} consider non-convex non-concave min-max games where the inner maximization problem  satisfies a PL condition, based on which they design a multi-step deterministic gradient descent ascent  with convergence to a stationary point. It is notable that our work is different in that (i) we explore the PL condition for the outer minimization problem instead of the inner maximization problem; (ii) we focus on designing stochastic algorithms instead of deterministic algorithms.  
{\bf Leveraging PL Condition for Minimization.} PL condition is first introduced by Polyak~\citep{polyak1963gradient}, which shows that gradient descent is able to enjoy linear convergence to a global minimum under this condition.~\citet{karimi2016linear} show that stochastic gradient descent, randomized coordinate descent, greedy coordinate descent are able to converge to a global minimum with faster rates under the PL condition. If the objective function has a finite-sum structure and satisfies PL condition, there are several non-convex SVRG-style algorithms~\citep{reddi2016stochastic,lei2017non,nguyen2017stochastic,zhou2018stochastic,li2018simple,wang2018spiderboost}, which are guaranteed to converge to a global minimum with a linear convergence rate. However, the stochastic algorithms in these works are developed for a minimization problem, and hence is not applicable to the min-max formulation for stochastic AUC maximization. To the best of our knowledge, \citet{liu2018fast} is the only work that leverages an equivalent condition to the PL condition (namely quadratic growth condition) to develop a stochastic primal-dual algorithm for AUC maximization with a fast rate. However, as mentioned before their algorithm and analysis rely on the convexity of the objective function, which does not hold for AUC maximization with a deep neural network.

Finally, we notice that PL condition is the key to many recent works in deep learning  for showing there is no  spurious local minima or for showing global convergence of gradient descent and stochastic gradient descent methods~\citep{hardt2016identity,li2017convergence,arora2018convergence,allen2018convergence,du2018gradient,du2018gradient1,li2018learning,allen2018convergence,zou2018stochastic,zou2019improved}. % Hardt and Ma~\cite{hardt2016identity} show that  a deep linear residual network with a square loss satisfies PL condition. 
Using the square loss, it has also been proved that the PL condition holds globally or locally for deep linear residual network~\citep{hardt2016identity}, deep linear network, one hidden layer neural network with Leaky ReLU activation~\citep{charles2017stability, zhou2017characterization}. Several studies~\citep{li2017convergence,arora2018convergence,allen2018convergence,du2018gradient,li2018learning} consider the trajectory of (stochastic) gradient descent on learning neural networks, and their analysis imply the PL condition in a certain form.  For example, %Li and Yuan~\cite{li2017convergence} consider the behavior of SGD on  a two-layer neural network with ReLU activation and show that SGD will enter a one-point strongly convex region (which implies that PL condition holds as shown in~\cite{karimi2016linear}). 
\citet{du2018gradient} show that when the width of a two layer neural network is sufficiently large, a global optimum would lie in the ball centered at the initial solution, in which PL condition holds. \citet{allen2018convergence} extends this insight further to overparameterized deep neural networks with ReLU activation, and show that the PL condition holds for a global minimum around a random initial solution.   %However, none of these analysis  are directly applicable for showing that the objective function in our setting satisfies PL condition. 

% A convergence analysis of gradient descent for deep linear neural networks

\vspace*{-0.1in}
\section{Preliminaries and Notations}
\vspace*{-0.1in}
Let $\|\cdot\|$ denote the Euclidean norm. A function $f(\x)$ is $\rho$-weakly convex if $f(\x)+\frac{\rho}{2}\|\x\|^2$ is convex, where  $\rho$ is the so-called weak-convexity parameter. A function $f(\x)$ satisfies PL condition with parameter $\mu>0$ if $f(\x)-f(\x_*)\leq \frac{1}{2\mu}\|\nabla f(\x)\|^2$, where $\x_*$ stands for the optimal solution of $f$.
Let $\z=(\x,y)\sim\P$ denote a random data following an unknown distribution $\P$, where $\x\in\X$ represents the feature vector and $y\in\Y=\{-1,+1\}$ represents the label. Denote by $\mathcal{Z}=\X\times\Y$ and by $p=\text{Pr}(y=1)=
\E_{y}\left[\mathbb{I}_{[y=1]}\right]$, where $\mathbb{I}(\cdot)$ is the indicator function. %We assume that $\X$ is a convex and compact set.

%Suppose we have a training dataset $\{(\x_i,y_i)\in\mathcal{X}\times\mathcal{Y},i=1,\ldots,n\}$ i.i.d. drawn from an underlying distribution $\P$ on $\X\times\Y$, where $\X\subset\R^d$ is a convex compact set, $\Y=\{-1,+1\}$. 
The area under the curve (AUC) on a population level for a scoring function $h:\X\rightarrow\R$ is defined as
\begin{align*}
\text{AUC}(h)=\text{Pr}\left(h(\x)\geq h(\x')\middle\vert y=1,y'=-1\right),
\end{align*}
where $\z=(\x,y)$ and $\z'=(\x',y')$ are drawn independently from $\P$.
By employing the squared loss as the surrogate for the indicator function that is a common choice used by previous studies~\citep{ying2016stochastic,gao2013one}, the AUC maximization problem can be formulated as
\begin{align*}
\min_{h\in\mathcal H}\E_{\z,\z'}\left[(1-h(\x)+h(\x'))^2\middle\vert y=1,y'=-1\right],
\end{align*}
where $\mathcal H$ denotes a hypothesis class. 
All previous works of AUC maximization assume $h(\x)=\w^\top\x$ for simplicity. Instead, we consider learning a general nonlinear model parameterized by $\w$, i.e. $h(\w; \x)$, which is not necessarily linear or convex in terms of $\w$ (e.g., $h(\w; \x)$ can be a score function defined by a neural network with weights denoted by $\w$). 
%Similar to~\citep{ying2016stochastic}, we restrict $\w$ on a bounded domain with radius $R$.
Hence, the corresponding optimization problem becomes
\begin{align}~\label{opt:obj}
\min_{\w\in \R^d}P(\w):= \E_{\z,\z'}\left[(1-h( \w; \x)+h(\w; \x'))^2 \middle\vert y=1,y'=-1\right]
\end{align}

The following proposition converts the original optimization problem~(\ref{opt:obj}) into a saddle-point problem, which is similar to Theorem 1 in~\citep{ying2016stochastic}. For completeness, the proof is included in the supplement.
\begin{prop}~\label{thm:spp}
	The optimization problem~(\ref{opt:obj}) is equivalent to
	\begin{align}~\label{opt:spp}
	\min_{\w\in\R^d,(a,b)\in\R^2}\max_{\alpha\in\R}f\left(\w,a,b,\alpha\right):=\E_{\z}\left[F\left(\w,a,b,\alpha;\z\right)\right],
	\end{align}
	where $\z=(\x,y)\sim\P$, and 
	\begin{equation*}
	\begin{aligned}
	F(\w,a,b,\alpha,\z)&=(1-p)\left(h(\w;\x)-a\right)^2\mathbb{I}_{[y=1]}+p(h(\w;\x)-b)^2\mathbb{I}_{[y=-1]}\\
	&+2\left(1+\alpha\right)\left(p h(\w;\x)\mathbb{I}_{[y=-1]}-(1-p)h(\w;\x)\mathbb{I}_{[y=1]}\right)-p(1-p)\alpha^2
	\end{aligned}
	\end{equation*}
\end{prop} 
\textbf{Remark}: It is notable that the min-max formulation~(\ref{opt:spp}) is more favorable than the original formulation~(\ref{opt:obj}) for developing a stochastic algorithm that updates the model parameters based on one example or a mini-batch of samples. For stochastic optimization of~(\ref{opt:obj}), one has to carefully sample both positive and negative examples, which is not allowed in an online setting.  It is notable that in the classical batch-learning setting, $p$ becomes the ratio of positive training examples and the expectation in~(\ref{opt:spp}) becomes average over $n$ individual functions. However, our algorithms are applicable to both batch-learning setting and online learning setting. 

% It is easy to see that $f(\w,a,b,\alpha)$ is strongly-concave in $\alpha$ if $0<p<1$.
Define $\v=(\w^\top,a,b)^\top$, $\phi(\v)=\max_{\alpha}f(\v,\alpha)$. It is clear that $\min_{\w}P(\w) =\min_{\v}\phi(\v)$ and $P(\w)\leq \phi(\v)$ for any $\v=(\w^\top,a,b)^\top$.  The following assumption is made throughout the paper.
\begin{ass}
	\label{ass:PL}
	(1) $\mu(\phi(\v)-\phi(\v_*))\leq\frac{1}{2}\|\nabla\phi(\v)\|^2,$
	where $\mu>0$ and $\v_*$ is the optimal solution of $\phi$. 
	%(2) $\E \left[h(\w;\x)\middle\vert y=-1\right]-\E \left[h(\w;\x)\middle\vert y=1\right]$ is $\tilde{L}$-Lipschitz continuous in $\w$. 
	(2) $h(\w; \x)$ is $\tilde{L}$-Lipschitz continuous in terms of $\w$ for all $\x$. 
	(3) $\phi(\v)$ is $L$-smooth.
	(4)  $\text{Var}\left[h(\w;\x)|y=-1\right]\leq\sigma^2, \text{Var}\left[h(\w;\x)|y=1\right]\leq \sigma^2$.
	(5) $0\leq h(\w;\x)\leq 1$.
	(6) Given a initial solution $\bar{\v}_0$, there exists $\Delta_0>0$ such that $\phi(\bar{\v}_0)-\phi(\v_*)\leq \Delta_0$, where $\v_*$ is the global minimum of $\phi$.
\end{ass}
%{\color{red}Keep the same notation: $\tilde L$ vs $\hat G$ below.}
\textbf{Remark}: The first condition is inspired by a PL condition on the objective function $P(\w)$ for learning a deep neural network. and the following Lemma~\ref{thm:PL1} establishes the connection. $h(\w;\x)\in[0,1]$ holds when $h$ is defined as the sigmoid function composited with the forward propagation function of a neural network.
\begin{lem}
	\label{thm:PL1}
	Suppose $\|\nabla_{\w}h(\w; \x)\|\leq \tilde L$ for all $\w$ and $\x$.
	% where $\v=(\w,a,b)$.
	If $P(\w)$ satisfies PL condition, i.e. there exists $\mu'>0$, such that $\mu' (P(\w)-\min_{\w}P(\w))\leq \frac{1}{2}\left\|\nabla_{\w}P(\w)\right\|^2,$
	then we have $\mu(\phi(\v)-\phi(\v_*))\leq\frac{1}{2}\|\nabla\phi(\v)\|^2$, 
	where $\mu=\frac{1}{\max\left(\frac{1}{2\min(p,1-p)}+\frac{2\tilde L^2}{\mu'\min(p^2,(1-p)^2)},\frac{2}{\mu'}\right)}$.
\end{lem}
\textbf{Remark}: The PL condition of $P(\w)$ could be proved for learning a neural network similar to existing studies, which is not the main focus of this paper. Nevertheless,  In Appendix~\ref{example:PL}, we provide an example for AUC maximization with one-hidden layer neural network.%By utilizing the lemma, we provide an example whose corresponding function $\phi$ satisfies PL condition (see Section \ref{example:PL} in the supplement).
%The justification of condition (1) in Assumption~\ref{ass:PL} for stochastic AUC maximization is illustrated in Section \ref{sec:PL}.

{\bf Warmup.} We first discuss the algorithms and their convergence results of~\citep{rafique2018non} applied to the considered min-max problem. They have algorithms for problems in batch-learning setting and online learning setting. Since the algorithms for the batch-learning setting have complexities scaling with $n$, we will concentrate on the algorithm for the online learning setting. The algorithm is presented in  Algorithm~\ref{alg:proximal-pd-rafique}, which is a direct application of Algorithm 2 of~\citep{rafique2018non} to an online setting. Since their analysis requires the domain of the primal and the dual variable to be bounded, hence we add a ball constraint on the primal variable and the dual variable as well. As long as $R_1$ and $R_2$ is sufficiently large, they should not affect the solution. The convergence result of  Algorithm~\ref{alg:proximal-pd-rafique} is stated below. 
\begin{thm}~\citep{rafique2018non}
	Suppose $f(\v,\alpha)$ is $\rho$-weakly convex in $\v$ and concave in $\alpha$. Let $\gamma=1/2\rho$, and define $\hat{\v}_{\tau}=\arg\min_{\v}\phi(\v)+\frac{1}{2\gamma}\|\v-\bar{\v}_{\tau}\|^2$. Algorithm~\ref{alg:proximal-pd-rafique} with $T_k =ck^2$ and   $K=\widetilde{O}(\epsilon^{-2})$ ensures that $\E\left[\text{dist}^2(0,\partial\phi(\hat{\v}_\tau))\right]\leq \frac{1}{\gamma^2}\E\|\hat{\v}_{\tau}-\bar{\v}_{\tau}\|^2\leq\epsilon^2$. The total iteration complexity is $\widetilde{O}(\epsilon^{-6})$.
\end{thm}
\vspace*{-0.1in}
{\bf Remark:}  Under the condition $\phi(\v)$ is smooth and the returned solution is within the added bounded ball constraint, the above result implies $\E[\|\nabla\phi(\bar\v_\tau)\|^2\leq \epsilon]$ with a complexity of $\widetilde O(1/\epsilon^3)$. It further implies that with a complexity of $\widetilde O(1/(\mu^3\epsilon^3))$ we have $\E[\phi(\bar\v_\tau) - \min_{\v}\phi(\v)]\leq \epsilon$ under the assumed PL condition. 

We can see that this complexity result under the PL condition of $\phi(\v)$ is worse than the typical complexity result of stochastic gradient descent method under the PL condition (i.e., $O(1/\epsilon)$)~\citep{karimi2016linear}. It remains an open problem how to design a stochastic primal-dual algorithm for solving $\min_{\v}\max_\alpha F(\v, \alpha)$ in order to achieve a complexity of $O(1/\epsilon)$  in terms of minimizing $\phi(\v)$. A naive idea is to solve the inner maximization problem of $\alpha$ first and the use SGD on the primal variable $\v$. However, this is not  viable since exact maximization over $\alpha$ is a non-trivial task.   
%Their algorithm assumes weak convexity of $f$ in terms of $\w$ and a bounded domain constraint, uses polynomially decaying stepsize, and returns a uniformly sampled solution. By this machinery, they are able to establish convergence to first order stationary point. However, this algorithm is designed for a general non-convex min-max problem and does not utilize the potential underlying structure of the objective function when using the deep neural network as a model. In addition, their polynomially decaying stepsize scheme and sampled solution are not practical in deep neural network training. For example, in the ResNet training~\cite{he2016deep}, it often uses geometrically decaying stepsize scheme and return the solution in the last iterate. To address these limitations, we exploit the PL condition and design practical algorithms with better complexity guarantees, which will be introduced in Section \ref{sec:algorithms}.

\begin{algorithm}[t]
	\caption{Proximally Guided Algorithm (PGA)~\citep{rafique2018non}}
	\begin{algorithmic}[1]
		%		\STATE \textbf{Input}: $R$, $D$, $c$, $\eta_{\v}$,  $\eta_{\alpha}$, $\gamma<\frac{1}{\rho}$
		%		\STATE Receive $\z_{-2}=(\x_{-2},y_{-2})$, $\z_{-1}=(\x_{-1},y_{-1})$ from $\P$
		%		\STATE Set $T_+=\mathbb{I}_{[y_{-2}=1]}+\mathbb{I}_{[y_{-1}=1]}$, $T_-=\mathbb{I}_{[y_{-2}=-1]}+\mathbb{I}_{[y_{-1}=-1]}$, $\ph=T_+/(T_++T_-)$
		%		\STATE $\bar{y}=(\mathbb{I}_{[y_{-2}=1]}+\mathbb{I}_{[y_{-1}=1]})/2$,  $\widehat{p(1-p)}=(\mathbb{I}_{[y_{-2}=1]}-\bar{y})^2+(\mathbb{I}_{[y_{-1}=1]}-\bar{y})^2$
		\STATE Initialize $\bar{\v}_0=\textbf{0}\in\R^{d+2}$, $\bar{\alpha}_0=0$, the global index $j=0$
		\FOR{$k=1,\ldots,K$}
		\STATE $\v_0^k=\bar{\v}_{k-1}$, $\alpha_0^k=\bar{\alpha}_{k-1}$, %$\eta_{\v}^{k}=\eta_{\v}/\sqrt{T}$, $\eta_{\alpha}^{k}=\eta_{\alpha}/\sqrt{T}$
		$\eta_k=\eta_0/k$, $T_k=T_0\cdot k^2$
		\FOR{$t=1,\ldots,T_k$}
		\STATE Receive $\z_{j}=(\x_j,y_j)$ from $\P$,  $\hat{\g}_{\v}=\nabla_{\v}F(\v_{t-1}^k,\alpha_{t-1}^k;\z_j)$, $\hat{\g}_{\alpha}=\nabla_{\alpha}F(\v_{t-1}^k,\alpha_{t-1}^k;\z_j)$
		\STATE $\v_{t}^{k}=\Pi_{\Omega_1}\left[\v_{t-1}^{k}-\eta_{k} \left(\hat{\g}_{\v}+\frac{1}{\gamma}(\v_{t-1}^{k}-\v_0^k)\right)\right]$, where $\Omega_1=\{\v: \|\v\|\leq R_1\}$
		\STATE $\alpha_t^k=\Pi_{\Omega_2}\left[\alpha_{t-1}^{k}+\eta_{k}\hat{\g}_{\alpha}\right]$, where $\Omega_2=\{\alpha:|\alpha|\leq R_2\}$
		%		\STATE Update $T_-=T_-+\mathbb{I}_{[y_j=-1]}$, $T_+=T_++\mathbb{I}_{[y_j=1]}$
		%		\STATE $\ph=T_+/(T_++T_-)$, $\bar{y}=\frac{(j+2)\bar{y}+\mathbb{I}_{[y_j=1]}}{j+3}$, $\widehat{p(1-p)}=\frac{(j+1)\widehat{p(1-p)}+(\mathbb{I}_{[y_j=1]}-\bar{y})^2}{j+2}$
		%		\STATE $j=j+1$
		\ENDFOR
		\STATE $\bar{\v}_{k}=\frac{1}{T_k}\sum_{t=1}^{T_k}\v_t^k$, $\bar{\alpha}_k=\frac{1}{T_k}\sum_{t=1}^{T_k}\alpha_t^k$
		\ENDFOR
		\STATE Sample $\tau$ uniformly randomly from $\{1,\ldots,K\}$
		\RETURN $\bar{\v}_{\tau}$, $\bar{\alpha}_{\tau}$
	\end{algorithmic}
	\label{alg:proximal-pd-rafique}
\end{algorithm}

\vspace*{-0.12in}
\section{Algorithms and Theoretical Analysis}
\vspace*{-0.12in}
\label{sec:algorithms}
In this section, we present two primal-dual algorithms for solving the min-max optimization problem (\ref{opt:spp}) with corresponding theoretical convergence results. For simplicity, we first assume the positive ratio $p$ is known in advance, which is true in the batch-learning setting.  Handling the unknown $p$ in an online learning setting is a simple extension, which will be discussed  in Section 4.3. The proposed algorithms follow the same proximal point framework proposed in~\citep{rafique2018non}, i.e., we solve the following convex-concave problems approximately and iteratively: 
\begin{align}~\label{opt:spp2}
\min_{\v}\max_{\alpha\in\R}\{f\left(\v,\alpha\right) + \frac{1}{2\gamma}\|\v -\v_0\|^2\}
\end{align} 
where $\gamma <1/L $ to ensure that the new objective function becomes convex and concave, and $\v_0$ is periodically updated. 
%Subsequently, define $\hat{\g}_t^k =\left(\nabla_{\v}F(\v_{t}^k,\alpha_{t}^k;\z)^\top+\frac{1}{\gamma}\left(\v_{t}^k-\v_0^k\right)^\top,-\nabla_{\alpha}F(\v_{t}^k,\alpha_{t}^k;\z)^\top\right)^\top$.
%%\g_t^k &=\left(\nabla_{\v}f(\v_{t}^k,\alpha_{t}^k)^\top+\frac{1}{\gamma}\left(\v_{t}^k-\v_0^k\right)^\top,-\nabla_{\alpha} f(\v_{t}^k,\alpha_{t}^k)^\top\right)^\top.

%	For simplicity of presentation,  we introduce the following notation:
%\begin{align*}
%\hat{\g}_t^k& =\left(\nabla_{\v}F(\v_{t}^k,\alpha_{t}^k;\z)^\top+\frac{1}{\gamma}\left(\v_{t}^k-\v_0^k\right)^\top,-\nabla_{\alpha}F(\v_{t}^k,\alpha_{t}^k;\z)^\top\right)^\top,\\
%%\g_t^k &=\left(\nabla_{\v}f(\v_{t}^k,\alpha_{t}^k)^\top+\frac{1}{\gamma}\left(\v_{t}^k-\v_0^k\right)^\top,-\nabla_{\alpha} f(\v_{t}^k,\alpha_{t}^k)^\top\right)^\top.
%\end{align*}
%which is stochastic gradient of the new objective in~(\ref{opt:spp2}) at the $k$-th outer loop.

%$$\sigma^2=\text{Var}\left[h(\w;\x)|y=-1\right]+\text{Var}\left[h(\w;\x)|y=1\right].$$
\vspace*{-0.1in}
\subsection{Proximal Primal-Dual Stochastic Gradient}
\vspace*{-0.1in}
Our first algorithm named Proximal Primal-Dual Stochastic Gradient (PPD-SG) is presented in Algorithm~\ref{alg:proximal-pd-PL}. Similar to Algorithm~\ref{alg:proximal-pd-rafique}, it has a nested loop, where the inner loop is to approximately solve a regularized min-max optimization problem~(\ref{opt:spp2}) using stochastic primal-dual gradient method, and the outer loop updates the reference point and learning rate. One key  difference is that PPD-SG uses a geometrically decaying step size scheme, while Algorithm~\ref{alg:proximal-pd-rafique} uses a polynomially decaying step size scheme. Another key difference is that at the end of $k$-th outer loop, we update the dual variable $\bar\alpha_k$ in Step 12, which is motivated by its closed-form solution given $\bar\v_k$. In particular, the given $\bar\v_k$, the dual solution that optimizes the inner maximization problem is given by: 
\begin{align*}
\alpha = \frac{\E[h(\bar\w_k; \x) I_{y=-1}] }{1-p} - \frac{\E[h(\bar\w_k; \x)I_{y=1}]}{p} =  \E_{\x}[h(\bar\w_k; \x)|y=-1]- \E_\x[h(\bar\w_k; \x)|y=1].
\end{align*} 
In the algorithm,  we only use a small number of samples in Step 11 to compute an estimation of the optimal $\alpha$ given $\bar\v_k$.  These differences are important for us to achieve  lower iteration complexity of PPD-SG. Next, we present our convergence results of PPD-SG. 
%\begin{ass}
%	\label{ass:boundedgradientl2}
%	$\|\hat{\g}_{t}^k\|_{2}\leq G$ holds for any $t$ and $k$.
%\end{ass}
\setlength{\textfloatsep}{0.1cm}
\setlength{\floatsep}{0.1cm}
\begin{algorithm}[t]
	\caption{Proximal Primal-Dual Stochastic Gradient (PPD-SG)}
	\begin{algorithmic}[1]
		%\STATE \textbf{Input}:  $\eta_0$, $\gamma$
		%\STATE Receive $\z_{-2}=(\x_{-2},y_{-2})$, $\z_{-1}=(\x_{-1},y_{-1})$ from $\P$
		%		\STATE Update $T_+,T_-,\ph,\widehat{p(1-p)}$ using data $\{\z_{-2},\z_{-1}\}$
		%\STATE Set $T_+=\mathbb{I}_{[y_{-2}=1]}+\mathbb{I}_{[y_{-1}=1]}$, $T_-=\mathbb{I}_{[y_{-2}=-1]}+\mathbb{I}_{[y_{-1}=-1]}$, $\ph=T_+/(T_++T_-)$
		%\STATE $\bar{y}=(\mathbb{I}_{[y_{-2}=1]}+\mathbb{I}_{[y_{-1}=1]})/2$,  $\widehat{p(1-p)}=(\mathbb{I}_{[y_{-2}=1]}-\bar{y})^2+(\mathbb{I}_{[y_{-1}=1]}-\bar{y})^2$
		\STATE Initialize $\bar{\v}_0=\textbf{0}\in\R^{d+2}$, $\bar{\alpha}_0=0$, the global index $j=0$
		\FOR{$k=1,\ldots,K$}
		\STATE $\v_0^k=\bar{\v}_{k-1}$, $\alpha_0^k=\bar{\alpha}_{k-1}$,  $\eta_k=\eta_0\exp\left(-(k-1)\frac{\mu/L}{5+\mu/L}\right)$
		\FOR{$t=1,\ldots,T_k-1$}
		\STATE Receive $\z_{j}=(\x_j,y_j)$ from $\P$, $\hat{\g}_{\v}=\nabla_{\v}F(\v_{t-1}^k,\alpha_{t-1}^k;\z_j)$, $\hat{\g}_{\alpha}=\nabla_{\alpha}F(\v_{t-1}^k,\alpha_{t-1}^k;\z_j)$
		\STATE $\v_{t}^{k}=\v_{t-1}^{k}-\eta_{k} \left(\hat{\g}_{\v}+\frac{1}{\gamma}(\v_{t-1}^{k}-\v_0^k)\right)$
		\STATE $\alpha_t^k=\alpha_{t-1}^{k}+\eta_{k}\hat{\g}_{\alpha}$
		\STATE $j=j+1$
		%\STATE Update $T_+,T_-,\ph,\widehat{p(1-p)}$,$\bar{y}$ using data $\{\z_{j}\}$ by Algorithm \ref{alg:update_stat}
		%		\STATE Update $T_-=T_-+\mathbb{I}_{[y_j=-1]}$, $T_+=T_++\mathbb{I}_{[y_j=1]}$
		%		\STATE $\ph=T_+/(T_++T_-)$, $\bar{y}=\frac{(j+2)\bar{y}+\mathbb{I}_{[y_j=1]}}{j+3}$, $\widehat{p(1-p)}=\frac{(j+1)\widehat{p(1-p)}+(\mathbb{I}_{[y_j=1]}-\bar{y})^2}{j+2}$
		%		\STATE $j=j+1$
		\ENDFOR
		\STATE $\bar{\v}_{k}=\frac{1}{T_k}\sum_{t=0}^{T_k-1}\v_t^k$
		\STATE Draw a minibatch $\{\z_j,\ldots,\z_{j+m_k-1}\}$ of size $m_k$
		\STATE $\bar{\alpha}_k=\frac{\sum_{i=j}^{j+m_k-1}h(\bar{\w}_k;\x_i)\mathbb{I}_{y_i=-1}}{\sum_{i=j}^{j+m_k-1}\mathbb{I}_{y_i=-1}}-\frac{\sum_{i=j}^{j+m_k-1}h(\bar{\w}_k;\x_i)\mathbb{I}_{y_i=1}}{\sum_{i=j}^{j+m_k-1}\mathbb{I}_{y_i=1}}$
		\STATE $j=j+m_k$
		%\STATE Update $T_+,T_-,\ph,\widehat{p(1-p)}$,$\bar{y}$ using data $\{\z_{j},\ldots,\z_{j+m-1}\}$ by Algorithm \ref{alg:update_stat}
		\ENDFOR
		%\STATE Sample $\tau$ uniformly randomly from $\{1,\ldots,K\}$
		\RETURN $\bar{\v}_{K}$, $\bar{\alpha}_{K}$
	\end{algorithmic}
	\label{alg:proximal-pd-PL}
\end{algorithm}
\begin{lem}[One Epoch Analysis of Algorithm \ref{alg:proximal-pd-PL}]
	\label{lemma:PDSGD2}
	Suppose Assumption \ref{ass:PL} and there exists $G>0$ such that $\|\hat{\g}_{t}^k\|_{2}\leq G$, where $\hat{\g}_t^k =\left(\nabla_{\v}F(\v_{t}^k,\alpha_{t}^k;\z)^\top+\frac{1}{\gamma}\left(\v_{t}^k-\v_0^k\right)^\top,-\nabla_{\alpha}F(\v_{t}^k,\alpha_{t}^k;\z)^\top\right)^\top$. Define $\phi_k(\v)=\phi(\v)+\frac{1}{2\gamma}\|\v-\bar{\v}_{k-1}\|^2,$ $\s_k=\arg\min_{\v\in\R^{d+2}} \phi_k(\v)$. Choosing $m_{k-1}\geq \frac{2(\sigma^2+C)}{p(1-p)\eta_k^2 G^2T_k}$ with $C=\frac{2}{\ln(\frac{1}{\max(p,1-p)})}\max(p,1-p)^{\frac{1}{\ln(1/\max(p,1-p))}}$, then we have 
	\begin{align*}
	\E_{k-1}\left[\phi_k(\bar{\v}_k)-\min_{\v}\phi_k(\v)\right]\leq  \frac{\|\bar{\v}_{k-1}-\s_k\|^2+16\tilde{L}^2\E_{k-1}\|\bar{\v}_{k-1}-\bar{\v}_k\|^2}{2\eta_k T_k}+4\eta_k G^2.
	\end{align*}
	where $\E_{k-1}$ stands for the conditional expectation conditioning on all the stochastic events until $\bar{\v}_{k-1}$ is generated.
\end{lem}
\begin{thm}
	\label{theorem:sgd}
	Suppose the same conditions in Lemma~\ref{lemma:PDSGD2} hold. Set $\eta_k=\eta_0\exp\left(-(k-1)\frac{\mu/L}{5+\mu/L}\right)$ and $T_k=\frac{\max(2,16\tilde{L}^2)}{L\eta_0}\exp\left((k-1)\frac{\mu/L}{5+\mu/L}\right)$, $m_k=\frac{2(\sigma^2+C)L}{p(1-p)G^2\eta_0\max(2,16\tilde{L}^2)}\exp\left(k\frac{\mu/L}{5+\mu/L}\right)$ with $C=\frac{2}{\ln(\frac{1}{\max(p,1-p)})}\max(p,1-p)^{\frac{1}{\ln(1/\max(p,1-p))}}$, $\gamma=\frac{1}{2L}$ in Algorithm \ref{alg:proximal-pd-PL}. To return $\bar{\v}_K$ such that $\E\left[\phi(\bar{\v}_K)-\phi(\v_*)\right]\leq \epsilon$, it suffices to choose  $K\geq  \left(\frac{5L}{\mu}+1\right)\max\left(\log\frac{2\Delta_0}{\epsilon},\log K+\log \frac{48G^2\eta_0}{5\epsilon}\right)$. The number of iterations is at most $\widetilde{O}\left(\frac{LG^2}{\mu^2\epsilon}\right)$, and the required number of samples is at most $\widetilde{O}\left(\frac{L^3\sigma^2}{\mu^2\epsilon}\right)$, where $\widetilde{O}(\cdot)$ hides logarithmic factors of $L,\mu,\epsilon,\delta$.  where $\widetilde{O}(\cdot)$ hides logarithmic factor of 
	%$\mu, \epsilon$.
	$L, \mu, \epsilon, G, \sigma$.
\end{thm}
\vspace*{-0.1in}
\textbf{Remark}: The above complexity result is similar to that of~\citep{karimi2016linear} for solving non-convex minimization problem under the PL condition up to a logarithmic factor. Compared with the complexity result of  Algorithm~\ref{alg:proximal-pd-rafique} discussed earlier, i.e., $\widetilde O(1/(\mu^3\epsilon^3))$, the above complexity in the order of $\widetilde  O(1/(\mu^2\epsilon))$ is much better - it not only improves the dependence on $\epsilon$ but also improves the dependence on $\mu$. %Let us compare our Algorithm~\ref{alg:proximal-pd-PL} with algorithms in~\cite{rafique2018non} under our setting. %With PL condition, their Algorithm 2 in their case D1 has complexity $\widetilde{O}(1/\epsilon^3)$ (see Theorem 1 in~\cite{rafique2018non}). Their Algorithm 2 in their case D2, Algorithm 3, 4 are not applicable in our setting since they need to assume finite sum structure.

\vspace*{-0.1in}
\subsection{Proximal Primal-Dual Adagrad}
\vspace*{-0.1in}
Our second algorithm named Proximal Primal-Dual Adagrad (PPD-Adagrad) is a AdaGrad-style algorithm. Since it only differs from PPD-SG in the updates of the inner loop, we only present the inner loop in Algorithm~\ref{alg:proximal-pdadagrad-PL}. The updates in the inner loop are similar to the adaptive updates of traditional AdaGrad~\citep{duchi2011adaptive}. We aim to achieve an adaptive convergence by using PPD-AdaGrad. The analysis of PPD-AdaGrad is inspired by the analysis of AdaGrad for non-convex minimization problems~\citep{chen2018universal}. The key difference is that we have to carefully deal with the primal-dual updates for the non-convex min-max problem. We summarize the convergence results of PPD-AdaGrad below. 
%Similar to PPD-SG, the Proximal Primal-Dual Adagrad (PPD-Adagrad) also has nested loops. The difference of PPD-SG and PPD-Adagrad lies in the approach for solving the regularized min-max problem in the inner loop: PPD-Adagrad employs Adagrad while PPD-SG invokes stochastic primal-dual gradient method. PPD-Adagrad has adaptive iteration complexity, as is shown in Theorem \ref{theorem:adagrad}.
%\begin{ass}
%	\label{ass:boundedgradient}
%	$\|\hat{\g}_{t}^k\|_{\infty}\leq \delta$ holds for any $t$ and $k$.
%\end{ass}
\begin{algorithm}[t]
	\caption{Inner Loop of Proximal Primal-Dual AdaGrad (PPD-AdaGrad)}
	\begin{algorithmic}[1]
		%\STATE \textbf{Input}:  $\eta_0$, $\gamma$
		%\STATE Receive $\z_{-2}=(\x_{-2},y_{-2})$, $\z_{-1}=(\x_{-1},y_{-1})$ from $\P$
		%		\STATE Update $T_+,T_-,\ph,\widehat{p(1-p)}$ using data $\{\z_{-2},\z_{-1}\}$
		%\STATE Set $T_+=\mathbb{I}_{[y_{-2}=1]}+\mathbb{I}_{[y_{-1}=1]}$, $T_-=\mathbb{I}_{[y_{-2}=-1]}+\mathbb{I}_{[y_{-1}=-1]}$, $\ph=T_+/(T_++T_-)$
		%\STATE $\bar{y}=(\mathbb{I}_{[y_{-2}=1]}+\mathbb{I}_{[y_{-1}=1]})/2$,  $\widehat{p(1-p)}=(\mathbb{I}_{[y_{-2}=1]}-\bar{y})^2+(\mathbb{I}_{[y_{-1}=1]}-\bar{y})^2$
		%\STATE Initialize $\bar{\v}_0=\textbf{0}\in\R^{d+2}$, $\bar{\alpha}_0=0$, the global index $j=0$
		%\FOR{$k=1,\ldots,K$}
		%\STATE $\v_0^k=\bar{\v}_{k-1}$, $\alpha_0^k=\bar{\alpha}_{k-1}$,  $\eta_k=\eta_0\exp\left(-\frac{k-1}{2}\frac{\mu/L}{5+\mu/L}\right)$
		\FOR{$t=1,\ldots,T_k-1$}
		\STATE Receive $\z_{j}=(\x_j,y_j)$ from $\P$, $\hat{\g}_{\v}=\nabla_{\v}F(\v_{t}^k,\alpha_{t}^k;\z_j)$, $\hat{\g}_{\alpha}=\nabla_{\alpha}F(\v_{t}^k,\alpha_{t}^k;\z_j)$
		\STATE  $\hat{\g}_t^k=[\hat{\g}_{\v}+\frac{1}{\gamma}(\v_t^k-\v_0^k);-\hat{\g}_{\alpha}]\in\R^{d+3}$, $\hat{\g}_{1:t}^k=[\hat{\g}_{1:t-1}^k,\hat{\g}_t^k]$, $s_{t,i}^{k}=\left\|\hat{\g}_{1:t,i}^{k}\right\|_2$, %$\g_{\alpha}^{1:t}=[\g_{\alpha}^{1:t-1},\hat{\g}_{\alpha}]$, $s_{\alpha}^{t,i}=\left\|\hat{\g}_{\alpha}^{1:t,i}\right\|_2$
		\STATE $H_t^{k}=\delta I+\text{diag}(s_{t}^k)$, $\psi_t^{k}(\u)=\frac{1}{2}\langle\u-\u_0^k,H_t^{k}(\u-\u_0^k)\rangle$, where $\u_0^k=[\v_0^k;\alpha_0^k]\in\R^{d+3}$
		%$H_t^{\alpha}=\delta I+\text{diag}(s_{\alpha}^{t})$, $\psi_t^{\alpha}(\x)=\frac{1}{2}\langle\x,H_t^{\alpha}\x\rangle$
		\STATE 
		$\u_{t+1}^{k}=\arg\min\limits_{\u}\left\{\eta_k\langle\frac{1}{t}\sum_{\tau=1}^{t}\hat{\g}_\tau^k,\u \rangle+\frac{1}{t}\psi_t^k(\u)\right\}$
		%$\v_{t+1}^{k}=\arg\min\limits_{\v}\left\{\eta_k\langle \hat{\g}_{\v}+\frac{1}{\gamma}(\v_{t}^{k}-\v_0^k),\v\rangle+B_{\psi_t^{\v}}(\v,\v_t^{k})\right\}$
		%\STATE $\alpha_{t+1}^k=\arg\min\limits_{\alpha}\left\{\eta_k\langle-\hat{\g}_{\alpha},\alpha\rangle+B_{\psi_t^{\alpha}}(\alpha,\alpha_t^k)\right\}$
		%\STATE $\v_{t}^{k}=\v_{t-1}^{k}-\eta_{k} \left(\hat{\g}_{\v}+\frac{1}{\gamma}(\v_{t-1}^{k}-\v_0^k)\right)$
		%\STATE $\alpha_{t+1}^k=\alpha_{t}^{k}+\eta_{k}\hat{\g}_{\alpha}$
		%\STATE $j=j+1$
		%\STATE Update $T_+,T_-,\ph,\widehat{p(1-p)}$ using data $\{\z_{j}\}$ by Algorithm \ref{alg:update_stat}
		%		\STATE Update $T_-=T_-+\mathbb{I}_{[y_j=-1]}$, $T_+=T_++\mathbb{I}_{[y_j=1]}$
		%		\STATE $\ph=T_+/(T_++T_-)$, $\bar{y}=\frac{(j+2)\bar{y}+\mathbb{I}_{[y_j=1]}}{j+3}$, $\widehat{p(1-p)}=\frac{(j+1)\widehat{p(1-p)}+(\mathbb{I}_{[y_j=1]}-\bar{y})^2}{j+2}$
		%		\STATE $j=j+1$
		\ENDFOR
		%\STATE $\bar{\v}_{k}=\frac{1}{T_k}\sum_{t=1}^{T_k}\v_t^k$
		%\STATE Receive minibatch $\{\z_j,\ldots,\z_{j+m_k-1}\}$ of size $m_k$
		%\STATE $\bar{\alpha}_k=\frac{\sum_{i=j}^{j+m_k-1}h(\bar{\w}_k;\x_j)\mathbb{I}_{y_j=-1}}{\sum_{i=j}^{j+m_k-1}\mathbb{I}_{y_j=-1}}-\frac{\sum_{i=j}^{j+m_k-1}h(\bar{\w}_k;\x_j)\mathbb{I}_{y_j=1}}{\sum_{i=j}^{j+m_k-1}\mathbb{I}_{y_j=1}}$
		%\STATE $j=j+m_k$
		%\STATE Update $T_+,T_-,\ph,\widehat{p(1-p)}$ using data $\{\z_{j},\ldots,\z_{j+m_k-1}\}$ by Algorithm \ref{alg:update_stat}
		%\ENDFOR
		%\STATE Sample $\tau$ uniformly randomly from $\{1,\ldots,K\}$
		%\RETURN $\bar{\v}_{K}$, $\bar{\alpha}_{K}$
	\end{algorithmic}
	\label{alg:proximal-pdadagrad-PL}
\end{algorithm}

\begin{lem}[One Epoch Analysis of Algorithm \ref{alg:proximal-pdadagrad-PL}]
	\label{lemma:PDadagrad}
	Suppose Assumption \ref{ass:PL} and $\|\hat{\g}_{t}^k\|_{\infty}\leq \delta$ hold.
	Define $\phi_k(\v)=\phi(\v)+\frac{1}{2\gamma}\|\v-\bar{\v}_{k-1}\|^2,$ $\s_k=\arg\min_{\v\in\R^{d+2}} \phi_k(\v)$. Choosing $m_{k-1}\geq \frac{2(\sigma^2+C)}{p(1-p)(d+3)\eta_k^2}$ with $C=\frac{2}{\ln(\frac{1}{\max(p,1-p)})}\max(p,1-p)^{\frac{1}{\ln(1/\max(p,1-p))}}$, and {\scriptsize $$T_k=\inf\left\{\tau:\tau\geq M_k\max\left(\frac{(\delta+\max_{i}\|\hat{\g}_{1:\tau,i}^{k}\|_2)\max(1,8\tilde{L}^2)}{c},2c(\sum_{i=1}^{d+3}\|\hat{\g}_{1:\tau,i}^{k}\|_2+(d+3)(\delta+\max_{i}\|\hat{\g}_{1:\tau,i}^k\|_2))\right)\right\}$$} with $M_k>0$, $c>0$, then we have
	%	\begin{align*}
	%		\E_{k-1}\left[\phi_k(\bar{\v}_k)-\min_{\v}\phi_k(\v)\right]\leq  \E_{k-1}\left(\frac{\delta+\max_{i}\|\hat{\g}_{1:T_k,i}^k\|_2}{2\eta_kT_k}\right)(4+3\tilde{L}^2)\|\bar{\v}_{k-1}-\s_k\|^2+\E_{k-1}\left(\frac{\eta_k}{T_k}\sum_{i=1}^{d+3}\|\hat{\g}_{1:T_k}^k\|_2\right)
	%	\end{align*}
	\begin{align*}
	\E_{k-1}\left[\phi_k(\bar{\v}_k)-\min_{\v}\phi_k(\v)\right]\leq \frac{c\left(\|\bar{\v}_{k-1}-\s_k\|_2^2+\E_{k-1}\|\bar{\v}_{k-1}-\bar{\v}_k\|_2^2\right)}{\eta_k M_k}+\frac{\eta_k}{cM_k}.
	\end{align*}
	where $\E_{k-1}$ stands for the conditional expectation conditioning on all the stochastic events until $\bar{\v}_{k-1}$ is generated.
\end{lem}
\begin{thm}
	\label{theorem:adagrad}
	Suppose the same conditions as in Lemma~\ref{lemma:PDadagrad} hold. Set $\eta_k=\eta_0\exp\left(-\frac{(k-1)}{2}\frac{\mu/L}{5+\mu/L}\right)$, $M_k=\frac{4c}{L\eta_0}\exp\left(\frac{(k-1)}{2}\frac{\mu/L}{5+\mu/L}\right) $, $m_k= \frac{2(\sigma^2+C)}{p(1-p)\eta_0^2(d+3)}\exp\left(k\frac{\mu/L}{5+\mu/L}\right)$ with $C=\frac{2}{\ln(\frac{1}{\max(p,1-p)})}\max(p,1-p)^{\frac{1}{\ln(1/\max(p,1-p))}}$, $\gamma=\frac{1}{2L}$ and  $T_k$ as in Lemma~\ref{lemma:PDadagrad} 
	%$$T_k=\inf\left\{\tau:\tau\geq M_k\max\left(\frac{(\delta+\max_{i}\|\hat{\g}_{1:\tau,i}^{k}\|_2)\max(1,8\tilde{L}^2)}{2c},c\left(\sum_{i=1}^{d+3}\|\hat{\g}_{1:\tau,i}^{k}\|_2+(d+3)\left(\delta+\max_{i}\|\hat{\g}_{1:\tau,i}^k\|_2\right)\right)\right)\right\},$$
	where $c=\frac{1}{\sqrt{d+3}}$. Suppose $\|\hat{\g}_{1:T_k,i}^k\|_2\leq \delta \cdot T_k^{\alpha}$ for $\forall k$, where $0\leq \alpha\leq\frac{1}{2}$. To return $\bar{\v}_K$ such that $\E\left[\phi(\bar{\v}_K)-\phi(\v_*)\right]\leq \epsilon$, it suffices to choose  $K\geq  \left(\frac{5L}{\mu}+1\right)\max\left(\log\frac{2\Delta_0}{\epsilon},\log K+\log \frac{\eta_0^2L}{5c^2\epsilon}\right)$. The number of iterations is at most $\widetilde{O}\left(\left(\frac{L\delta^2 d}{\mu^2\epsilon}\right)^{\frac{1}{2(1-\alpha)}}\right)$, and the required number of samples is at most $\widetilde{O}\left(\frac{L^3\sigma^2}{\mu^2\epsilon}\right)$, where $\widetilde{O}(\cdot)$ hides logarithmic factors of $L,\mu,\epsilon,\delta$.
\end{thm}
\vspace*{-0.1in}
\textbf{Remark}: When the cumulative growth of stochastic gradient is slow, i.e., $\alpha<1/2$, the number of iterations is less than that in Theorem \ref{theorem:sgd}, which exhibits adaptive iteration complexity.

%\vspace*{-0.1in}

\subsection{Extensions}
%\vspace*{-0.1in}
\begin{algorithm}[t]
	\caption{Update $T_+,T_-,\ph,\widehat{p(1-p)},\bar{y}$ given data $\{\z_j,\ldots,\z_{j+m-1}\}$}
	\label{alg:keep}
	\begin{algorithmic}[1]
		\STATE Update $T_-=T_-+\sum_{i=j}^{j+m-1}\mathbb{I}_{[y_j=-1]}$, $T_+=T_++\sum_{i=j}^{j+m-1}\mathbb{I}_{[y_j=1]}$
		\STATE $\ph=T_+/(T_++T_-)$, $\bar{y}=\frac{(j+2)\bar{y}+\sum_{i=j}^{j+m-1}\mathbb{I}_{[y_i=1]}}{j+m+2}$, $\widehat{p(1-p)}=\frac{(j+1)\widehat{p(1-p)}+\sum_{i=j}^{j+m-1}(\mathbb{I}_{[y_i=1]}-\bar{y})^2}{j+m+1}$
	\end{algorithmic}
	\label{alg:update_stat}
\end{algorithm}

{\bf Setting $\eta_k, T_k, m_k$.} It is notable that the setting of $\eta_k, T_k, m_k$ depends on unknown parameters $\mu$, $L$, etc., which are typically unknown. One heuristic to address this issue is that we can decrease $\eta_k$ by a constant factor larger than $1$ (e.g., 2 or 5 or 10), and similarly increase $T_k$ and $m_k$ by a constant factor. Another heuristic is to decrease the step size by a constant factor when the performance on a validation data saturates~\citep{krizhevsky2012imagenet}.%, and choose a reasonably large value for $m_k$.  %We use the second heuristic in our experiments. 

{\bf Variants when $p$ is unknown}. In the online learning setting when $p$ is unknown, the stochastic gradients of $f$ in both $\v$ and $\alpha$ are not directly available. To address this issue, we can keep unbiased estimators for both $p$ and $p(1-p)$ which are independent of the new arrived data, and update these estimators during the optimization procedure. All values depending on $p$ and $p(1-p)$ (i.e., $F,\g_{\v},\g_{\alpha}$) are estimated by substituting $p$ and $p(1-p)$ by $\ph$ and $\widehat{p(1-p)}$ (i.e., $\hat{F},\hat{\g}_{\v},\hat{\g}_{\alpha}$) respectively. 
%Specifically,
%\begin{equation*}
%\begin{aligned}
%\hat{F}(\w,a,b,\alpha,\z)&=(1-\ph)\left(h(\w;\x)-a\right)^2\mathbb{I}_{[y=1]}+\ph(h(\w;\x)-b)^2\mathbb{I}_{[y=-1]}\\
%&+2\left(1+\alpha\right)\left(\ph h(\w;\x)\mathbb{I}_{[y=-1]}-(1-\ph)h(\w;\x)\mathbb{I}_{[y=1]}\right)-\widehat{p(1-p)}\alpha^2
%\end{aligned}
%\end{equation*}
%\begin{align*}
%\hat{\g}_{\v}(\v,\alpha,\z)=\nabla_{\v}\hat{F}(\w,a,b,\alpha,\z)\\
%\hat{\g}_{\alpha}(\v,\alpha,\z)=\nabla_{\alpha}\hat{F}(\w,a,b,\alpha,\z)\\
%\end{align*}
The approach for keeping unbiased estimator $\hat{p}$ and $\widehat{p(1-p)}$ during the optimization is described in Algorithm \ref{alg:keep}, where $j$ is the global index, and $m$ is the number of examples received.
%in one iteration.
%{\color{red} What is definition of one iteration here? }

{\bf Extensions to multi-class problems.} In the previous analysis, we only consider the binary classification problem. We can extend it to the multi-class setting. To this end, we first introduce the definition of AUC in this setting according to~\citep{hand2001simple}. Suppose there are $c$ classes, we have $c$ scoring functions for each class, namely $h(\w_1;\x),\ldots,h(\w_c;\x)$. We assume that these scores are normalized such that $\sum_{k=1}^ch(\w_c; \x) = 1$. 
Note that if these functions are implemented by a deep neural network, they can share the lower layers and have individual last layer of connections. The AUC is defined as
\begin{equation}
\label{def:AUC:multiclass}
\text{AUC}(h)=\frac{1}{c(c-1)}\sum_{i=1}^{c}\sum_{j\neq i}\text{Pr}\left(h(\w_i, \x)\geq  h(\w_i; \x')\middle\vert y=i, y'=j\right),
\end{equation}
%where $\hat p_i(\x)$ stands for the probability that a sample $\x$ is predicted as in the $i$-th class, i.e., $\hat p_i(\x)=\frac{\exp(h(\w_i;\x))}{\sum_{k=1}^{c}\exp(h(\w_k;\x))}$. Noting that $\text{Pr}\left(\hat p_i(\x)\geq \hat p_j(\x')\middle\vert y=i, y'=j\right)=\text{Pr}\left(h(\w_i;\x)\geq h(\w_i;\x')\middle\vert y=i, y'=j\right)$ 
Similar to Proposition~\ref{thm:spp}, we can cast the problem into 
\begin{equation}
\label{eq:multiclassAUC}
\min_{\w,\a,\b}\max_{\alpha}\frac{1}{c(c-1)}\sum_{i=1}^{c}\sum_{j\neq i}\E_{\z}\left[F_{ij}\left(\w_i,a_{ij},b_{ij},\alpha_{ij};\z\right)\right],
\end{equation}
where $\w=[\w_1,\ldots,\w_c]$, $\a,\b,\alpha\in\R^{c\times c}$, $i,j=1,\ldots,c$, $\z=(\x,y)\sim\P$, $p_i=\text{Pr}(y=i)$, and 
\begin{equation*}
\begin{aligned}
F_{ij}(\w_i,a_{ij},b_{ij},\alpha_{ij},\z)&=p_j\left(h(\w_i;\x)-a_{ij}\right)^2\mathbb{I}_{[y=i]}+p_i(h(\w_i;\x)-b_{ij})^2\mathbb{I}_{[y=j]}\\
&+2\left(1+\alpha_{ij}\right)\left(p_i h(\w_i;\x)\mathbb{I}_{[y=j]}-p_jh(\w_i;\x)\mathbb{I}_{[y=i]}\right)-p_ip_j\alpha_{ij}^2.
\end{aligned} 
\end{equation*}
Then we can modify our algorithms to accommodate the multiple class pairs. We can also add another level of sampling of class pairs into computing the stochastic gradients. 

\begin{figure}[ht]
	\centering
	%\hspace{-10pt}
	\includegraphics[scale=0.19]{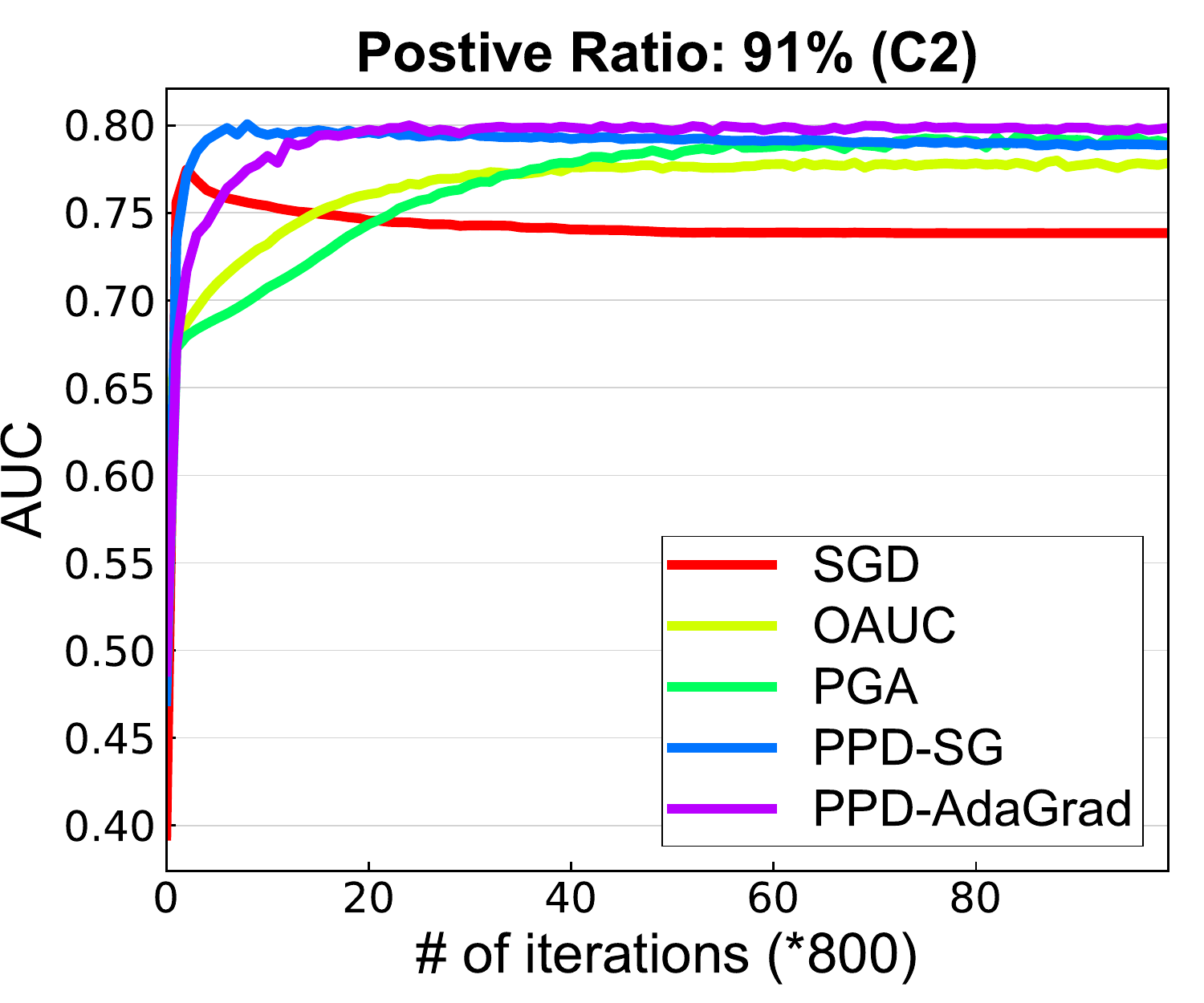}
	\hspace{10pt}\includegraphics[scale=0.19]{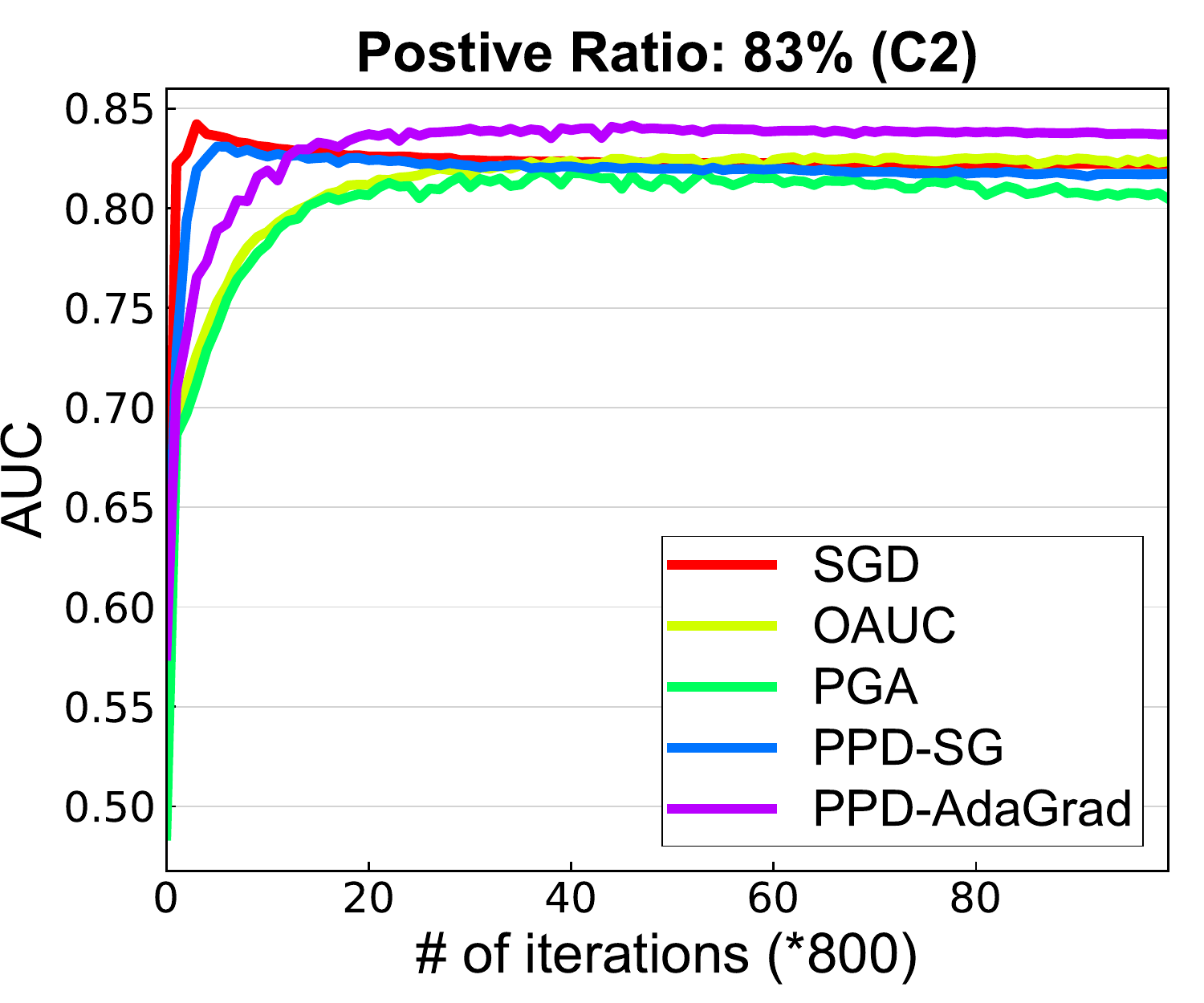}
	\hspace{10pt}\includegraphics[scale=0.19]{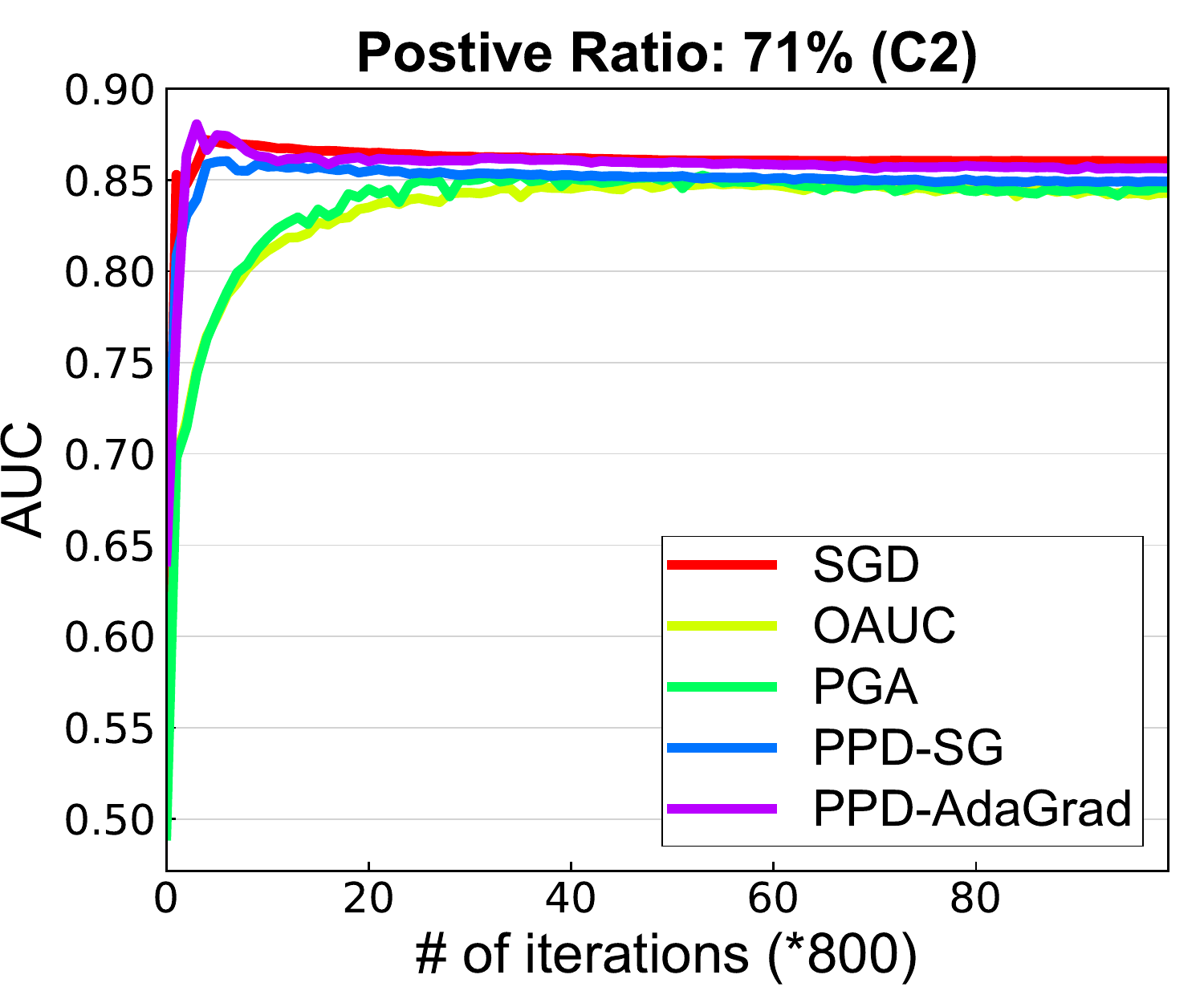}
	\hspace{10pt}\includegraphics[scale=0.19]{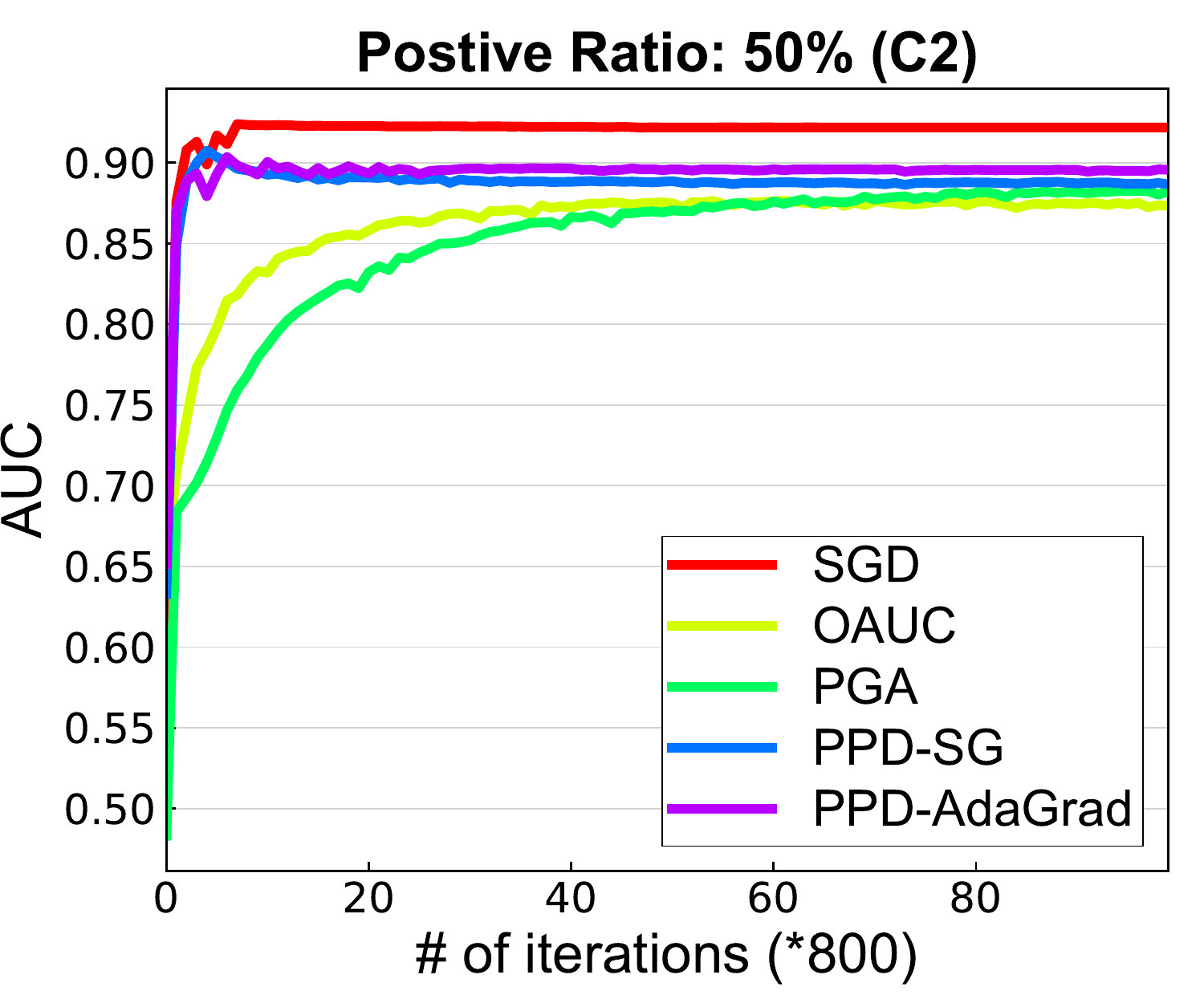}
	
	%\hspace{-10pt}
	\includegraphics[scale=0.19]{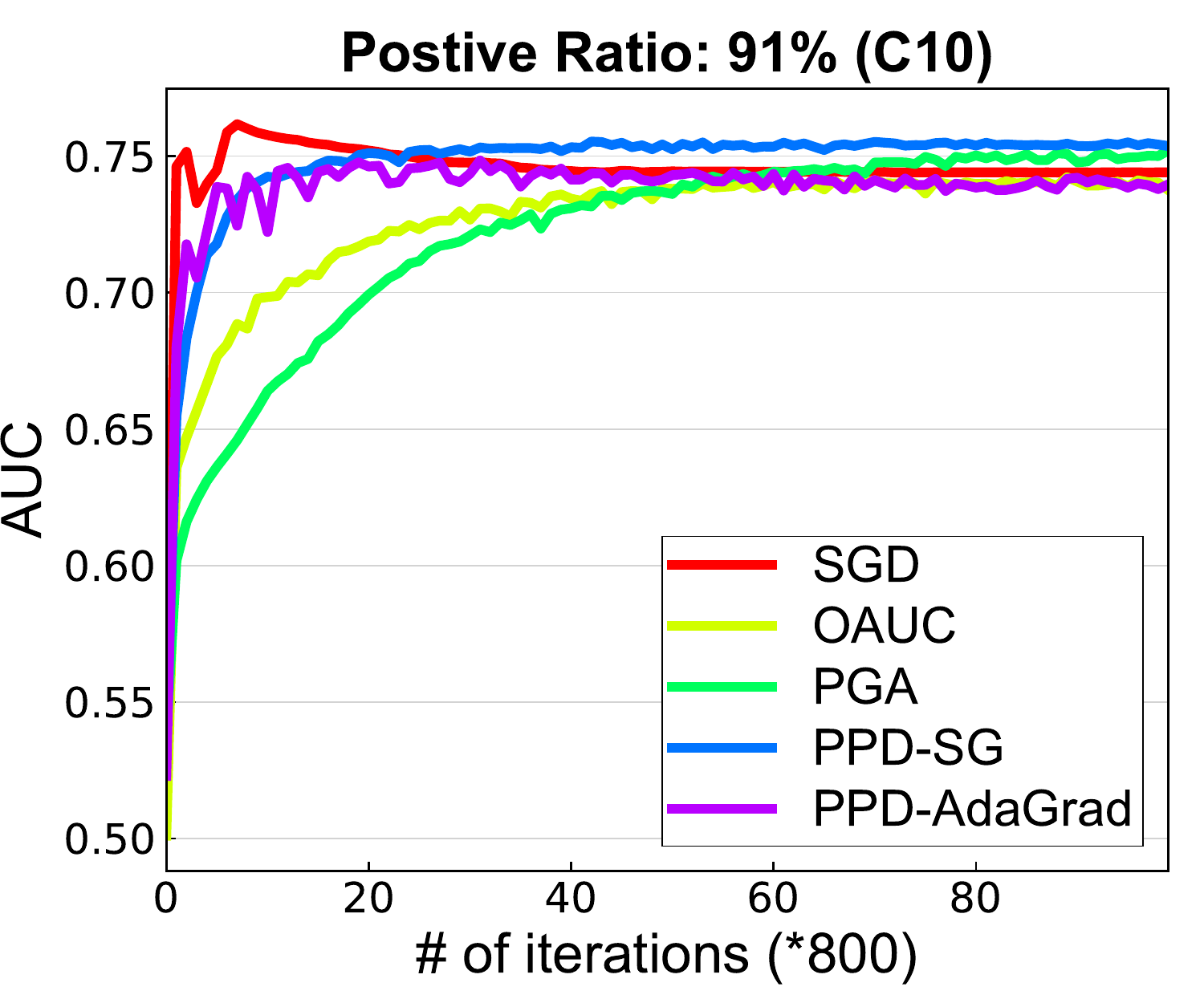}
	\hspace{10pt}\includegraphics[scale=0.19]{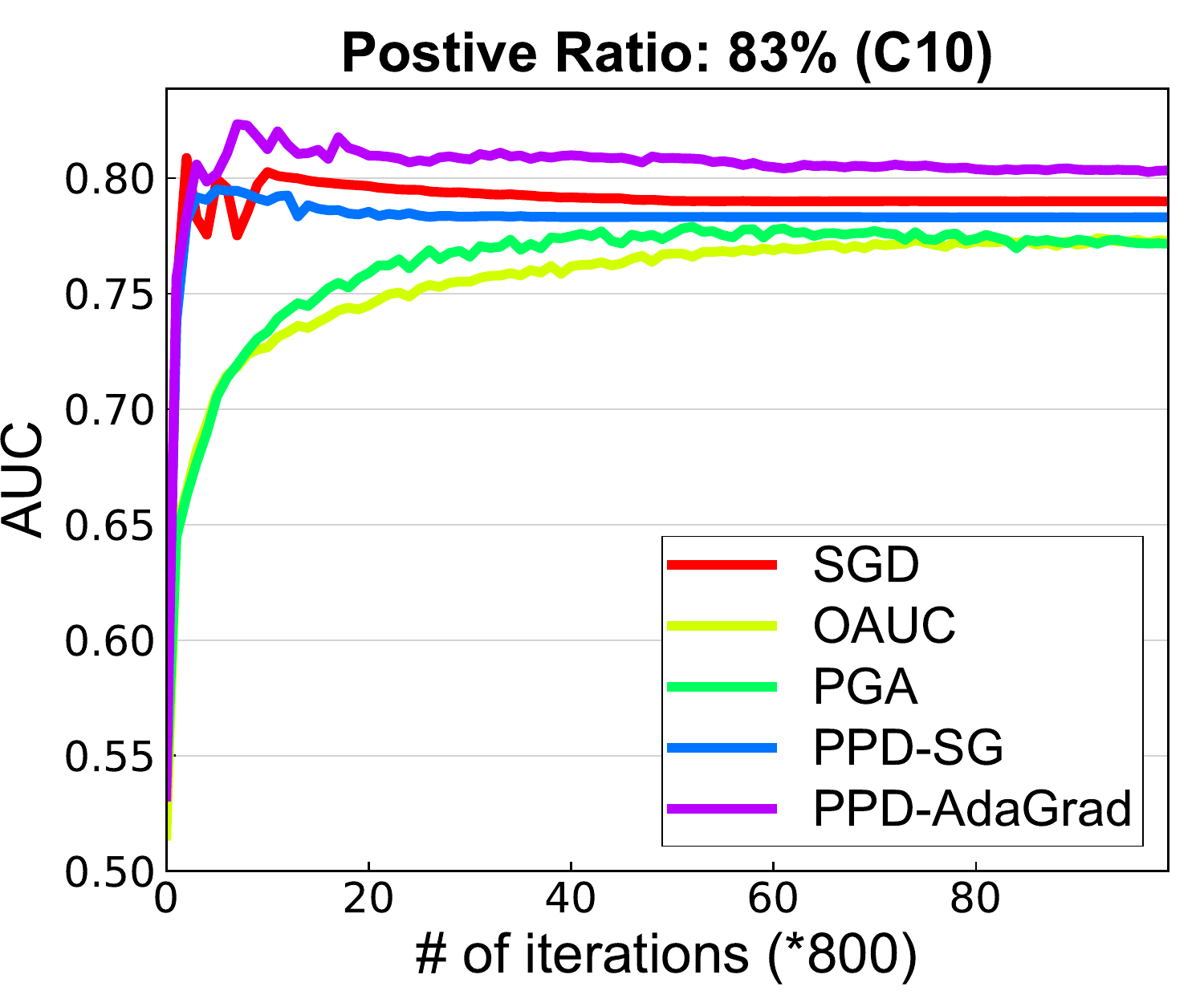}
	\hspace{10pt}\includegraphics[scale=0.19]{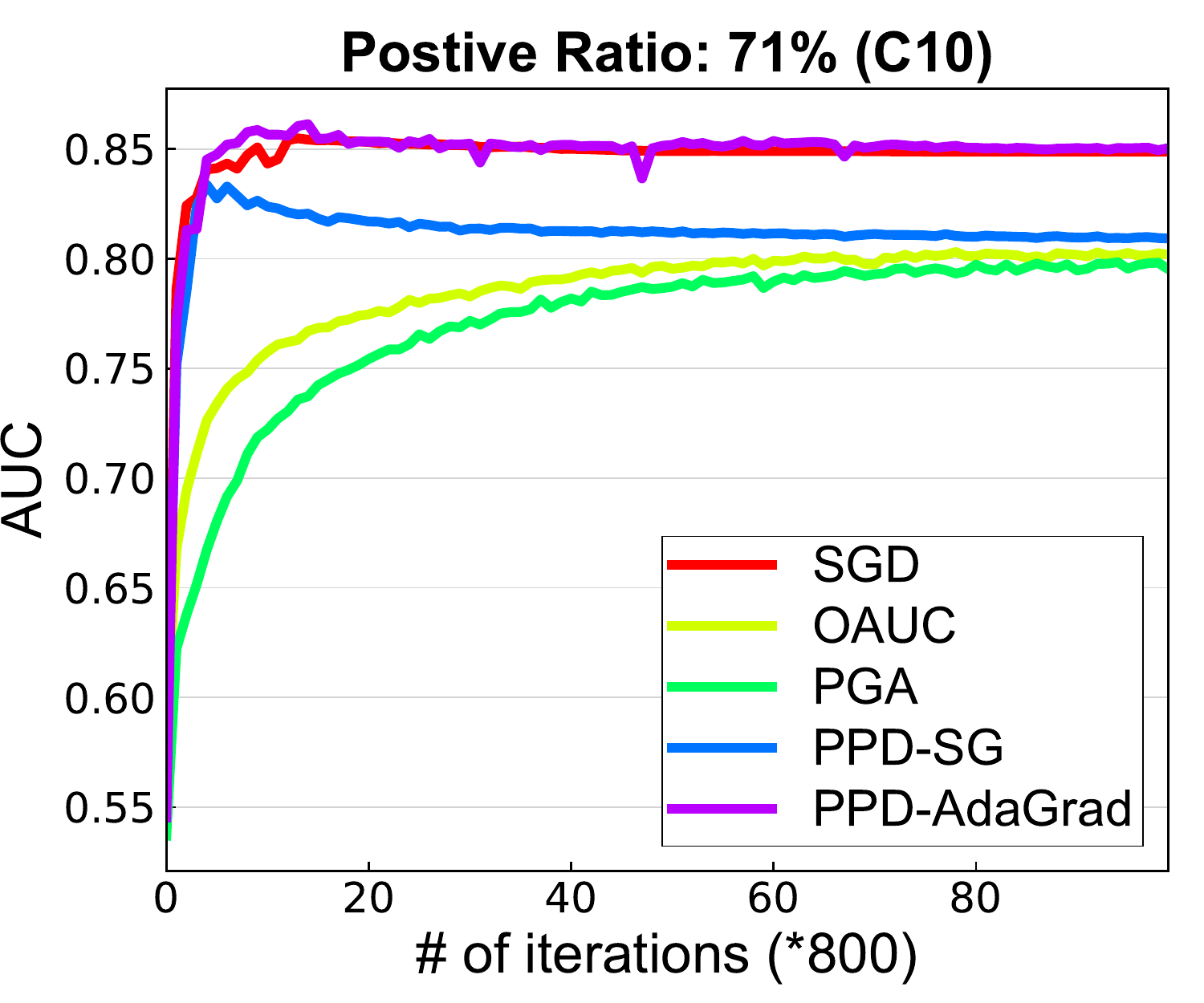}
	\hspace{10pt}\includegraphics[scale=0.19]{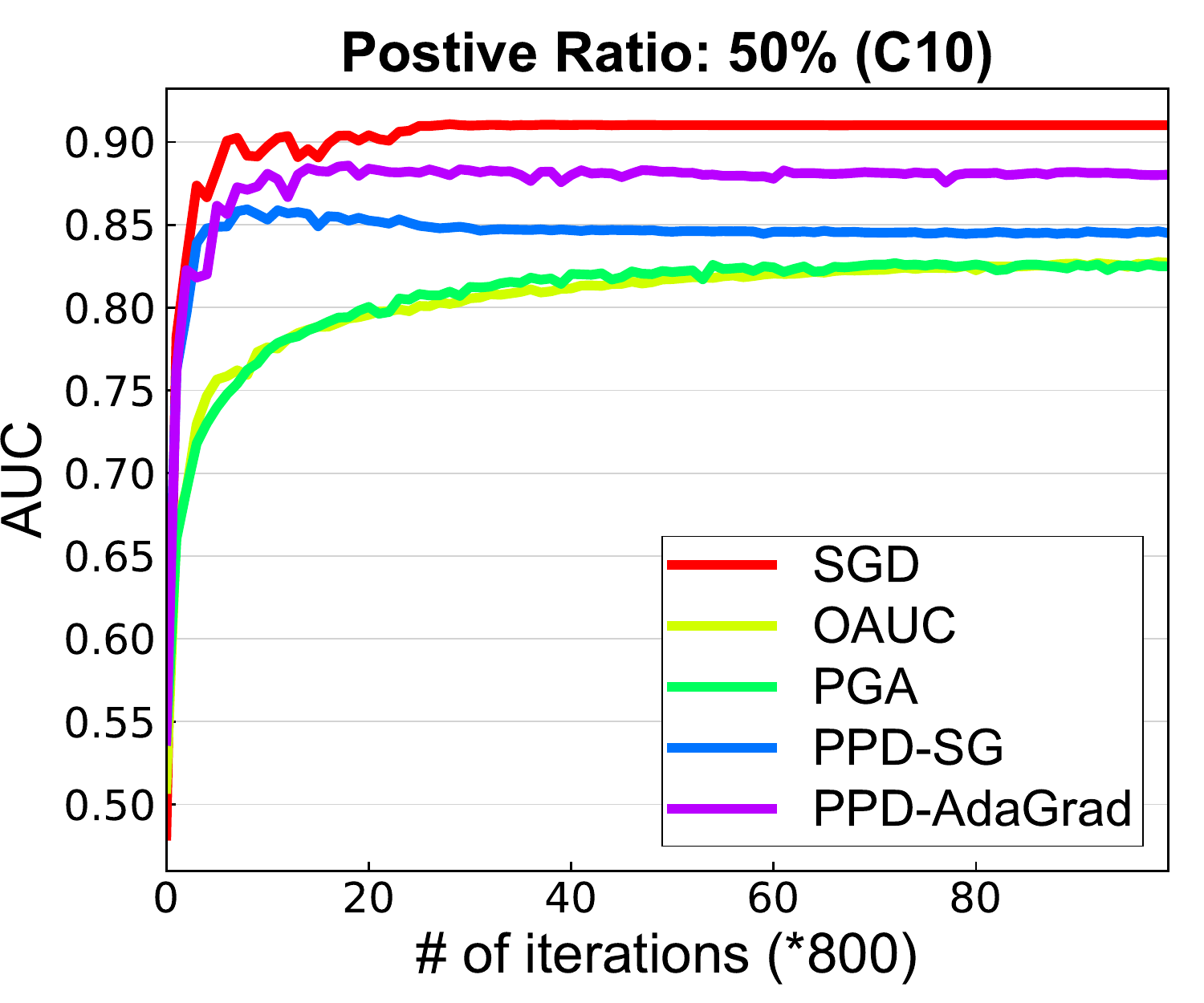}
	
	%\hspace{-10pt}
	\includegraphics[scale=0.19]{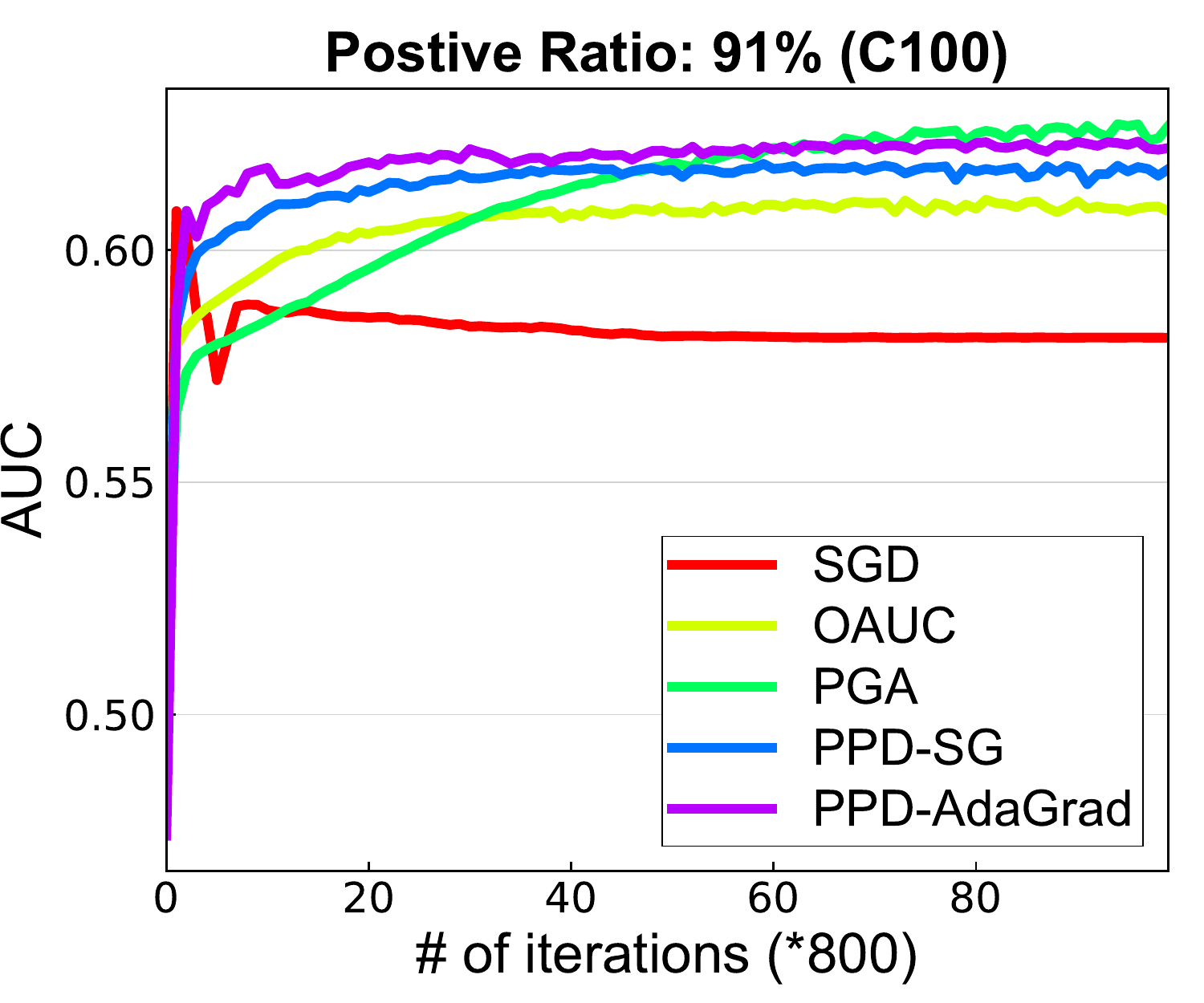}
	\hspace{10pt}\includegraphics[scale=0.19]{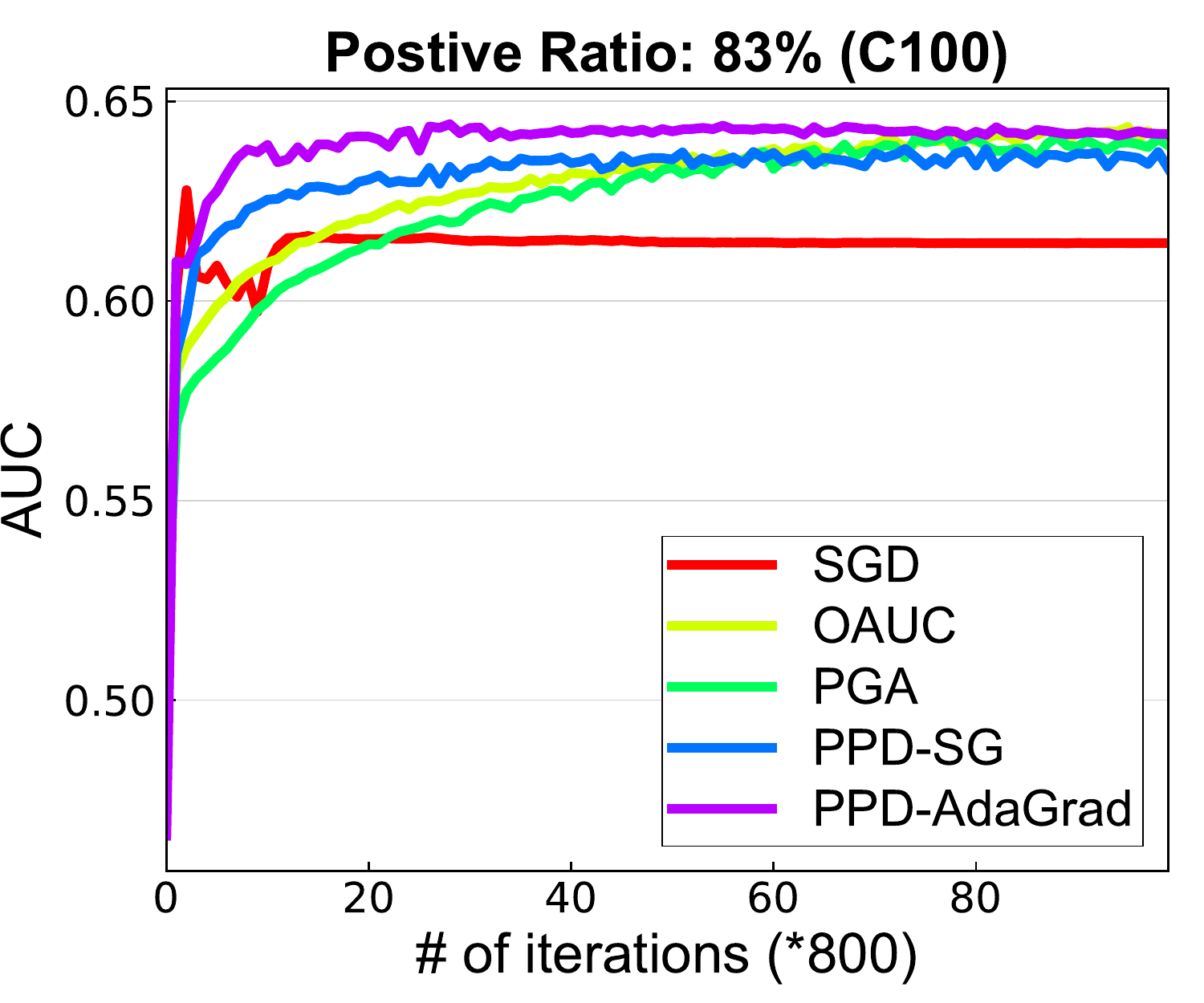}
	\hspace{10pt}\includegraphics[scale=0.19]{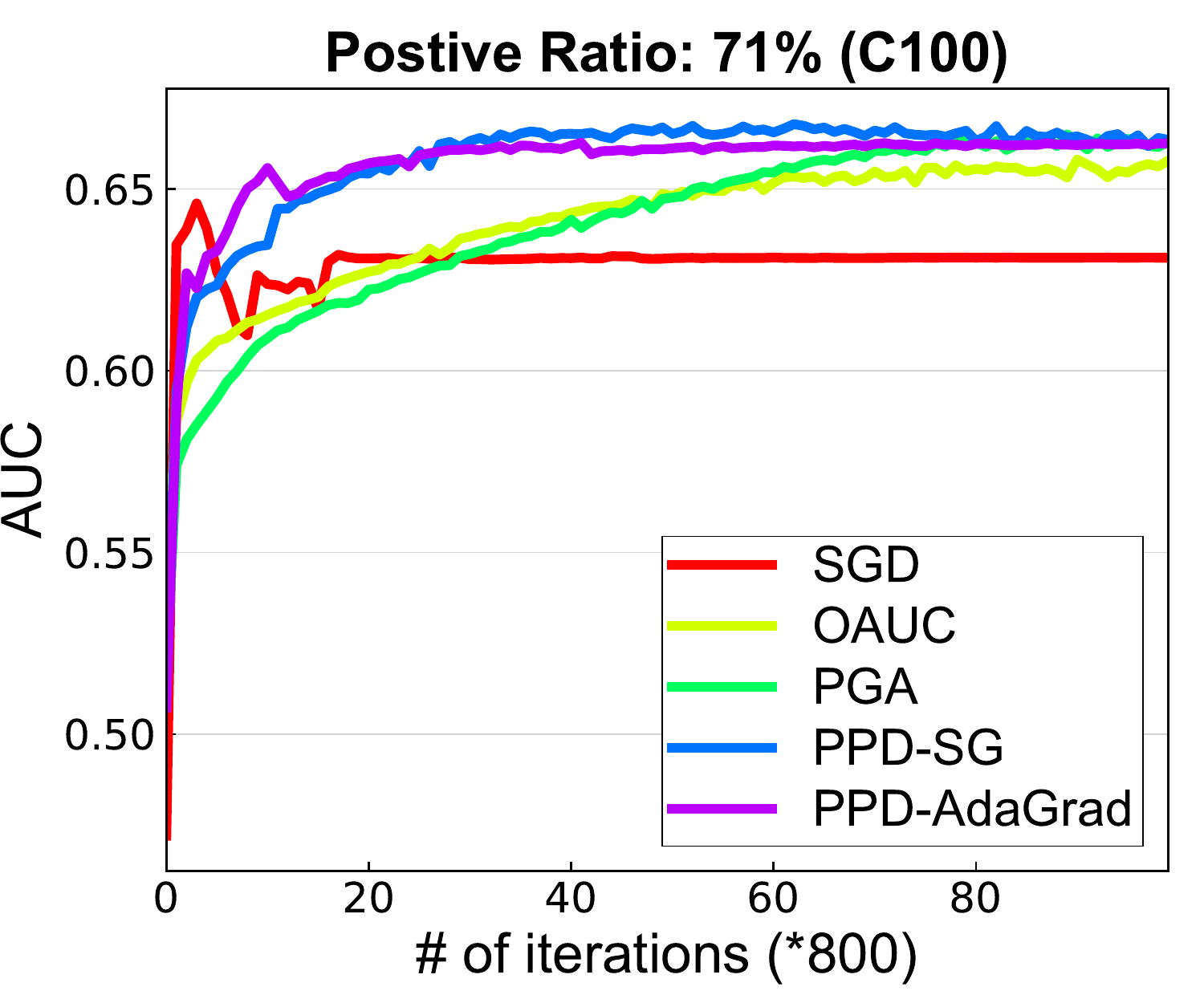}
	\hspace{10pt}\includegraphics[scale=0.19]{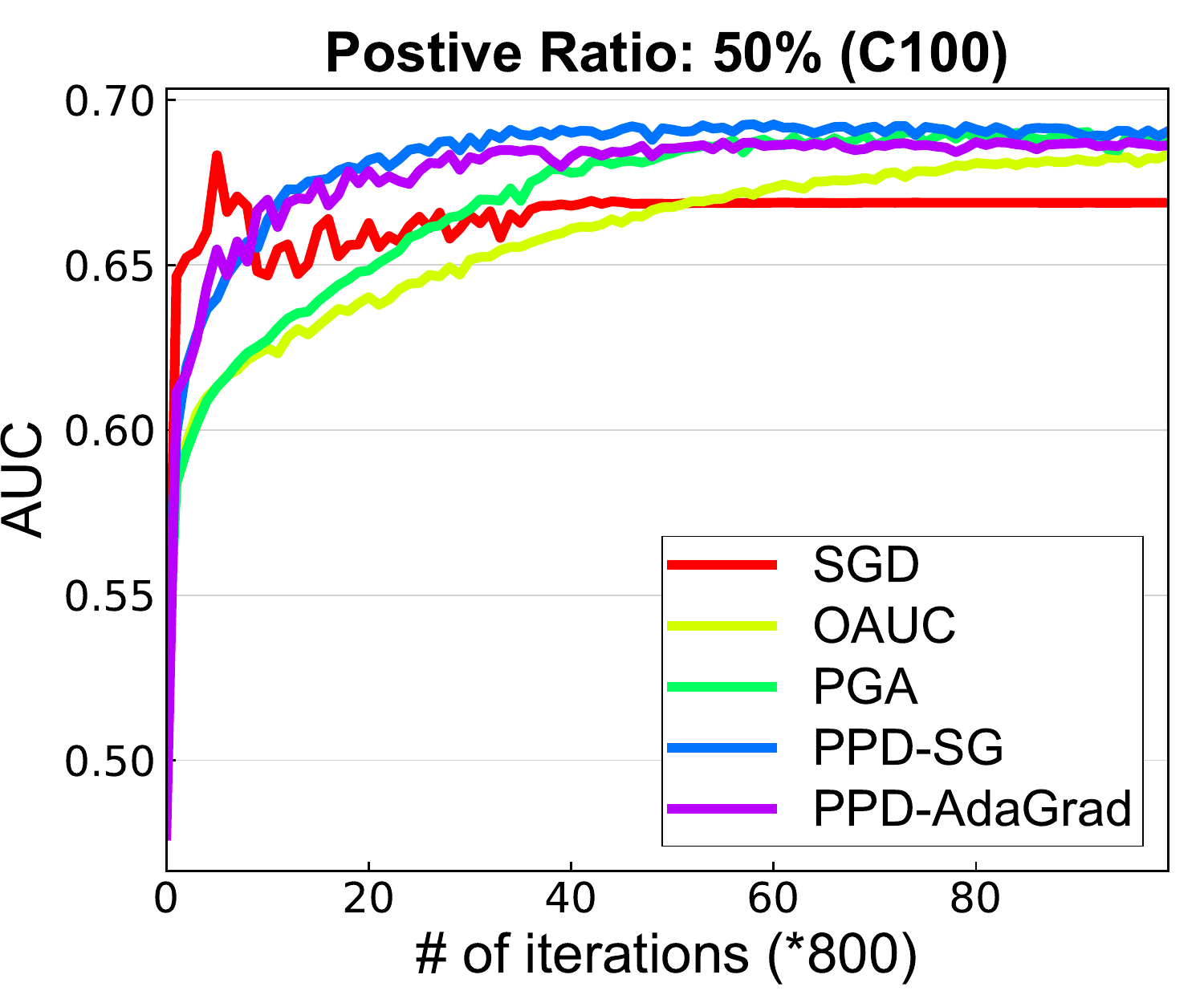}
	
	%\hspace{-10pt}
	\includegraphics[scale=0.19]{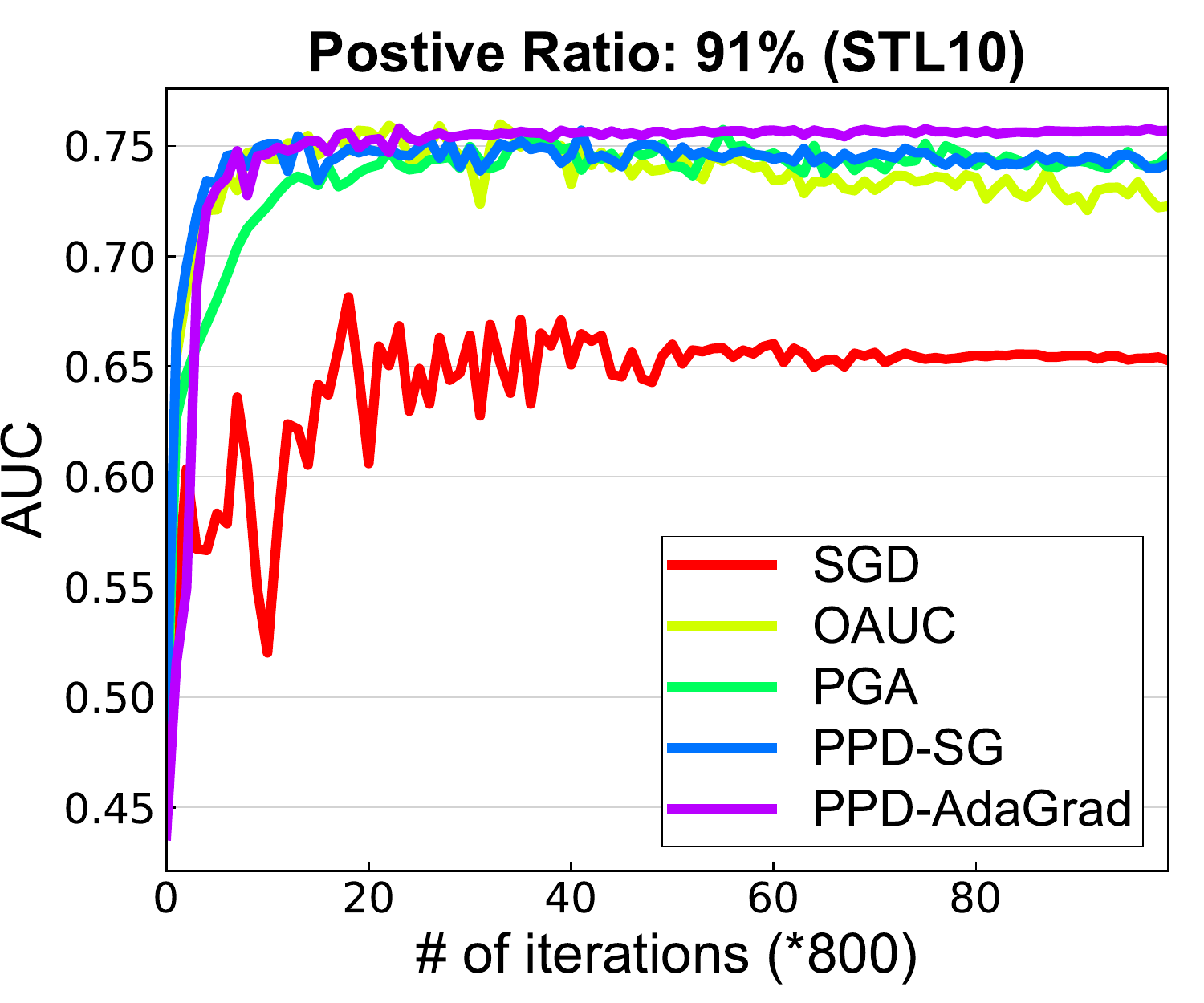}
	\hspace{10pt}\includegraphics[scale=0.19]{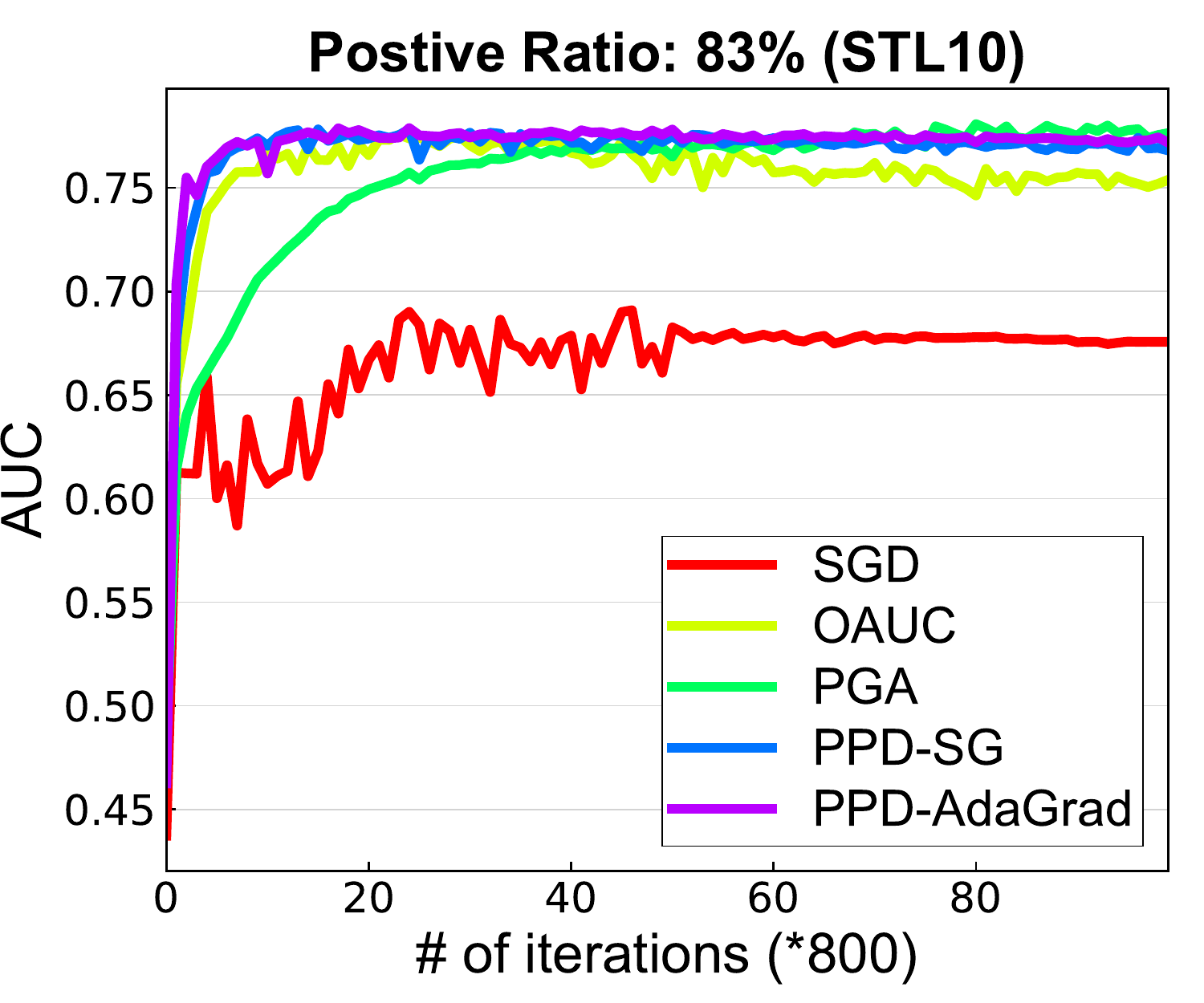}
	\hspace{10pt}\includegraphics[scale=0.19]{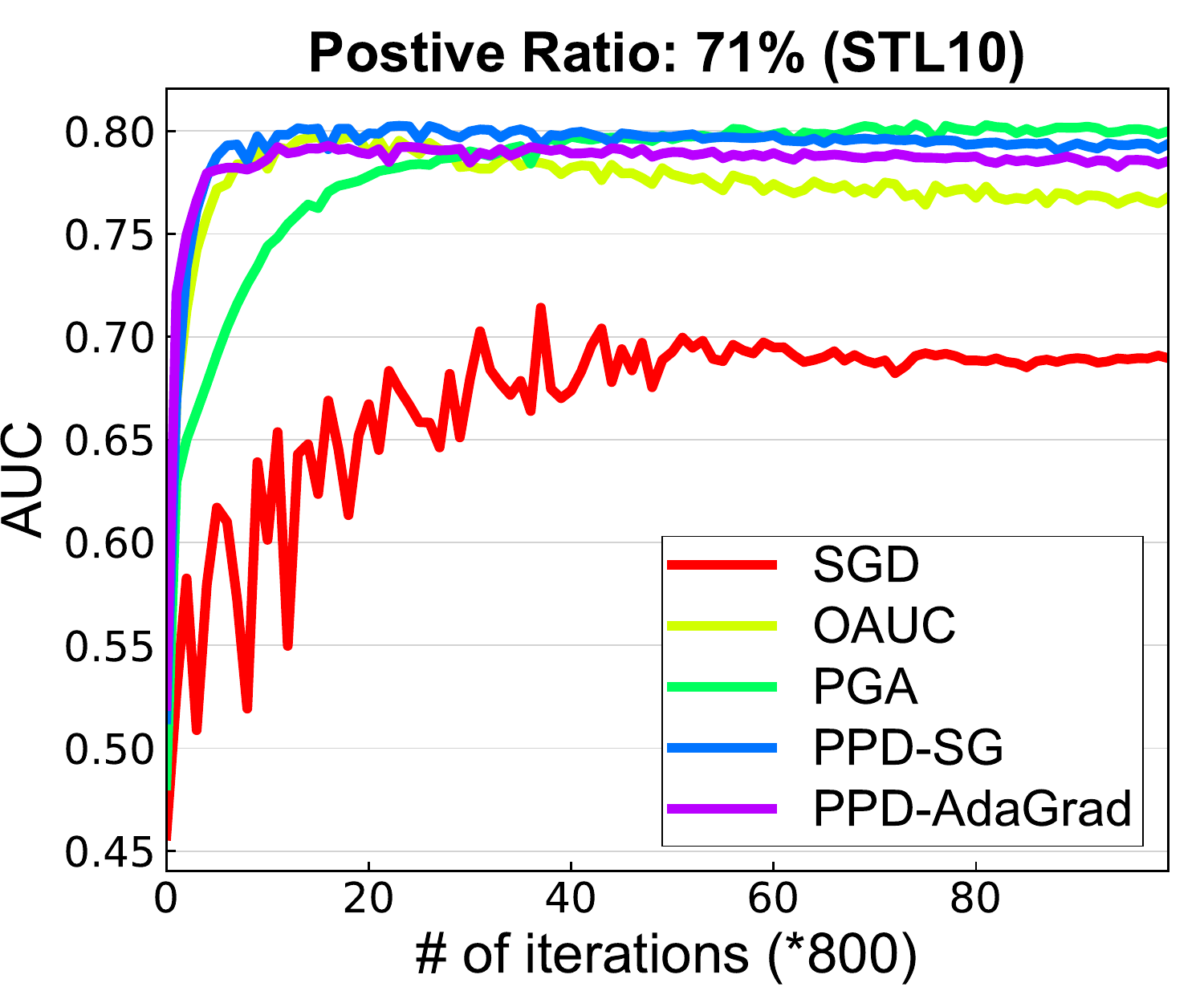}
	\hspace{10pt}\includegraphics[scale=0.19]{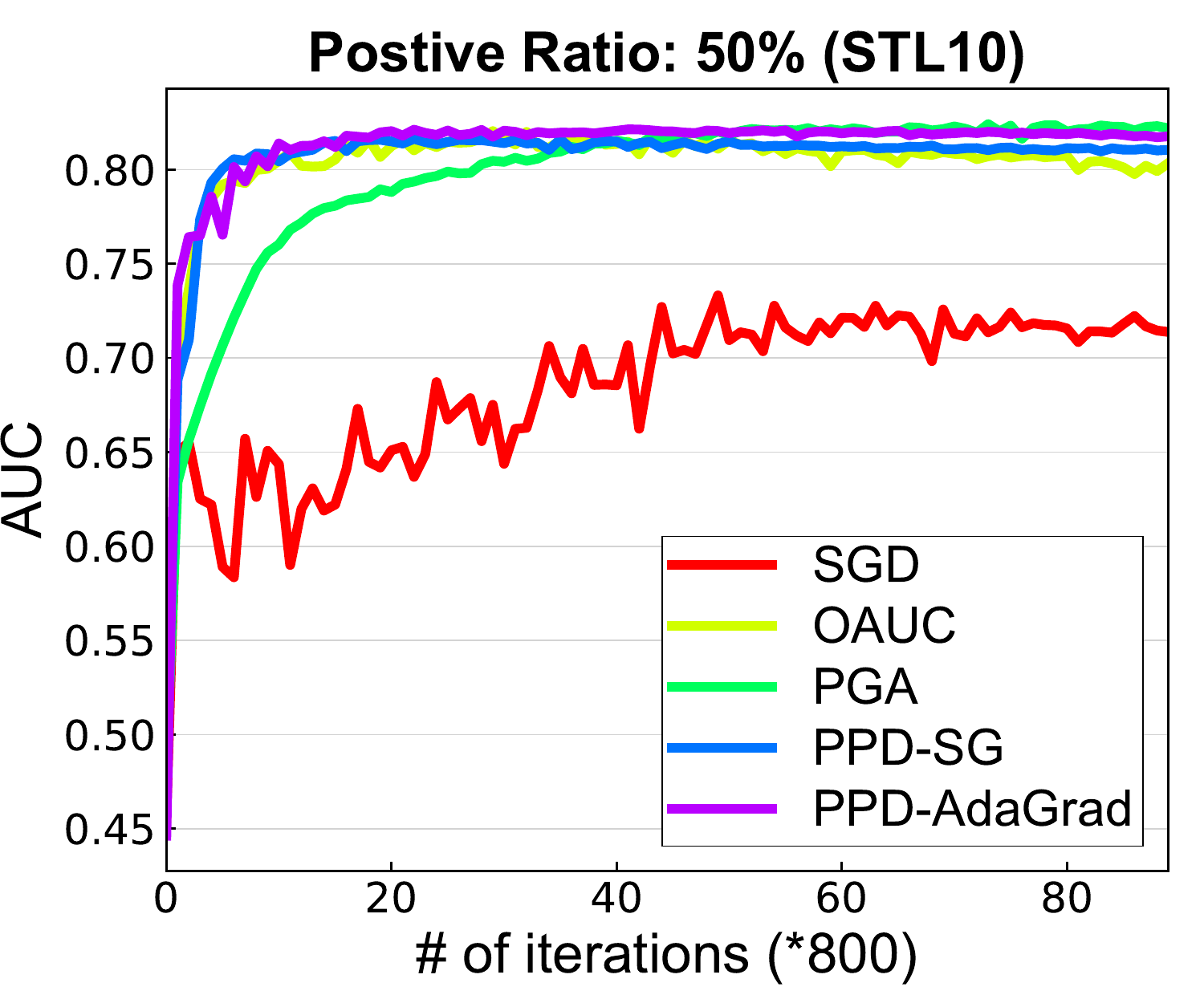}
	\vspace*{-0.08in}
	\caption{Comparison of testing AUC on  Cat\&Dog, CIFAR10, CIFAR100 and STL10.}
	\label{fig:0}
	\vspace*{0.08in}
\end{figure}

\section{Experimental Results}
\label{exp:settings}
\vspace*{-0.15in}
In this section, we present some empirical results to verify the effectiveness of the proposed algorithms. We compare our algorithms (PPD-SG and PPD-AdaGrad) with three baseline methods including PGA (Algorithm~\ref{alg:proximal-pd-rafique}), Online AUC method~\citep{ying2016stochastic} (OAUC) that directly employs the standard primal-dual stochastic gradient method with a decreasing step size for solving the min-max formulation, and the standard stochastic gradient descent (SGD)  for minimizing cross-entropy loss. Comparing with PGA and OAUC allows us to verify the effectiveness  of the proposed algorithms for solving the same formulation, and comparing with SGD allows us to verify the effectiveness of maximizing AUC for imbalanced data. We use a residual network with 20 layers (ResNet-20) to implement the deep neural network for all algorithms.
% The code is included in the .
%Code can be downloaded from \cite{code}.

We use the stagewise step size strategy as in \citep{he2016deep} for SGD, i.e. the step size is decreased by 10 times at 40K, 60K.   % for validation purpose on
%\textcolor{red}{We randomly choose 1k, 5k, 5k, 100 samples from C2, C10, C100, and STL10 respectively for . After the best hyperparameter is selected, the validation data is also used in the training.}
For PPD-SG and PPD-AdaGrad, we set $T_s = T_03^k, \eta_s = \eta_0/3^k$.  $T_0, \eta_0$ are tuned on a validation data. The value of $\gamma$ is tuned for PGA and the same value is used for PPD-SG and PPD-AdaGrad. The initial step size is tuned in [0.1, 0.05, 0.01, 0.008, 0.005] and $T_0$ is tuned in $[200 \sim 2000]$ for each algorithm separately. The batch size is set to 128. For STL10, we use a smaller batch size 32 due to the limited training data. 

We conduct the comparisons on four benchmark datasets, i.e., Cat\&Dog (C2), CIFAR10 (C10), CIFAR100 (C100), STL10. STL10 is an extension of CIFAR10 and the images are acquired from ImageNet. Cat\&Dog is from Kaggle containing 25,000 images of dogs and cats and we choose an 80:20 split to construct training and testing set. We use 19k/1k, 45k/5k, 45k/5k, 4k/1k training/validation split on C2, C10, C100, and STL10 respectively. For each dataset, we construct multiple binary classification tasks with varying imbalanced ratio of number negative examples to number of positive examples. For details of construction of binary classification tasks, please refer to the Appendix~\ref{dataset:prep}.

We report the convergence of AUC on testing data in Figure \ref{fig:0}, where the title shows the ratio of the majority class to the minority class. The results about the convergence of AUC versus the time in seconds are also presented in Figure~\ref{fig:add_exp}. From the results we can see that  for the balanced settings with ratio equal to 50\%, SGD performs consistently better than other methods on C2 and CIFAR10 data. However, it is worse than AUC optimization based methods on CIFAR100 and STL10.  For imbalanced settings, AUC maximization based methods are more advantageous than SGD in most cases. In addition, PPD-SG and PPD-AdaGrad are mostly better than other baseline algorithms. In certain cases, PPD-AdaGrad can be faster than PPD-SG. Finally,  we  observe  even better performance (in Appendix) by a mixed strategy that pre-trains the model with SGD and then switchs to PPD-SG. %In overall, v2 and v3 are comparative. ( Need to be modified; wait for new results(adagrad))
\vspace*{-0.15in}
\section{Conclusion}
\vspace*{-0.1in}
In this paper, we consider stochastic AUC maximization problem when the predictive model is a deep neural network. By building on the saddle point reformulation and exploring Polyak-\L{}ojasiewicz condition in deep learning, we have proposed two algorithms with state-of-the-art complexities for stochastic AUC maximization problem. We have also demonstrated the efficiency of our proposed algorithms on several benchmark datasets, and the experimental results indicate that our algorithms converge faster than other baselines. One may consider to extend the analysis techniques to other problems with the min-max formulation.
% We believe that our analysis technique can be extended more general problems. 
\section*{Acknowledgments}
\vspace*{-0.1in}
The authors thank the anonymous reviewers for their helpful comments. M. Liu, Z. Yuan and T. Yang are
partially supported by National Science Foundation CAREER Award 1844403.
\newpage
\bibliography{iclr2020_conference}
\bibliographystyle{iclr2020_conference}

\appendix
\section{Appendix}
\subsection{Proof of Proposition \ref{thm:spp}}
\begin{proof}
	It suffices to prove that 
	\begin{equation}\label{spp:eq1}
	\E_{\z,\z'}\left[(1-h(\w;\x)+h(\w;\x'))^2 \middle\vert y=1,y'=-1\right]=1+\frac{\min_{(a,b)\in\R^2}\max_{\alpha\in\R}\E_{\z}\left[F(\w,a,b,\alpha;\z)\right]}{p(1-p)}
	\end{equation}
	Note that 
	\begin{equation}\label{spp:eq2}
	\begin{aligned}
	\text{LHS}&=1+\E\left[h^2(\w;\x)\middle\vert y=1\right]+\E\left[h^2(\w;\x')\middle\vert y'=-1\right]-2\E\left[h(\w;\x)\middle\vert y=1\right]+2\E\left[h(\w;\x')\middle\vert y'=-1\right]\\
	&\quad\quad -2\left(\E\left[h(\w;\x)\middle\vert y=1\right]\right)\left(\E\left[h(\w;\x')\middle\vert y'=-1\right]\right)\\
	&=1+\E\left[h^2(\w;\x)\middle\vert y=1\right]-\left(\E\left[h(\w;\x)\middle\vert y=1\right]\right)^2+\E\left[h^2(\w;\x')\middle\vert y'=-1\right]-\left(\E\left[h(\w;\x')\middle\vert y'=-1\right]\right)^2\\
	&\quad\quad-2\E\left[h(\w;\x)\middle\vert y=1\right]+2\E\left[h(\w;\x')\middle\vert y'=-1\right]+\left(\E[h(\w;\x)\middle\vert y=1]-\E[h(\w;\x')\middle\vert y'=-1]\right)^2\\
	&=1+\min_{(a,b)\in\R^2}\E\left[(h(\w;\x)-a)^2\middle\vert y=1\right]+\E\left[(h(\w;\x')-b)^2\middle\vert y'=-1\right]-2\E\left[h(\w;\x)\middle\vert y=1\right]\\
	&\quad\quad+2\E\left[h(\w;\x')\middle\vert y'=-1\right]+\max_{\alpha\in\R}\left[2\alpha\left(\E\left[h(\w;\x')\middle\vert y'=-1\right]-\E\left[h(\w;\x)\middle\vert y=1\right]\right)-\alpha^2\right]\\
	&=1+\min_{(a,b)\in\R^2}\max_{\alpha\in\R}\E_{\z}\left\{\frac{1}{p}\left(h(\w;\x)-a\right)^2\mathbb{I}_{[y=1]}+\frac{1}{1-p}(h(\w;\x)-b)^2\mathbb{I}_{[y=-1]}\right.\\
	&\quad\quad \left.+2\left(1+\alpha\right)\left(\frac{1}{1-p}h(\w;\x)\mathbb{I}_{[y=-1]}-\frac{1}{p}h(\w;\x)\mathbb{I}_{[y=1]}\right)-\alpha^2\right\}\\
	&=1+\frac{\min_{(a,b)\in\R^2}\max_{\alpha\in\R}\E_{\z}\left[F(\w,a,b,\alpha;\z)\right]}{p(1-p)}=\text{RHS}.
	\end{aligned}
	\end{equation}
	Note that the optimal values of $a,b,\alpha$ are chosen as $a^*=\E\left[h(\w;\x)\middle\vert y=1\right]$, $b=\E\left[h(\w;\x')\middle\vert y'=-1\right]$, $\alpha^*=\E\left[h(\w;\x')\middle\vert y'=-1\right]-\E\left[h(\w;\x)\middle\vert y=1\right]$. 
\end{proof}

\subsection{Proof of Lemma \ref{lemma:PDSGD2}}
\begin{proof}
	Define $\alpha_{*,k}=\arg\max_{\alpha}f(\bar{\v}_k,\alpha)$, $\u=(\v^\top,\alpha)^\top\in\R^{d+3}$, $\u_{*,k}=(\v_*^\top, \alpha_{*,k})^\top$, $\u_t^k=((\v_t^k)^\top,\alpha_t^k)^\top$, $\g_t^k =\left(\nabla_{\v}f(\v_{t}^k,\alpha_{t}^k)^\top+\frac{1}{\gamma}\left(\v_{t}^k-\v_0^k\right)^\top,-\nabla_{\alpha} f(\v_{t}^k,\alpha_{t}^k)^\top\right)^\top$.
	\begin{align*}
	\phi_k(\bar{\v}_k)-\min_{\v}\phi_k(\v)&\stackrel{(a)}{=}\max_{\alpha}\left[f(\bar{\v}_k,\alpha)+\frac{1}{2\gamma}\|\bar{\v}_k-\bar{\v}_{k-1}\|^2\right]-\min_{\v}\max_{\alpha}\left[f(\v,\alpha)+\frac{1}{2\gamma}\|\v-\bar{\v}_{k-1}\|^2\right]\\
	&\stackrel{(b)}{\leq} \left[f(\bar{\v}_k,\alpha_{*,k})+\frac{1}{2\gamma}\|\bar{\v}_k-\bar{\v}_{k-1}\|^2\right]-\left[f(\s_k,\bar{\alpha}_k)+\frac{1}{2\gamma}\|\s_k-\bar{\v}_{k-1}\|^2\right]\\
	&\stackrel{(c)}\leq \frac{\|\bar{\v}_{k-1}-\s_k\|^2}{2\eta_k T_k}+\frac{\|\bar{\alpha}_{k-1}-\alpha_{*,k}\|^2}{2\eta_k T_k}+\eta_k G^2 + \frac{\sum_{t=0}^{T_k-1}(\u_t^k-\u_{*,k})^\top (\g_t^k-\hat{\g}_t^k)}{T_k},	
	\end{align*}
	where (a) comes from the definition of $\phi_k$, (b) holds because $\min_{\v}\max_{\alpha}\left[f(\v,\alpha)+\frac{1}{2\gamma}\|\v-\bar{\v}_{k-1}\|^2\right]\geq f(\s_k,\bar{\alpha}_k)+\frac{1}{2\gamma}\|\s_k-\bar{\v}_{k-1}\|^2 $, (c) comes from the standard analysis of primal-dual stochastic gradient method.

	Denote $\E_{k-1}$ by taking the conditional expectation conditioning on all the stochastic events until $\bar{\v}_{k-1}$ is generated.
	Taking $\E_{k-1}$ on both sides and noting that $\hat{\g}_t^k$ is an unbiased estimator of $\g_t^k$ for $\forall t,k$, we have
	\begin{align*}
	\E_{k-1}\left[\phi_k(\bar{\v}_k)-\min_{\v}\phi_k(\v)\right]\leq  \frac{\|\bar{\v}_{k-1}-\s_k\|^2}{2\eta_k T_k}+\frac{\E_{k-1}\|\bar{\alpha}_{k-1}-\alpha_{*,k}\|^2}{2\eta_k T_k}+\eta_k G^2+\mathbf{I},
	\end{align*} 
	where $$\mathbf{I}=\E_{k-1}\left[\frac{\sum_{t=0}^{T_k-1}(\alpha_t^k-\alpha_{*,k}) \left(-\nabla_{\alpha}f(\v_t^k,\alpha_t^k)-(-\nabla_{\alpha}F(\v_{t}^k,\alpha_t^k;\xi_t^k))\right)}{T_k}\right].$$
	Define  $\widetilde{\alpha}_{0}^{k} =\alpha_0^k$ and  $$\widetilde{\alpha}_{t+1}^k=\arg\min_{\alpha}\left(-\nabla_{\alpha}f(\v_t^k,\alpha_t^k)-(-\nabla_{\alpha}F(\v_{t}^k,\alpha_t^k;\xi_t^k))\right)\alpha+\frac{1}{2\eta_k}(\alpha-\widetilde{\alpha}_{t}^k).$$  
	By first-order optimality condition, we have
	\begin{equation}
	\label{eq:new1}
	\begin{aligned}
			&\left(\widetilde{\alpha}_t^k-\alpha_{*,k}\right)\left(-\nabla_{\alpha}f(\v_t^k,\alpha_t^k)-(-\nabla_{\alpha}F(\v_{t}^k,\alpha_t^k;\xi_t^k))\right)\\
			&\leq \frac{(\widetilde{\alpha}_{t}^k-\alpha_{*,k})^2-(\widetilde{\alpha}_{t+1}^k-\alpha_{*,k})^2}{2\eta_k}+\frac{\eta_k}{2}\left(-\nabla_{\alpha}f(\v_t^k,\alpha_t^k)-(-\nabla_{\alpha}F(\v_{t}^k,\alpha_t^k;\xi_t^k))\right)^2
	\end{aligned}
\end{equation}
Note that 
\begin{equation*}
	\begin{aligned}
	\mathbf{I} &= \E_{k-1}\left[\frac{\sum_{t=0}^{T_k-1}(\alpha_t^k-\widetilde{\alpha}_t^k+\widetilde{\alpha}_t^k-\alpha_{*,k}) \left(-\nabla_{\alpha}f(\v_t^k,\alpha_t^k)-(-\nabla_{\alpha}F(\v_{t}^k,\alpha_t^k;\xi_t^k))\right)}{T_k}\right]\\
	&= \E_{k-1}\left[\frac{\sum_{t=0}^{T_k-1}(\widetilde{\alpha}_t^k-\alpha_{*,k}) \left(-\nabla_{\alpha}f(\v_t^k,\alpha_t^k)-(-\nabla_{\alpha}F(\v_{t}^k,\alpha_t^k;\xi_t^k))\right)}{T_k}\right]\\
	&\leq \frac{\E_{k-1}(\widetilde{\alpha}_0^k-\alpha_{*,k})^2}{2\eta_k T_k}+\eta_k G^2=\frac{\E_{k-1}(\bar{\alpha}_{t-1}-\alpha_{*,k})^2}{2\eta_k T_k}+\eta_k G^2
	\end{aligned}
\end{equation*}
where the first inequality holds due to~(\ref{eq:new1}). Hence we have
\begin{equation*}
	\E_{k-1}\left[\phi_k(\bar{\v}_k)-\min_{\v}\phi_k(\v)\right]\leq  \frac{\|\bar{\v}_{k-1}-\s_k\|^2}{2\eta_k T_k}+\frac{\E_{k-1}\|\bar{\alpha}_{k-1}-\alpha_{*,k}\|^2}{\eta_k T_k}+2\eta_k G^2.
\end{equation*}

Define $\x_{j:j+m_{k-1}-1}=(\x_{j},\ldots,\x_{j+m_{k-1}-1})$,  $y_{j:j+m_{k-1}-1}=(y_{j},\ldots,y_{j+m_{k-1}-1})$, and $\tilde{f}(\x_{j:j+m_{k-1}-1},y_{j:j+m_{k-1}-1})=\frac{\sum_{i=j}^{j+m_{k-1}-1}h(\bar{\w}_{k-1};\x_i)\mathbb{I}_{y_i=y}}{\sum_{i=j}^{j+m_{k-1}-1}\mathbb{I}_{y_i=y}}-\E_{\x}[h(\bar\w_{k-1}; \x)|y]$. Note that $0\leq h \leq 1$. Then we know that
\begin{align*}
&\E_{\x_{j:j+m_{k-1}-1}}(\tilde{f}^2(\x_{j:j+m_{k-1}-1},y_{j:j+m_{k-1}-1})\vert y_{j:j+m_{k-1}-1})\\
&\leq \frac{\sigma^2}{\sum_{i=j}^{j+m_{k-1}-1}\mathbb{I}_{y_i=y}}\cdot\mathbb{I}_{\left(\sum_{i=j}^{j+m_{k-1}-1}\mathbb{I}_{y_i=y}>0\right)}+1\cdot\mathbb{I}_{\left(\sum_{i=j}^{j+m_{k-1}-1}\mathbb{I}_{y_i=y}=0\right)}.
\end{align*}

Hence
\begin{equation*}
	\begin{aligned}
	&\E_{k-1}\left[\tilde{f}^2(\x_j,\ldots,\x_{j+m_{k-1}-1},y_j,\ldots,y_{j+m_{k-1}-1})\right]\\
	&=\E_{y_{j:j+m_{k-1}-1}}\left[\E_{\x_{j:j+m_{k-1}-1}}(\tilde{f}^2(\x_{j:j+m_{k-1}-1},y_{j:j+m_{k-1}-1})\vert y_{j:j+m_{k-1}-1}\right]\\
	&\leq \E_{y_{j:j+m_{k-1}-1}}\left[\frac{\sigma^2}{\sum_{i=j}^{j+m_{k-1}-1}\mathbb{I}_{y_i=y}}\cdot\mathbb{I}_{\left(\sum_{i=j}^{j+m_{k-1}-1}\mathbb{I}_{y_i=y}>0\right)}+1\cdot\mathbb{I}_{\left(\sum_{i=j}^{j+m_{k-1}-1}\mathbb{I}_{y_i=y}=0\right)}\right]\\
	&\leq \frac{\sigma^2}{m_{k-1}\text{Pr}(y_i=y)}+(1-\text{Pr}\left(y_i=y\right))^{m_{k-1}}.\\
	%&\leq \E_{\x}(f^2(y)|y=1)+\E_{\x}(f^2(y)|y=-1)\\
	%&\leq \frac{\sigma^2}{\sum_{i=j}^{j+m_k-1}\mathbb{I}_{y_i=y}}\cdot\text{Pr}\left(\sum_{i=j}^{j+m_k-1}\mathbb{I}_{y_i=-1}>0\right)+1\cdot\text{Pr}\left(\sum_{i=j}^{j+m_k-1}\mathbb{I}_{y_i=-1}\right)+\frac{\sigma^2}{\sum_{i=j}^{j+m_k-1}\mathbb{I}_{y_i=1}}\cdot\text{Pr}\left(\sum_{i=j}^{j+m_k-1}\mathbb{I}_{y_i=y}>0\right)+1\cdot\text{Pr}\left(\sum_{i=j}^{j+m_k-1}\mathbb{I}_{y_i=1}\right)
	\end{aligned}
\end{equation*}
%Define $\delta(y)=\text{Pr}\left(\sum_{i=j}^{j+m_k-1}\mathbb{I}_{y_i=y}\right)$, then we have
%$$\E_{\x}(f^2(y)\vert y)\leq \frac{\sigma^2}{\sum_{i=j}^{j+m_k-1}\mathbb{I}_{y_i=y}}\cdot(1-\delta(y))+\delta(y)$$

Hence we have
		\begin{equation*}
\begin{aligned}
&\E_{k-1}\|\bar{\alpha}_{k-1}-\alpha_{*,k-1}\|^2\\
&=\E_{k-1}\left[\frac{\sum_{i=j}^{j+m_{k-1}-1}h(\bar{\w}_{k-1};\x_i)\mathbb{I}_{y_i=-1}}{\sum_{i=j}^{j+m_{k-1}-1}\mathbb{I}_{y_i=-1}}- \E_{\x}[h(\bar\w_{k-1}; \x)|y=-1]\right.\\&\qquad\left.+\E_\x[h(\bar\w_{k-1}; \x)|y=1]-\frac{\sum_{i=j}^{j+m_k-1}h(\bar{\w}_{k-1};\x_i)\mathbb{I}_{y_i=1}}{\sum_{i=j}^{j+m_{k-1}-1}\mathbb{I}_{y_i=1}}\right]^2\\
&\leq  \frac{2\sigma^2}{m_{k-1}\text{Pr}(y_i=-1)}+2(1-\text{Pr}\left(y_i=-1\right))^{m_{k-1}}+ \frac{2\sigma^2}{m_{k-1}\text{Pr}(y_i=1)}+2(1-\text{Pr}\left(y_i=1\right))^{m_{k-1}}\\
	&=\frac{2\sigma^2}{m_{k-1} p(1-p)}+2p^{m_{k-1}}+2(1-p)^{m_{k-1}} \leq 2\left(\frac{\sigma^2}{m_{k-1} p(1-p)}+2(\max(p,1-p))^{m_{k-1}}\right)\\
	&\stackrel{(a)}{\leq} 2\left(\frac{\sigma^2}{m_{k-1}p(1-p)}+\frac{C}{m_{k-1}}\right)\leq \frac{2(\sigma^2+C)}{m_{k-1} p(1-p)}.
\end{aligned}
\end{equation*}
where $C=\frac{2}{\ln(\frac{1}{\max(p,1-p)})}\max(p,1-p)^{\frac{1}{\ln(1/\max(p,1-p))}}$, and (a) holds since the function $x\max(p,1-p)^x$ achieves its maximum at point $x=1/\ln(1/\max(p, 1-p))$.

	By the update of $\bar{\alpha}_{k-1}$, $2\tilde{L}$-Lipschitz continuity of $\E \left[h(\w;\x)\middle\vert y=-1\right]-\E \left[h(\w;\x)\middle\vert y=1\right]$, and noting that $\alpha_{*,k}=\E\left[h(\bar{\w}_k;\x)\middle\vert y=-1\right]-\E\left[h(\bar{\w}_k;\x)\middle\vert y=1\right]$, we have
	%and the fact that $\|\a+\b+\c\|^2\leq 3\|\a\|^2+3\|\b\|^2+3\|\c\|^2$, we have
	\begin{align*}
	&\E_{k-1}\|\bar{\alpha}_{k-1}-\alpha_{*,k}\|^2=\E_{k-1}\left\|\bar{\alpha}_{k-1}-\alpha_{*,k-1}+\alpha_{*,k-1}-\alpha_{*,k}\right\|^2\\
	&\leq \E_{k-1} \left(2\|\bar{\alpha}_{k-1}-\alpha_{*,k-1}\|^2+2\|\alpha_{*,k-1}-\alpha_{*,k}\|^2\right)\leq \frac{4(\sigma^2+C)}{m_{k-1} p(1-p)}+8\tilde{L}^2\E_{k-1}\|\bar{\v}_{k-1}-\bar{\v}_{k}\|^2. %3\tilde{L}^2\|\bar{\v}_{k-1}-\s_k\|^2+\frac{3\sigma^2}{m},
	\end{align*}

	%where $m$ is the minibatch size, and $\sigma^2=$

	%Hence, we have
	%\begin{align*}
	%\E_{k-1}\left[\phi_k(\bar{\v}_k)-\min_{\v}\phi_k(\v)\right]\leq  \frac{\|\bar{\v}_{k-1}-\s_k\|^2}{2\eta_k T_k}+\frac{\tilde{L}^2\|\bar{\v}_{k-1}-\s_k\|^2+\frac{\sigma^2}{m}}{2\eta_k T_k}+\eta_k G^2.
	%\end{align*}
	%Conditioning on all stochastic events until $\bar{\v}_{k-1}$ is generated, 
	
	Taking $m_{k-1}\geq \frac{2(\sigma^2+C)}{p(1-p)\eta_k^2 G^2T_k}$, then we have 
	\begin{align*}
	\E_{k-1}\left[\phi_k(\bar{\v}_k)-\min_{\v}\phi_k(\v)\right]\leq  \frac{\|\bar{\v}_{k-1}-\s_k\|^2+16\tilde{L}^2\E_{k-1}\|\bar{\v}_{k-1}-\bar{\v}_k\|^2}{2\eta_k T_k}+4\eta_k G^2.
	\end{align*}
\end{proof}

\subsection{Proof of Theorem \ref{theorem:sgd}}
\begin{proof}
	Define $\phi_k(\v)=\phi(\v)+\frac{1}{2\gamma}\|\v-\bar{\v}_{k-1}\|^2$. We can see that $\phi_k(\v)$ is convex and smooth function since $\gamma\leq 1/L$. The smoothness parameter of $\phi_k$ is $\hat{L}=L+\gamma^{-1}$. Define $\s_k=\arg\min_{\v\in\R^{d+2}}\phi_k(\v)$. According to Theorem 2.1.5 of~\citep{nesterov2013introductory}, we have
	\begin{align}
	\label{eq:b3}
	\|\nabla\phi_k(\bar{\v}_k)\|^2\leq 2\hat{L}(\phi_k(\bar{\v}_k)-\phi_k(\s_k)).
	\end{align}
	%123
	Combining (\ref{eq:b3}) with Lemma \ref{lemma:PDSGD2} yields
	\begin{align}
	\label{eq:b2}
	\E_{k-1}\|\nabla\phi_k(\bar{\v}_k)\|^2\leq 2\hat{L}\left(\frac{\|\bar{\v}_{k-1}-\s_k\|^2+16\tilde{L}^2\E_{k-1}\|\bar{\v}_{k-1}-\bar{\v}_k\|^2}{2\eta_k T_k}+4\eta_k G^2\right).
	\end{align}
	Note that $\phi_k(\bar{\v})$ is $(\gamma^{-1} - L)$-strongly convex, and $\gamma=\frac{1}{2L}$, we have 
	\begin{equation}
	\label{eq:sc}
	\phi_k(\bar{\v}_{k-1})\geq \phi_k(\s_k)+\frac{L}{2}\|\bar{\v}_{k-1}-\s_k\|^2.
	\end{equation}
	
	Plugging in $\s_k$ into Lemma \ref{lemma:PDSGD2} and combining (\ref{eq:sc}) yield
	\begin{align*}
	&\E_{k-1}[\phi(\bar{\v}_k)+L\|\bar{\v}_k-\bar{\v}_{k-1}\|^2]\\
	&\leq \phi_k(\bar{\v}_{k-1})-\frac{L}{2}\|\bar{\v}_{k-1}-\s_k\|^2+\frac{\|\bar{\v}_{k-1}-\s_k\|^2+16\tilde{L}^2\E_{k-1}\|\bar{\v}_{k-1}-\bar{\v}_k\|^2}{2\eta_k T_k}+4\eta_kG^2.
	\end{align*}
	By using $\eta_kT_k L\geq \max(2,16\tilde{L}^2)$, rearranging the terms, and noting that $\phi_k(\bar{\v}_{k-1})=\phi(\bar{\v}_{k-1})$, we have
	\begin{equation}
	\label{eq:b1}
	\frac{\|\bar{\v}_{k-1}-\s_k\|^2+16\tilde{L}^2\E_{k-1}\|\bar{\v}_{k-1}-\bar{\v}_k\|^2}{2\eta_k T_k}\leq \phi(\bar{\v}_{k-1})-	\E_{k-1}\left[\phi(\bar{\v}_k)\right] +4\eta_k G^2.
	\end{equation}
	Combining (\ref{eq:b1}) and (\ref{eq:b2}) yields
	\begin{equation}
	%\label{eq:thm:2}
	\E_{k-1}\|\nabla\phi_k(\bar{\v}_k)\|^2\leq 6L\left(\phi(\bar{\v}_{k-1})-	\E_{k-1}\left[\phi(\bar{\v}_k)\right] +8\eta_k G^2\right).
	\end{equation}
	Taking expectation on both sides over all randomness until $\bar{\v}_{k-1}$ is generated and by the tower property, we have
	\begin{equation}
	\label{eq:thm:2}
	\E\|\nabla\phi_k(\bar{\v}_k)\|^2\leq 6L\left(\E\left[\phi(\bar{\v}_{k-1})-\phi(\v_*)\right]-	\E\left[\phi(\bar{\v}_k)-\phi(\v_*)\right] +8\eta_k G^2\right).
	\end{equation}
	Note that $\phi(\v)$ is $L$-smooth and hence is $L$-weakly convex, so we have 
	\begin{equation}
	\label{eq:thm:1}
	\begin{aligned}
	&\phi(\bar{\v}_{k-1})\geq\phi(\bar{\v}_{k})+ \left\langle\nabla\phi(\bar{\v}_k),\bar{\v}_{k-1}-\bar{\v}_k\right\rangle-\frac{L}{2}\|\bar{\v}_{k-1}-\bar{\v}_k\|^2\\
	&=\phi(\bar{\v}_k)+\left\langle\nabla\phi(\bar{\v}_k)+2L(\bar{\v}_k-\bar{\v}_{k-1}), \bar{\v}_{k-1}-\bar{\v}_k\right\rangle+\frac{3}{2}L\|\bar{\v}_{k-1}-\bar{\v}_{k}\|^2\\
	&\stackrel{(a)}{=}\phi(\bar{\v}_k)+\left\langle\nabla\phi_k(\bar{\v}_k),\bar{\v}_{k-1}-\bar{\v}_k\right\rangle
	+\frac{3}{2}L\|\bar{\v}_{k-1}-\bar{\v}_{k}\|^2\\
	&\stackrel{(b)}{=}\phi(\bar{\v}_k)-\frac{1}{2L}\left\langle \nabla\phi_k(\bar{\v}_k),\nabla\phi_k(\bar{\v}_k)-\nabla\phi(\bar{\v}_k)\right\rangle+\frac{3}{8L}\|\nabla\phi_k(\bar{\v}_k)-\nabla\phi(\bar{\v}_k)\|^2\\
	%&= \phi(\bar{\v}_k)-\frac{1}{2L}\|\nabla\phi_k(\bar{\v}_k)\|^2+\frac{1}{2L}\left\langle\nabla\phi_k(\bar{\v}_k),\nabla\phi(\bar{\v}_k)\right\rangle+\frac{3}{8L}\|\nabla\phi_k(\bar{\v}_k)\|^2-\frac{3}{4L}\left\langle\nabla\phi_k(\bar{\v}_k),\nabla\phi(\bar{\v}_k)\right\rangle\\
	&=\phi(\bar{\v}_k)-\frac{1}{8L}\|\nabla\phi_k(\bar{\v}_k)\|^2-\frac{1}{4L}\left\langle\nabla\phi_k(\bar{\v}_k),\nabla\phi(\bar{\v}_k)\right\rangle+\frac{3}{8L}\|\nabla\phi(\bar{\v}_k)\|^2,
	\end{aligned}
	\end{equation}
	where (a) and (b) hold by the definition of $\phi_k$. 
	
	Rearranging the terms in (\ref{eq:thm:1}) yields
	\begin{equation}
	\label{thm:eq:3}
	\begin{aligned}
	\phi(\bar{\v}_k)-\phi(\bar{\v}_{k-1})&\leq \frac{1}{8L}\|\nabla\phi_k(\bar{\v}_k)\|^2+\frac{1}{4L}\left\langle\nabla\phi_k(\bar{\v}_k),\nabla\phi(\bar{\v}_k)\right\rangle-\frac{3}{8L}\|\nabla\phi(\bar{\v}_k)\|^2\\
	&\stackrel{(a)}{\leq} \frac{1}{8L}\|\nabla\phi_k(\bar{\v}_k)\|^2+\frac{1}{8L}\left(\|\nabla\phi_k(\bar{\v}_k)\|^2+\|\nabla\phi(\bar{\v}_k)\|^2\right)-\frac{3}{8L}\|\nabla\phi(\bar{\v}_k)\|^2\\
	&=\frac{1}{4L}\|\nabla\phi_k(\bar{\v}_k)\|^2-\frac{1}{4L}\|\nabla\phi(\bar{\v}_k)\|^2\\
	&\stackrel{(b)}{\leq} \frac{1}{4L}\|\nabla\phi_k(\bar{\v}_k)\|^2-\frac{\mu}{2L}\left(\phi(\bar{\v}_k)-\phi(\v_*)\right),
	\end{aligned}
	\end{equation} 
	where (a) holds by using $\left\langle \a,\b\right\rangle\leq \frac{1}{2}(\|\a\|^2+\|\b\|^2)$, and (b) holds by the PL property of $\phi$.

	Define $\Delta_k=\phi(\bar{\v}_k)-\phi(\v_*)$. Combining (\ref{eq:thm:2}) and (\ref{thm:eq:3}), we can see that
	\begin{align*}
	\E[\Delta_k-\Delta_{k-1}]\leq\frac{3}{2}\left(\E[\Delta_{k-1}-\Delta_{k}]+8\eta_kG^2\right)-\frac{\mu}{2L}\E[\Delta_k],
	\end{align*}
	which implies that
	\begin{align*}
	\left(\frac{5}{2}+\frac{\mu}{2L}\right)\E[\Delta_k]\leq \frac{5}{2}\E[\Delta_{k-1}]+12\eta_kG^2.
	\end{align*}
	As a result, we have
	\begin{align*}
	\E[\Delta_k]&\leq \frac{5}{5+\mu/L}\E[\Delta_{k-1}]+\frac{24\eta_kG^2}{5+\mu/L}=\left(1-\frac{\mu/L}{5+\mu/L}\right)\left(\E[\Delta_{k-1}]+\frac{24}{5}\eta_kG^2\right)\\
	&\leq \left(1-\frac{\mu/L}{5+\mu/L}\right)^k\E[\Delta_0]+\frac{24}{5}G^2\sum_{j=1}^{k}\eta_j\left(1-\frac{\mu/L}{5+\mu/L}\right)^{k+1-j}.
	\end{align*}
	By setting $\eta_k=\eta_0\exp\left(-(k-1)\frac{\mu/L}{5+\mu/L}\right)$, we have
	\begin{align*}
	\E[\Delta_k]&\leq \left(1-\frac{\mu/L}{5+\mu/L}\right)^k\E[\Delta_0]+\frac{24}{5}G^2\eta_0\sum_{j=1}^{k}\exp\left(-k\frac{\mu/L}{5+\mu/L}\right)\\
	&\leq \exp\left(-k\frac{\mu/L}{5+\mu/L}\right)\Delta_0+\frac{24}{5}G^2\eta_0k\exp\left(-k\frac{\mu/L}{5+\mu/L}\right).
	\end{align*}
	To achieve $\E[\Delta_k]\leq \epsilon$, it suffices to let $K$ satisfy $\exp\left(-K\frac{\mu/L}{5+\mu/L}\right)\leq \min\left(\frac{\epsilon}{2\Delta_0},\frac{5\epsilon}{48KG^2\eta_0}\right)$, i.e. $K\geq \left(\frac{5L}{\mu}+1\right)\max\left(\log\frac{2\Delta_0}{\epsilon},\log K+\log \frac{48G^2\eta_0}{5\epsilon}\right)$.

	Since $\eta_kT_kL\geq \max(2,16\tilde{L}^2)$, by the setting of $\eta_k$, we set $T_k=\frac{\max(2,16\tilde{L}^2)}{L\eta_0}\exp\left((k-1)\frac{\mu/L}{5+\mu/L}\right)$. Then the total iteration complexity is
	\begin{align*}
	\sum_{k=1}^{K}T_k\leq\frac{\max(2,16\tilde{L}^2)}{L\eta_0}\cdot\frac{\exp\left(K\frac{\mu/L}{5+\mu/L}\right)-1}{\exp\left(\frac{\mu/L}{5+\mu/L}\right)-1}=\widetilde{O}\left(\frac{KG^2}{\mu\epsilon}\right)=\widetilde{O}\left(\frac{LG^2}{\mu^2\epsilon}\right).
	\end{align*}
	The required number of samples is
	\begin{align*}
	\sum_{k=1}^{K}m_k=\frac{2(\sigma^2+C)L}{p(1-p)G^2\eta_0\max(2,16\tilde{L}^2)}\cdot\frac{\exp\left(K\frac{\mu/L}{5+\mu/L}\right)-1}{\exp\left(\frac{\mu/L}{5+\mu/L}\right)-1}=\widetilde{O}\left(\frac{L^3\sigma^2}{\mu^2\epsilon}\right).
	\end{align*}
\end{proof}

\subsection{Proof of Lemma \ref{lemma:PDadagrad}}
\begin{proof}
	Define $\alpha_{*,k}=\arg\max\limits_{\alpha}f(\bar{\v}_k,\alpha)$, $\u=(\v^\top,\alpha)^\top\in\R^{d+3}$, $\u_{*,k}=(\v_*^\top, \alpha_{*,k})^\top$, $\u_t^k=((\v_t^k)^\top,\alpha_t^k)^\top$.
	\begin{equation}
	\label{ineq:main:adagrad}
	\begin{aligned}
	&\phi_k(\bar{\v}_k)-\min_{\v}\phi_k(\v)\stackrel{(a)}{=}\max_{\alpha}\left[f(\bar{\v}_k,\alpha)+\frac{1}{2\gamma}\|\bar{\v}_k-\bar{\v}_{k-1}\|^2\right]-\min_{\v}\max_{\alpha}\left[f(\v,\alpha)+\frac{1}{2\gamma}\|\v-\bar{\v}_{k-1}\|^2\right]\\
	&\stackrel{(b)}{\leq} \left[f(\bar{\v}_k,\alpha_{*,k})+\frac{1}{2\gamma}\|\bar{\v}_k-\bar{\v}_{k-1}\|^2\right]-\left[f(\s_k,\bar{\alpha}_k)+\frac{1}{2\gamma}\|\s_k-\bar{\v}_{k-1}\|^2\right]\\
	&\stackrel{(c)}\leq \frac{1}{T_k}\sum_{t=1}^{T_k}\left[f(\v_t^k,\alpha_{*,k})+\frac{1}{2\gamma}\|\v_t^k-\bar{\v}_{k-1}\|^2-\left(f(\s_k,\alpha_t^k)+\frac{1}{2\gamma}\|\s_k-\bar{\v}_{k-1}\|^2\right)\right]\\
	&=\frac{1}{T_k}\sum_{t=1}^{T_k}\left[\left(f(\v_t^k,\alpha_{*,k})+\frac{1}{2\gamma}\|\v_t^k-\bar{\v}_{k-1}\|^2\right)-\left(f(\v_t^k,\alpha_t^k)+\frac{1}{2\gamma}\|\v_t^k-\bar{\v}_{k-1}\|^2\right)\right.\\
	&\quad\quad\left.+\left(f(\v_t^k,\alpha_t^k)+\frac{1}{2\gamma}\|\v_t^k-\bar{\v}_{k-1}\|^2\right)-\left(f(\s_k,\alpha_t^k)+\frac{1}{2\gamma}\|\s_k-\bar{\v}_{k-1}\|^2\right)\right]\\
	&\leq \frac{1}{T_k}\sum_{t=1}^{T_k}\left\langle\nabla_{\v} \left(f(\v_t^k,\alpha_t^k)+\frac{1}{2\gamma}\|\v_t^k-\v_t^0\|^2\right),\v_t^k-\s_k\right\rangle+\left\langle-\nabla_{\alpha} \left(f(\v_t^k,\alpha_t^k)+\frac{1}{2\gamma}\|\v_t^k-\v_t^0\|^2\right),\alpha_t^k-\alpha_{*,k}\right\rangle\\
	&=\frac{\sum_{t=1}^{T_k}\left\langle\u_t^k-\u_{*,k},\hat\g_t^k\right\rangle}{T_k}+\frac{\sum_{t=1}^{T_k}\left\langle\u_t^k-\u_{*,k},\g_t^k-\hat\g_t^k\right\rangle}{T_k}\\
	&=\mathbf{I} + \mathbf{II}
	%&\stackrel{()}\leq \frac{\|\bar{\v}_{k-1}-\s_k\|^2}{2\eta_k T_k}+\frac{\|\bar{\alpha}_{k-1}-\alpha_{*,k}\|^2}{2\eta_k T_k}+\eta_k G^2 + \frac{\sum_{t=1}^{T_k}(\u_t^k-\u_{*,k})^\top (\G(\u_t^k)-\hat{\G}(\u_t^k))}{T_k},	
	\end{aligned}
	\end{equation}
	where (a) comes from the definition of $\phi_k$, (b) holds because $\min\limits_{\v}\max\limits_{\alpha}\left[f(\v,\alpha)+\frac{1}{2\gamma}\|\v-\bar{\v}_{k-1}\|^2\right]\geq f(\s_k,\bar{\alpha}_k)+\frac{1}{2\gamma}\|\s_k-\bar{\v}_{k-1}\|^2 $, (c) holds by Jensen's inequality. \\
	
	Now we bound $\mathbf{I}$ and $\mathbf{II}$ separately. Define $\|\u\|_{H}=\sqrt{\u^\top H\u}$, $\psi_0^{k}(\u)=0$,  $\psi_{T_k}^{k,*}$ to be the conjugate of $\frac{1}{\eta_k}\psi_{T_k}^k$, which is
	%\vspace*{-0.1in}
	%\begin{equation*}
	$\psi_{T_k}^{k,*}(\g)=\sup_{\u}\left\{\langle\g,\u\rangle-\frac{1}{\eta_k}\psi_{T_k}^{k}(\u)\right\}.$
	%\end{equation*}
	Note that 
	\begin{equation}
	\label{ineq:adagrad1}
	\begin{aligned}
	T_k\cdot\mathbf{I}&=\sum_{t=1}^{T_k}\left\langle\hat{\g}_t^k,\u_t^k\right\rangle-\sum_{t=1}^{T_k}\left\langle\hat{\g}_t^k,\u_{*,k}\right\rangle-\frac{1}{\eta_k}\psi_{T_k}^k(\u_{*,k})+\frac{1}{\eta_k}\psi_{T_k}^k(\u_{*,k})\\
	&\leq \frac{1}{\eta_k}\psi_{T_k}^k(\u_{*,k})+\sum_{t=1}^{T_k}\left\langle\hat{\g}_t^k,\u_t^k\right\rangle+\sup_{\u}\left\{\left\langle-\sum_{t=1}^{T_k}\hat{\g}_t^k,\u\right\rangle-\frac{1}{\eta_k}\psi_{T_k}^k(\u_{*,k})\right\}\\
	&=\frac{1}{\eta_k}\psi_{T_k}^k(\u_{*,k})+\sum_{t=1}^{T_k}\left\langle\hat{\g}_t^k,\u_t^k\right\rangle+\psi_{T_k}^{k,*}\left(-\sum_{t=1}^{T_k}\hat{\g}_t^k\right),
	\end{aligned}
	\end{equation}
	where the last equality holds by the definition of $\psi_{T_k}^{k,*}$.
	
	In addition, note that 
	\begin{equation}
	\label{ineq:adagrad2}
	\begin{aligned}
	\psi_{T_k}^{k,*}\left(-\sum_{t=1}^{T_k}\hat{\g}_t^k\right)
	&\stackrel{(a)}{=}\left\langle-\sum_{t=1}^{T_k}\hat{\g}_{t}^k,\u_{T_k+1}^k\right\rangle-\frac{1}{\eta_k}\psi_{T_k}^{k}(\u_{T_k+1}^k)
	\stackrel{(b)}{\leq}\left\langle-\sum_{t=1}^{T_k}\hat{\g}_{t}^k,\u_{T_k+1}\right\rangle-\frac{1}{\eta_k}\psi_{T_k-1}^{k}(\u_{T_k+1}^k)\\
	&\leq \sup_{\u}\left\{\left\langle-\sum_{t=1}^{T_k}\hat{\g}_k^t,\u\right\rangle-\frac{1}{\eta_k}\psi_{T_k-1}^k(\u)\right\}
	=\psi_{T_k-1}^{k,*}\left(-\sum_{t=1}^{T_k}\hat{\g}_t^k\right)\\
	&\stackrel{(c)}{\leq} \psi_{T_k-1}^{k,*}\left(-\sum_{t=1}^{T_k-1}\hat\g_t^k\right)+\left\langle-\g_{T_k}^k,\nabla\psi_{T_k-1}^{k,*}\left(-\sum_{t=1}^{T_k-1}\hat\g_t^k\right)\right\rangle+\frac{\eta_k}{2}\|\hat\g_{T_k}\|_{\psi_{T_k-1}^{k,*}}^2,
	\end{aligned}
	\end{equation}
	where (a) holds due to the update of the Algorithm \ref{alg:proximal-pdadagrad-PL}, (b) holds since $\psi_{t+1}^k(\u)\geq\psi_{t}^{k}(\u)$, (c) holds by the $\eta_k$-smoothness of $\psi_{t}^{k,*}$ with respect to $\|\cdot\|_{\psi_{t}^{k,*}}=\|\cdot\|_{(H_{t}^k)^{-1}}$.
	%comes from the standard analysis of primal-dual stochastic gradient method. 
	
	%Combining (\ref{ineq:adagrad1}) and
	By (\ref{ineq:adagrad2}) and noting that $\nabla\psi_{T_k-1}^{k,*}\left(-\sum_{t=1}^{T_k-1}\hat\g_t^k\right)=\u_{T_k}^k$, we have
	\begin{equation}
	\label{ineq:adagrad3}
	\begin{aligned}
	\sum_{t=1}^{T_k}\left\langle\hat{\g}_t^k,\u_t^k\right\rangle+\psi_{T_k}^{k,*}\left(-\sum_{t=1}^{T_k}\hat{\g}_t^k\right)\leq \sum_{t=1}^{T_k-1}\left\langle\hat{\g}_t^k,\u_t^k\right\rangle+\psi_{T_k-1}^{k,*}\left(-\sum_{t=1}^{T_k-1}\hat\g_t^k\right)+\frac{\eta_k}{2}\|\hat\g_{T_k}\|_{\psi_{T_k-1}^{k,*}}^2
	\end{aligned}
	\end{equation}
	Using (\ref{ineq:adagrad3}) recursively and noting that $\psi_0^k(\u)=0$, we know that 
	\begin{equation}
	\label{ineq:adagrad4}
	\begin{aligned}
	\sum_{t=1}^{T_k}\left\langle\hat{\g}_t^k,\u_t^k\right\rangle+\psi_{T_k}^{k,*}\left(-\sum_{t=1}^{T_k}\hat{\g}_t^k\right)\leq \frac{\eta_k}{2}\sum_{t=1}^{T_k}\|\hat\g_t^k\|_{\psi_{t-1}^{k,*}}^2
	\end{aligned}
	\end{equation}
	
	Combining (\ref{ineq:adagrad1}) and (\ref{ineq:adagrad4}), we have
	\begin{equation}
	\label{ineq:adagrad5}
	\mathbf{I}\leq \frac{1}{\eta_k T_k}\psi_{T_k}^k(\u_{*,k})+\frac{\eta_k}{2T_k}\sum_{t=1}^{T_k}\|\hat\g_t^k\|_{\psi_{t-1}^{k,*}}^2
	\end{equation}
	
	By Lemma 4 of~\citep{duchi2011adaptive} and setting $\delta\geq \max_{t}\|\hat\g_t^k\|_{\infty}$, we know that $\sum_{t=1}^{T_k}\|\hat\g_t^k\|_{\psi_{t-1}^{k,*}}^2\leq 2\sum_{i=1}^{d+3}\|\hat{\g}_{1:T_k}^k\|_2$, and hence
	\begin{equation}
	\label{ineq:adagrad6}
	\begin{aligned}
	\mathbf{I}&\leq \frac{1}{\eta_k T_k}\psi_{T_k}^k(\u_{*,k})+\frac{\eta_k}{T_k}\sum_{i=1}^{d+3}\|\hat{\g}_{1:T_k}^k\|_2\\
	&=\frac{\delta\|\u_1^k-\u_{*,k}\|_2^2}{2\eta_k T_k}+\frac{\left\langle\u_1^k-\u_{*,k}, \text{diag}(s_{T_k}^k)(\u_1^k-\u_{*,k})\right\rangle}{2\eta_k T_k}+\frac{\eta_k}{T_k}\sum_{i=1}^{d}\|\hat{\g}_{1:T_k}^k\|_2\\
	&\leq \frac{\delta+\max_{i}\|\hat{\g}_{1:T_k,i}^k\|_2}{2\eta_kT_k}\|\u_1^k-\u_{*,k}\|_2^2+\frac{\eta_k}{T_k}\sum_{i=1}^{d+3}\|\hat{\g}_{1:T_k}^k\|_2
	\end{aligned}
	\end{equation}
	
	%\frac{1}{\eta_k}\psi_{T_k}^k(\u_{*,k})
	Denote $\E_{k-1}$ by taking the conditional expectation conditioning on filtration $\mathcal{F}_{k-1}$, where $\mathcal{F}_{k-1}$ is the $\sigma$-algebra generated by all random variables until $\bar{\v}_{k-1}$ is generated.
	%Noting that $\hat{\g}$ is an unbiased estimator of $\g$, we know that $\E_{k-1}(\mathbf{II})=0$. 
	Taking $\E_{k-1}$ on both sides of (\ref{ineq:main:adagrad}), and employing (\ref{ineq:adagrad6}) yields
	\begin{equation}
	\begin{aligned}
	&\E_{k-1}\left[\phi_k(\bar{\v}_k)-\min_{\v}\phi_k(\v)\right]\\
	&\leq  \E_{k-1}\left[\frac{\delta+\max_{i}\|\hat{\g}_{1:T_k,i}^k\|_2}{2\eta_kT_k}\left(\|\bar{\v}_{k-1}-\s_k\|_2^2+\|\bar{\alpha}_{k-1}-\alpha_{*,k}\|_2^2\right)+\frac{\eta_k}{T_k}\sum_{i=1}^{d+3}\|\hat{\g}_{1:T_k}^k\|_2\right]+\E_{k-1}(\mathbf{II})\\
	&=\left(\|\bar{\v}_{k-1}-\s_k\|_2^2\right)\E_{k-1}\left(\frac{\delta+\max_{i}\|\hat{\g}_{1:T_k,i}^k\|_2}{2\eta_kT_k}\right)+\E_{k-1}\left(\frac{\delta+\max_{i}\|\hat{\g}_{1:T_k,i}^k\|_2}{2\eta_kT_k}\|\bar{\alpha}_{k-1}-\alpha_{*,k}\|_2^2\right)\\
	&\quad\quad+\E_{k-1}\left(\frac{\eta_k}{T_k}\sum_{i=1}^{d+3}\|\hat{\g}_{1:T_k}^k\|_2\right)+\E_{k-1}(\mathbf{II})
	\end{aligned} 
	\end{equation}
	where the equality holds since $\bar{\v}_{k-1}-\s_k$ is measurable with respect to $\mathcal{F}_{k-1}$.
	%  $\frac{\delta+\max_{i}\|\hat{\g}_{1:T_k,i}^k\|_2}{2\eta_kT_k}$ and $\bar{\alpha}_{k-1}-\alpha_{*,k}$ are independent conditioning on $\mathcal{F}_{k-1}$. 
	
	%By the update of $\bar{\alpha}_{k-1}$, $\tilde{L}$-Lipschitz continuity of $h(\w;\x)$, and noting that $\alpha_{*,k}=\E\left[h(\w;\x)\middle\vert y=-1\right]-\E\left[h(\w;\x)\middle\vert y=1\right]$, we have
	%and the fact that $\|\a+\b+\c\|^2\leq 3\|\a\|^2+3\|\b\|^2+3\|\c\|^2$, we have
	Note that 
	\begin{align*}
	&\E_{k-1}\left(\frac{\delta+\max_{i}\|\hat{\g}_{1:T_k,i}^k\|_2}{2\eta_kT_k}\|\bar{\alpha}_{k-1}-\alpha_{*,k}\|_2^2\right)=\E_{k-1}\left(\frac{\delta+\max_{i}\|\hat{\g}_{1:T_k,i}^k\|_2}{2\eta_kT_k}\left\|\bar{\alpha}_{k-1}-\alpha_{*,k-1}+\alpha_{*,k-1}-\alpha_{*,k}\right\|^2\right)\\
	&\leq \E_{k-1}\left(\frac{\delta+\max_{i}\|\hat{\g}_{1:T_k,i}^k\|_2}{2\eta_kT_k}\left(2\|\bar{\alpha}_{k-1}-\alpha_{*,k-1}\|^2+2\|\alpha_{*,k-1}-\alpha_{*,k}\|^2\right)\right)\\
	&\stackrel{(a)}{=}\E_{k-1}\left(\frac{\delta+\max_{i}\|\hat{\g}_{1:T_k,i}^k\|_2}{2\eta_kT_k}\right)\E_{k-1}\left(2\|\bar{\alpha}_{k-1}-\alpha_{*,k-1}\|^2\right)+\E_{k-1}\left(\frac{\delta+\max_{i}\|\hat{\g}_{1:T_k,i}^k\|_2}{2\eta_kT_k}\cdot 2\|\alpha_{*,k-1}-\alpha_{*,k}\|^2\right)\\
	&\stackrel{(b)}{\leq} \E_{k-1}\left(\frac{\delta+\max_{i}\|\hat{\g}_{1:T_k,i}^k\|_2}{2\eta_kT_k}\right)\cdot\frac{4(\sigma^2+C)}{m_{k-1} p(1-p)}+\E_{k-1}\left(\frac{\delta+\max_{i}\|\hat{\g}_{1:T_k,i}^k\|_2}{2\eta_kT_k}\cdot 2\|\alpha_{*,k-1}-\alpha_{*,k}\|^2\right)\\
	&\stackrel{(c)}{\leq} \E_{k-1}\left(\frac{\delta+\max_{i}\|\hat{\g}_{1:T_k,i}^k\|_2}{2\eta_kT_k}\right)\cdot\frac{4(\sigma^2+C)}{m_{k-1}p(1-p)}+\E_{k-1}\left(\frac{\delta+\max_{i}\|\hat{\g}_{1:T_k,i}^k\|_2}{2\eta_kT_k}\cdot 8\tilde{L}^2\|\bar{\v}_{k-1}-\bar{\v}_k\|^2\right)
	\end{align*}
	where $(a)$ holds because $\bar{\alpha}_{k-1}-\alpha_{*,k-1}$ and $\frac{\delta+\max_{i}\|\hat{\g}_{1:T_k,i}^k\|_2}{2\eta_kT_k}$ are independent conditioning on $\mathcal{F}_{k-1}$, (b) holds because of the update of $\bar{\alpha}_{k-1}$ and $\alpha_{*,k}=\E\left[h(\bar{\w}_k;\x)\middle\vert y=-1\right]-\E\left[h(\bar{\w}_k;\x)\middle\vert y=1\right]$, (c) holds due to the $2\tilde{L}$-Lipschitz continuity of $\E\left[h(\w;\x)\middle\vert y=-1\right]-\E\left[h(\w;\x)\middle\vert y=1\right]$. 
	%\begin{align*}
	%&\E_{k-1}\|\bar{\alpha}_{k-1}-\alpha_{*,k}\|^2=\E_{k-1}\left\|\bar{\alpha}_{k-1}-\alpha_{*,k-1}+\alpha_{*,k-1}-\alpha_{*,k}\right\|^2\\
	%&\leq \E_{k-1} \left(2\|\bar{\alpha}_{k-1}-\alpha_{*,k-1}\|^2+2\|\alpha_{*,k-1}-\alpha_{*,k}\|^2\right)\leq \frac{2\sigma^2}{m}+2\tilde{L}^2\E_{k-1}\|\bar{\v}_{k-1}-\bar{\v}_{k}\|^2 %3\tilde{L}^2\|\bar{\v}_{k-1}-\s_k\|^2+\frac{3\sigma^2}{m},
	%\end{align*}
	
	%where $m$ is the minibatch size, and $\sigma^2=$

	%Hence, we have
	%\begin{align*}
	%\E_{k-1}\left[\phi_k(\bar{\v}_k)-\min_{\v}\phi_k(\v)\right]\leq  \frac{\|\bar{\v}_{k-1}-\s_k\|^2}{2\eta_k T_k}+\frac{\tilde{L}^2\|\bar{\v}_{k-1}-\s_k\|^2+\frac{\sigma^2}{m}}{2\eta_k T_k}+\eta_k G^2.
	%\end{align*}
	%Conditioning on all stochastic events until $\bar{\v}_{k-1}$ is generated, 
	Taking $m_{k-1}\geq \frac{2(\sigma^2+C)}{p(1-p)(d+3)\eta_k^2}$, then we have 
	\begin{align*}
	&\quad\E_{k-1}\left[\phi_k(\bar{\v}_k)-\min_{\v}\phi_k(\v)\right]\\
	&\leq  \E_{k-1}\left(\frac{\delta+\max_{i}\|\hat{\g}_{1:T_k,i}^k\|_2}{2\eta_kT_k}\right)\|\bar{\v}_{k-1}-\s_k\|^2+\E_{k-1}\left(\frac{\eta_k}{T_k}\sum_{i=1}^{d+3}\|\hat{\g}_{1:T_k,i}^k\|_2\right)+\E_{k-1}(\mathbf{II})\\
	&\quad\quad+\E_{k-1}\left(\frac{\delta+\max_{i}\|\hat{\g}_{1:T_k,i}^k\|_2}{2T_k}\right)\cdot 2\eta_k(d+3)+\E_{k-1}\left(\frac{\delta+\max_{i}\|\hat{\g}_{1:T_k,i}^k\|_2}{2\eta_kT_k}\cdot 8\tilde{L}^2\|\bar{\v}_{k-1}-\bar{\v}_k\|^2\right)\\
	&=  \E_{k-1}\left(\frac{\delta+\max_{i}\|\hat{\g}_{1:T_k,i}^k\|_2}{2\eta_kT_k}\right)\|\bar{\v}_{k-1}-\s_k\|^2+\E_{k-1}\left[\frac{\eta_k}{T_k}\left(\sum_{i=1}^{d+3}\|\hat{\g}_{1:T_k,i}^k\|_2+(d+3)\left(\delta+\max_{i}\|\hat{\g}_{1:T_k,i}^k\|_2\right)\right)\right]\\
	&\quad\quad+\E_{k-1}(\mathbf{II})+\E_{k-1}\left(\frac{\delta+\max_{i}\|\hat{\g}_{1:T_k,i}^k\|_2}{2\eta_kT_k}\cdot 8\tilde{L}^2\|\bar{\v}_{k-1}-\bar{\v}_k\|^2\right)\\
	\end{align*}
		Define $\widetilde{\alpha}_0^k=\alpha_0^k$ and  $$\widetilde{\alpha}_{t+1}^k=\arg\min_{\alpha}\left[\frac{\eta_k}{t}\sum_{\tau=1}^{t}\left(-\nabla_{\alpha}f(\v_t^k,\alpha_t^k)-(-\nabla_{\alpha}F(\v_{t}^k,\alpha_t^k;\xi_t^k))\right)\alpha+\frac{1}{t}\psi_t^k(\alpha)\right],$$ 
		where $\psi_t^k(\alpha)=\psi_t^k(\u)$ in which $\u=[0,\ldots,0,\alpha]$ and $\u_0^k=[0,\ldots,0,\alpha_0^k]$.
	By setting
	$$T_k=\inf\left\{\tau:\tau\geq M_k\max\left(\frac{(\delta+\max_{i}\|\hat{\g}_{1:\tau,i}^{k}\|_2)\max(1,8\tilde{L}^2)}{c},2c\left(\sum_{i=1}^{d+3}\|\hat{\g}_{1:\tau,i}^{k}\|_2+(d+3)\left(\delta+\max_{i}\|\hat{\g}_{1:\tau,i}^k\|_2\right)\right)\right)\right\},$$ then $T_k$ is a stopping time which is bounded almost surely. By stopping time argument, we have
	
		\begin{equation*}
		\begin{aligned}
		&\E_{k-1}\left[\frac{\sum_{t=1}^{T_k}(\v_t^k-\v_*)^\top \left(\nabla_{\v}f(\v_t^k,\alpha_t^k)-\nabla_{\v}F(\v_{t}^k,\alpha_t^k;\xi_t^k))\right)}{T_k}\right]=0\\
		&\E_{k-1}\left[\frac{\sum_{t=1}^{T_k}(\alpha_t^k-\widetilde{\alpha}_t^k)^\top \left(-\nabla_{\alpha}f(\v_t^k,\alpha_t^k)-\left(-\nabla_{\alpha}F(\v_{t}^k,\alpha_t^k;\xi_t^k)\right)\right)}{T_k}\right]=0
		\end{aligned}
		\end{equation*}
		Hence we know that 
		\begin{equation*}
			\begin{aligned}
			\E_{k-1}\left(\mathbf{II}\right)=\E_{k-1}\left[\frac{\sum_{t=1}^{T_k}(\widetilde{\alpha}_t^k-\alpha_{*,k}) \left(-\nabla_{\alpha}f(\v_t^k,\alpha_t^k)-(-\nabla_{\alpha}F(\v_{t}^k,\alpha_t^k;\xi_t^k))\right)}{T_k}\right].
			\end{aligned}
		\end{equation*}
			Note that the variance of stochastic gradient is smaller than its second moment, we can follow the similar analysis of bounding $\mathbf{I}$ to show that
			 $$\E_{k-1}\left(\mathbf{II}\right)\leq \E_{k-1}\left[\frac{\delta+\max_{i}\|\hat{\g}_{1:T_k,i}^k\|_2}{2\eta_kT_k}\|\u_1^k-\u_{*,k}\|_2^2+\frac{\eta_k}{T_k}\sum_{i=1}^{d+3}\|\hat{\g}_{1:T_k}^k\|_2\right].$$
			 Following the same analysis of bounding the RHS of~(\ref{ineq:adagrad6}), we know that
	%Hence,
	%By stopping time argument, we have $\E_{k-1}(\mathbf{II})=0$, and hence 
	\begin{align*}
	\E_{k-1}\left[\phi_k(\bar{\v}_k)-\min_{\v}\phi_k(\v)\right]\leq \frac{c\left(\|\bar{\v}_{k-1}-\s_k\|_2^2+\E_{k-1}\|\bar{\v}_{k-1}-\bar{\v}_k\|_2^2\right)}{\eta_k M_k}+\frac{\eta_k}{cM_k}.
	\end{align*}
\end{proof}
\subsection{Proof of Theorem~\ref{theorem:adagrad}}
\begin{proof}
	Define $\phi_k(\v)=\phi(\v)+\frac{1}{2\gamma}\|\v-\bar{\v}_{k-1}\|^2$. We can see that $\phi_k(\v)$ is convex and smooth function since $\gamma\leq 1/L$. The smoothness parameter of $\phi_k$ is $\hat{L}=L+\gamma^{-1}$. Define $\s_k=\arg\min_{\v\in\R^{d+2}}\phi_k(\v)$. According to Theorem 2.1.5 of~\citep{nesterov2013introductory}, we have
	\begin{align}
	\label{ineq:b3}
	\|\nabla\phi_k(\bar{\v}_k)\|^2\leq 2\hat{L}(\phi_k(\bar{\v}_k)-\phi_k(\s_k)).
	\end{align}
	%123
	Combining (\ref{ineq:b3}) with Lemma \ref{lemma:PDadagrad} yields
	\begin{align}
	\label{ineq:b2}
	\E_{k-1}\|\nabla\phi_k(\bar{\v}_k)\|^2\leq 2\hat{L}\left(\frac{c\left(\|\bar{\v}_{k-1}-\s_k\|_2^2+\E_{k-1}\|\bar{\v}_{k-1}-\bar{\v}_k\|_2^2\right)}{\eta_k M_k}+\frac{\eta_k}{cM_k} \right).
	\end{align}
	Note that $\phi_k(\bar{\v})$ is $(\gamma^{-1} - L)$-strongly convex, and $\gamma=\frac{1}{2L}$, we have 
	\begin{equation}
	\label{ineq:sc}
	\phi_k(\bar{\v}_{k-1})\geq \phi_k(\s_k)+\frac{L}{2}\|\bar{\v}_{k-1}-\s_k\|^2.
	\end{equation}
	Plugging in $\s_k$ into Lemma \ref{lemma:PDadagrad} and combining (\ref{ineq:sc}) yield
	\begin{align*}
	\E_{k-1}[\phi(\bar{\v}_k)+L\|\bar{\v}_k-\bar{\v}_{k-1}\|^2]\leq \phi_k(\bar{\v}_{k-1})-\frac{L}{2}\|\bar{\v}_{k-1}-\s_k\|^2+\frac{c\left(\|\bar{\v}_{k-1}-\s_k\|_2^2+\E_{k-1}\|\bar{\v}_{k-1}-\bar{\v}_k\|_2^2\right)}{\eta_k M_k}+\frac{\eta_k}{cM_k}
	\end{align*}
	By taking $\eta_k M_k L\geq 4c$, rearranging the terms, and noting that $\phi_k(\bar{\v}_{k-1})=\phi(\bar{\v}_{k-1})$, we have
	\begin{equation}
	\label{ineq:b1}
	\frac{c\left(\|\bar{\v}_{k-1}-\s_k\|_2^2+\E_{k-1}\|\bar{\v}_{k-1}-\bar{\v}_k\|_2^2\right)}{\eta_k M_k}\leq \phi(\bar{\v}_{k-1})-	\E_{k-1}\left[\phi(\bar{\v}_k)\right] +\frac{\eta_k}{cM_k}.
	\end{equation}
	Combining (\ref{ineq:b1}) and (\ref{ineq:b2}) yields
	\begin{equation}
	%\label{eq:thm:2}
	\E_{k-1}\|\nabla\phi_k(\bar{\v}_k)\|^2\leq 6L\left(\phi(\bar{\v}_{k-1})-	\E_{k-1}\left[\phi(\bar{\v}_k)\right] +2\frac{\eta_k}{cM_k}\right).
	\end{equation}
	Taking expectation on both sides over all randomness until $\bar{\v}_{k-1}$ is generated and by the tower property, we have
	\begin{equation}
	\label{ineq:thm:2}
	\E\|\nabla\phi_k(\bar{\v}_k)\|^2\leq 6L\left(\E\left[\phi(\bar{\v}_{k-1})-\phi(\v_*)\right]-	\E\left[\phi(\bar{\v}_k)-\phi(\v_*)\right] +\frac{2\eta_k} {cM_k}\right).
	\end{equation}
	Note that $\phi(\v)$ is $L$-smooth and hence is $L$-weakly convex, so we have 
	\begin{equation}
	\label{ineq:thm:1}
	\begin{aligned}
	&\phi(\bar{\v}_{k-1})\geq\phi(\bar{\v}_{k})+ \left\langle\nabla\phi(\bar{\v}_k),\bar{\v}_{k-1}-\bar{\v}_k\right\rangle-\frac{L}{2}\|\bar{\v}_{k-1}-\bar{\v}_k\|^2\\
	&=\phi(\bar{\v}_k)+\left\langle\nabla\phi(\bar{\v}_k)+2L(\bar{\v}_k-\bar{\v}_{k-1}), \bar{\v}_{k-1}-\bar{\v}_k\right\rangle+\frac{3}{2}L\|\bar{\v}_{k-1}-\bar{\v}_{k}\|^2\\
	&\stackrel{(a)}{=}\phi(\bar{\v}_k)+\left\langle\nabla\phi_k(\bar{\v}_k),\bar{\v}_{k-1}-\bar{\v}_k\right\rangle
	+\frac{3}{2}L\|\bar{\v}_{k-1}-\bar{\v}_{k}\|^2\\
	&\stackrel{(b)}{=}\phi(\bar{\v}_k)-\frac{1}{2L}\left\langle \nabla\phi_k(\bar{\v}_k),\nabla\phi_k(\bar{\v}_k)-\nabla\phi(\bar{\v}_k)\right\rangle+\frac{3}{8L}\|\nabla\phi_k(\bar{\v}_k)-\nabla\phi(\bar{\v}_k)\|^2\\
	%&= \phi(\bar{\v}_k)-\frac{1}{2L}\|\nabla\phi_k(\bar{\v}_k)\|^2+\frac{1}{2L}\left\langle\nabla\phi_k(\bar{\v}_k),\nabla\phi(\bar{\v}_k)\right\rangle+\frac{3}{8L}\|\nabla\phi_k(\bar{\v}_k)\|^2-\frac{3}{4L}\left\langle\nabla\phi_k(\bar{\v}_k),\nabla\phi(\bar{\v}_k)\right\rangle\\
	&=\phi(\bar{\v}_k)-\frac{1}{8L}\|\nabla\phi_k(\bar{\v}_k)\|^2-\frac{1}{4L}\left\langle\nabla\phi_k(\bar{\v}_k),\nabla\phi(\bar{\v}_k)\right\rangle+\frac{3}{8L}\|\nabla\phi(\bar{\v}_k)\|^2,
	\end{aligned}
	\end{equation}
	where (a) and (b) hold by the definition of $\phi_k$. 
	
	Rearranging the terms in (\ref{ineq:thm:1}) yields
	\begin{equation}
	\label{thm:ineq:3}
	\begin{aligned}
	\phi(\bar{\v}_k)-\phi(\bar{\v}_{k-1})&\leq \frac{1}{8L}\|\nabla\phi_k(\bar{\v}_k)\|^2+\frac{1}{4L}\left\langle\nabla\phi_k(\bar{\v}_k),\nabla\phi(\bar{\v}_k)\right\rangle-\frac{3}{8L}\|\nabla\phi(\bar{\v}_k)\|^2\\
	&\stackrel{(a)}{\leq} \frac{1}{8L}\|\nabla\phi_k(\bar{\v}_k)\|^2+\frac{1}{8L}\left(\|\nabla\phi_k(\bar{\v}_k)\|^2+\|\nabla\phi(\bar{\v}_k)\|^2\right)-\frac{3}{8L}\|\nabla\phi(\bar{\v}_k)\|^2\\
	&=\frac{1}{4L}\|\nabla\phi_k(\bar{\v}_k)\|^2-\frac{1}{4L}\|\nabla\phi(\bar{\v}_k)\|^2\\
	&\stackrel{(b)}{\leq} \frac{1}{4L}\|\nabla\phi_k(\bar{\v}_k)\|^2-\frac{\mu}{2L}\left(\phi(\bar{\v}_k)-\phi(\v_*)\right),
	\end{aligned}
	\end{equation} 
	where (a) holds by using $\left\langle \a,\b\right\rangle\leq \frac{1}{2}(\|\a\|^2+\|\b\|^2)$, and (b) holds by the PL property of $\phi$.

	Define $\Delta_k=\phi(\bar{\v}_k)-\phi(\v_*)$. Combining (\ref{ineq:thm:2}) and (\ref{thm:ineq:3}), we can see that
	\begin{align*}
	\E[\Delta_k-\Delta_{k-1}]\leq\frac{3}{2}\left(\E[\Delta_{k-1}-\Delta_{k}]+\frac{2\eta_k}{cM_k}\right)-\frac{\mu}{2L}\E[\Delta_k],
	\end{align*}
	which implies that
	\begin{align*}
	\left(\frac{5}{2}+\frac{\mu}{2L}\right)\E[\Delta_k]\leq \frac{5}{2}\E[\Delta_{k-1}]+\frac{3\eta_k}{cM_k}.
	\end{align*}
	As a result, we have
	\begin{align*}
	\E[\Delta_k]&\leq \frac{5}{5+\mu/L}\E[\Delta_{k-1}]+\frac{6(\eta_k/cM_k)}{5+\mu/L}=\left(1-\frac{\mu/L}{5+\mu/L}\right)\left(\E[\Delta_{k-1}]+\frac{6\eta_k}{5cM_k}\right)\\
	&\leq \left(1-\frac{\mu/L}{5+\mu/L}\right)^k\E[\Delta_0]+\frac{6}{5c}\sum_{j=1}^{k}\frac{\eta_j}{M_j}\left(1-\frac{\mu/L}{5+\mu/L}\right)^{k+1-j}.
	\end{align*}
	By setting $\eta_k=\eta_0\exp\left(-\frac{(k-1)}{2}\frac{\mu/L}{5+\mu/L}\right)$, $M_k=\frac{4c}{L\eta_0}\exp\left(\frac{(k-1)}{2}\frac{\mu/L}{5+\mu/L}\right) $ at $k$-th stage, we have
	\begin{align*}
	\E[\Delta_k]&\leq \left(1-\frac{\mu/L}{5+\mu/L}\right)^k\E[\Delta_0]+\frac{\eta_0^2 L}{10c^2}\sum_{j=1}^{k}\exp\left(-k\frac{\mu/L}{5+\mu/L}\right)\\
	&\leq \exp\left(-k\frac{\mu/L}{5+\mu/L}\right)\Delta_0+\frac{\eta_0^2 L}{10c^2}k\exp\left(-k\frac{\mu/L}{5+\mu/L}\right).
	\end{align*}
	To achieve $\E[\Delta_K]\leq \epsilon$, it suffices to let $K$ satisfy $\exp\left(-K\frac{\mu/L}{5+\mu/L}\right)\leq \min\left(\frac{\epsilon}{2\Delta_0},\frac{5c^2\epsilon}{K\eta_0^2L}\right)$, i.e. $K\geq \left(\frac{5L}{\mu}+1\right)\max\left(\log\frac{2\Delta_0}{\epsilon},\log K+\log \frac{\eta_0^2L}{5c^2\epsilon}\right)$.

	%Since $\eta_k M L\geq 4c$, by the setting of $\eta_k$, 
	%we set $M\geq \frac{4c}{L\eta_0}\exp\left((k-1)\frac{\mu/L}{5+\mu/L}\right) $. %$T_k=\frac{2(4+3\widetilde{L}^2)}{L\eta_0}\exp\left((k-1)\frac{\mu/L}{5+\mu/L}\right)$. 
	Take $c=\frac{1}{\sqrt{d+3}}$. If $\|\hat{\g}_{1:T_k,i}^k\|_2\leq \delta \cdot T_k^{\alpha}$ for $\forall k$, where $0\leq \alpha\leq\frac{1}{2}$, and note that when $\tau\geq 1$, 
	\begin{align*}
	&\max\left(\frac{(\delta+\max_{i}\|\hat{\g}_{1:\tau,i}^{k}\|_2)\max(1,8\tilde{L}^2)}{2c},c\left(\sum_{i=1}^{d+3}\|\hat{\g}_{1:\tau,i}^{k}\|_2+(d+3)\left(\delta+\max_{i}\|\hat{\g}_{1:\tau,i}^k\|_2\right)\right)\right)\\
	&\leq \left[(4+8\tilde{L}^2)\sqrt{d+3}\right]\delta\cdot \tau^\alpha
	\end{align*}
	%	Note that $\sigma^2$ is the variance of a one dimensional random variable, hence when $d$ is large, $\frac{2\sigma^2}{\sqrt{d+3}}$ is negligible. Without loss of generality, we can ass
	so we have $T_k\leq\frac{4c}{L\eta_0}\exp\left(\frac{(k-1)}{2}\frac{\mu/L}{5+\mu/L}\right)
	\cdot \left[(4+8\tilde{L}^2)\sqrt{d+3}\right]\delta T_k^\alpha$, and hence $$T_k\leq \left(\frac{4\delta c}{L\eta_0}\exp\left(\frac{(k-1)}{2}\frac{\mu/L}{5+\mu/L}\right)
	\cdot \left[(4+8\tilde{L}^2)\sqrt{d+3}\right]\right)^{\frac{1}{1-\alpha}}.$$
	%	Note that $\sigma^2$ is the variance of a one dimensional random variable, and hence when $d$ is large, $\frac{2\sigma^2}{\sqrt{d+3}}$ is dominated by $(2+2\tilde{L}^2)\sqrt{d+3}$.

	Noting that $c=\frac{1}{\sqrt{d+3}}$, we can see that the total iteration complexity is
	\begin{align*}
	\sum_{k=1}^{K}T_k\leq\left(\frac{4\delta(4+8\tilde{L}^2)}{L\eta_0}\right)^{\frac{1}{1-\alpha}}\cdot\frac{\exp\left(K\frac{\mu/L}{(5+\mu/L)(2-2\alpha)}\right)-1}{\exp\left(\frac{\mu/L}{(5+\mu/L)(2-2\alpha)}\right)-1}=\widetilde{O}\left(\left(\frac{L\delta^2 d}{\mu^2\epsilon}\right)^{\frac{1}{2(1-\alpha)}}\right).
	\end{align*}
	The required number of samples is
	\begin{align*}
	\sum_{k=1}^{K}m_k=\frac{2(\sigma^2+C)}{p(1-p)\eta_0^2(d+3)}\cdot\frac{\exp\left(K\frac{\mu/L}{5+\mu/L}\right)-1}{\exp\left(\frac{\mu/L}{5+\mu/L}\right)-1}=\widetilde{O}\left(\frac{L^3\sigma^2}{\mu^2\epsilon}\right).
	\end{align*}
\end{proof}
\subsection{Proof of Lemma~\ref{thm:PL1}}
\begin{proof}
	%To be done.
	For any fixed $\w$, define $(a_{\w}^*,b_{\w}^*)=\arg\min\limits_{a,b}\phi(\w,a,b)$ ($\phi(\w,a,b)$ is strongly convex in terms of $(a,b)$, so the argmin is well-defined and unique). Note that 
	\begin{align*}
	\phi(\v)-\phi(\v_*)=\phi(\w,a,b)-\min_{\w,a,b}\phi(\w,a,b)
	=\phi(\w,a,b)-\phi(\w,a_{\w}^*,b_{\w}^*)+\phi(\w,a_{\w}^*,b_{\w}^*)-\min_{\w,a,b}\phi(\w,a,b)
	\end{align*}
	We bound $\phi(\w,a,b)-\phi(\w,a_{\w}^*,b_{\w}^*)$ and $\phi(\w,a_{\w}^*,b_{\w}^*)-\min_{\w,a,b}\phi(\w,a,b)$ respectively:
	%\vspace*{-0.5in}
	\begin{itemize}
		\item Note that $\phi(\w,a,b)$ is strong convex in $(a,b)$ with modulus $2\min(p,1-p)$, so the PL condition holds, which means that  
		\begin{align*}
		\phi(\w,a,b)-\phi(\w,a_{\w}^*,b_{\w}^*)\leq \frac{1}{4\min(p,1-p)}\|\nabla_{(a,b)}\phi(\w,a,b)\|^2
		\end{align*}
		\item 
		\begin{align*}
		&\phi(\w,a_{\w^*},b_{\w^*})-\min\limits_{\w,a,b}\phi(\w,a,b)=\min_{a,b}\phi(\w,a,b)-\min\limits_{\w,a,b}\phi(\w,a,b)\leq \frac{1}{2\mu}\left\|\nabla_{\w}\min_{a,b}\phi(\w,a,b)\right\|^2\\
		&=\frac{1}{2\mu}\left\|\nabla_{\w}\phi(\w,a,b)+\nabla_{\w}\phi(\w,a_{\w}^*,b_{\w}^*)-\nabla_{\w}\phi(\w,a,b)\right\|^2\\
		&\leq  \frac{1}{2\mu} \left(2\left\|\nabla_{\w}\phi(\w,a,b)-\nabla_{\w}\phi(\w,a_{\w}^*,b_{\w}^*)\right\|^2+2\left\|\nabla_{\w}\phi(\w,a,b)\right\|^2\right)\\
		&\leq \frac{1}{2\mu} \left(8\tilde L^2\left\|(a,b)-(a_{\w}^*,b_{\w}^*)\right\|^2+2\left\|\nabla_{\w}\phi(\w,a,b)\right\|^2\right)\\
		&\leq \frac{1}{2\mu}\left(\frac{8\tilde L^2}{4\min(p^2,(1-p)^2)}\left\|\nabla_{(a,b)}\phi(\w,a,b)\right\|^2+2\left\|\nabla_{\w}\phi(\w,a,b)\right\|^2\right),
		\end{align*}
		where the last inequality holds since $\phi(\w,a,b)$ is strongly convex in $(a,b)$ with modulus $2\min(p,1-p)$.
	\end{itemize}
	%\vspace*{-1in}
	Combining these two cases, we know that
	%\begin{align*}
	$\phi(\v)-\phi(\v_*)\leq \frac{1}{2\mu'}\|\nabla\phi(\v)\|^2$,
	%\end{align*}
	where $\mu'=\frac{1}{\max\left(\frac{1}{2\min(p,1-p)}+\frac{2\hat{G}^2}{\mu\min(p^2,(1-p)^2)},\frac{2}{\mu}\right)}$.
\end{proof}
\vspace*{-0.1in}
\subsection{An example that satisfies PL condition}
%\vspace*{-0.3in}
\label{example:PL}
\paragraph{One Hidden Layer Neural Network}
One hidden neural network satisfies $h(\w;\x)=\sigma(\w^\top\x)$, where $\sigma$ is the activation function. We have the following theorem:
\begin{thm}
	\label{thm:onehiddenlayer}
	Let $\sigma$ be the Leaky ReLU activation function such that $\sigma(z)=c_1z$ for $z>0$ and $\sigma(z)=c_2z$ if $z\leq 0$. If $\E[\x|y=1]=\E[\x|y=-1]=0$, $\E\left[\x\x'^\top\middle\vert y=1,y=-1\right]=\textbf{0}_{d\times d}$, then $$f(\w):=\E_{\z,\z'}\left[(1-\sigma(\w^\top\x)+\sigma(\w^\top\x'))^2 \middle\vert y=1,y'=-1\right]$$ satisfies PL condition with $\mu=2\min(c_1^2,c_2^2)\left[\lambda_\text{min}\left(\E\left[\x\x^\top\middle\vert y=1\right]\right)+\lambda_\text{min}\left(\E\left[\x\x^\top\middle\vert y=-1\right]\right)\right]$, where $\lambda_{\text{min}}$ stands for the minimum eigenvalue.
\end{thm}
%\vspace*{-0.3in}
\textbf{Remark}: Consider the case that $\x$ is a zero mean Gaussian distribution with non-degenerate convariance matrix. Then $\mu>0$ since the minimum eigenvalue appeared in the expression of $\mu$ is positive. 
%\vspace*{-0.5in}
\begin{proof}
	Define $g_1(x)=(1-x)^2$, $g_2(\w)=\sigma(\w^\top\x)-\sigma(\w^\top\x')$, $F(\w)=(1-\sigma(\w^\top\x)+\sigma(\w^\top\x'))^2$. We know that 
	$f(\w)=\E_{\z,\z'}\left[F(\w)\middle\vert y=1,y'=-1\right]$, $F(\w)=g_1(g_2(\w))$.
	For fixed $\x$, $\x'$, we can write $\sigma(\w^\top\x)$ and $\sigma(\w^\top\x')$ as $a\w^\top\x$ and $b\w^\top\x'$ respectively, and it is obvious that $a^2\geq \min(c_1^2,c_2^2)$ and $b^2\geq \min(c_1^2,c_2^2)$.
	%Define $a=c_1\mathbb{I}_{[\w^\top\x>0]}+c_2\mathbb{I}_{[\w^\top\x<0]}$, $b=a=c_1\mathbb{I}_{[\w^\top\x'>0]}+c_2\mathbb{I}_{[\w^\top\x'<0]}$.
	Note that $g_1$ is $2$-strongly convex. Since the conditional expectation perserves the strong convexity, as a result, for $\forall \w$, let $\w_p$ be the closest optimal point of $\w$ such that $f_*=f(\w_p)$, we have
	% \sigma(\w_*^\top\x)-\sigma(\w^\top\x)-\left(\sigma(\w_*^\top\x')-\sigma(\w^\top\x')\right)
	%	\begin{align*}
	%		f(\w_*)&\geq f(\w)+\left\langle \nabla f(\w), \E\left[g_2(\w_*)-g_2(\w)\middle\vert y=1,y'=-1\right]\right\rangle+\left(\E\left[g_2(\w_*)-g_2(\w)\middle\vert y=1,y'=-1\right]\right)^2\\
	%		&\geq f(\w)+\left\langle \nabla f(\w), \w_*-\w \right\rangle		
	%	\end{align*}
	\begin{align*}
	&f(\w_p)-f(\w)=\E\left[g_1(g_2(\w_p))\middle\vert y=1,y'=-1 \right]-\E\left[g_1(g_2(\w))\middle\vert y=1,y'=-1\right]\\
	&\geq
	\E\left[\left\langle   \nabla g_1(g_2(\w)),g_2(\w_p)-g_2(\w)\right\rangle\middle\vert y=1,y=-1\right] +\E\left[(g_2(\w)-g_2(\w_p))^2\middle\vert y=1,y'=-1\right]\\
	&=\E\left[\left\langle   2(g_2(\w)-1),(g_2(\w_p)-g_2(\w))\right\rangle\middle\vert y=1,y=-1\right]+\E\left[(g_2(\w)-g_2(\w_p))^2\middle\vert y=1,y'=-1\right]\\ 
	&=\E\left[\left\langle   -2(1-a\w^\top\x+b\w^\top\x'),(a\x^\top-b\x'^\top)(\w_p-\w)\right\rangle\middle\vert y=1,y'=-1\right]\\ &\quad\quad +\E\left[\left((a\x^\top-b\x'^\top)(\w_p-\w)\right)^2\middle\vert y=1,y'=-1\right]\\
	&=\E\left[\left\langle   2(1-a\w^\top\x+b\w^\top\x')(b\x'-a\x),\w_p-\w\right\rangle\middle\vert y=1,y'=-1\right]\\ &\quad\quad +\E\left[\left((a\x^\top-b\x'^\top)(\w_p-\w)\right)^2\middle\vert y=1,y'=-1\right]\\
	&=\left\langle \nabla f(\w), \w_p-\w\right\rangle+\E\left[(\w_p-\w)^\top(a\x-b\x')(a\x^\top-b\x'^\top)(\w_p-\w)\middle\vert y=1,y'=-1\right]\\
	&=\left\langle \nabla f(\w), \w_p-\w\right\rangle+(\w_p-\w)^\top\E\left[(a^2\x\x^\top+b^2\x'\x'^\top)\middle\vert y=1,y'=-1\right](\w_p-\w)\\
	&\geq \left\langle \nabla f(\w), \w_p-\w\right\rangle+(\w_p-\w)^\top\lambda_{\text{min}}\left(\E\left[(a^2\x\x^\top+b^2\x'\x'^\top)\middle\vert y=1,y'=-1\right]\right)(\w_p-\w)\\
	&\stackrel{(*)}{\geq}\left\langle \nabla f(\w), \w_p-\w\right\rangle+\frac{2\lambda_\text{min}\left(\E\left[a^2\x\x^\top\middle\vert y=1\right]\right)+2\lambda_\text{min}\left(\E\left[b^2\x\x^\top\middle\vert y=-1\right]\right)}{2}\|\w_p-\w\|^2\\
	&\geq\left\langle \nabla f(\w), \w_p-\w\right\rangle+\frac{2\min(c_1^2,c_2^2)\left[\lambda_\text{min}\left(\E\left[\x\x^\top\middle\vert y=1\right]\right)+\lambda_\text{min}\left(\E\left[\x\x^\top\middle\vert y=-1\right]\right)\right]}{2}\|\w_p-\w\|^2\\
	&\geq \min_{\w'}\left[\left\langle \nabla f(\w), \w'-\w\right\rangle+\frac{2\min(c_1^2,c_2^2)\left[\lambda_\text{min}\left(\E\left[\x\x^\top\middle\vert y=1\right]\right)+\lambda_\text{min}\left(\E\left[\x\x^\top\middle\vert y=-1\right]\right)\right]}{2}\|\w'-\w\|^2\right]\\
	&=-\frac{1}{4\min(c_1^2,c_2^2)\left[\lambda_\text{min}\left(\E\left[\x\x^\top\middle\vert y=1\right]\right)+\lambda_\text{min}\left(\E\left[\x\x^\top\middle\vert y=-1\right]\right)\right]}\|\nabla f(\w)\|^2,
	\end{align*}
	where $(*)$ holds since $\lambda_{\text{min}}(A+B)\geq \lambda_{\text{min}}(A)+\lambda_{\text{min}}(B)$, and the last inequality holds since $a^2\geq \min(c_1^2,c_2^2)$ and $b^2\geq \min(c_1^2,c_2^2)$.
	%where $\lambda=a^2\sigma_\text{min}\left[\x\x^\top\middle\vert y=1\right]+b^2\sigma_\text{min}\left[\x\x^\top\middle\vert y=-1\right]$
\end{proof}

\subsection{Dataset Preparation}
\label{dataset:prep}
We construct the datasets in the following ways: For CIFAR10/STL10, we label the first 5 classes as negative ("-") class and the last 5 classes as positive ("+") class, which leads to a 50/50 class ratio. For CIFAR100, we label the first 50 classes as negative ("-") class and the last 50 classes as positve ("+") class. For the imbalanced cases, we randomly remove 90\%, 80\%, 60\% data from negative samples on all training data, which lead to 91/9, 83/17, 71/29 ratio respectively. For testing data, we keep them unchanged.

\subsection{More Experiments}
\begin{figure}[ht]
	\centering
	%\hspace{-10pt}
	\includegraphics[scale=0.2]{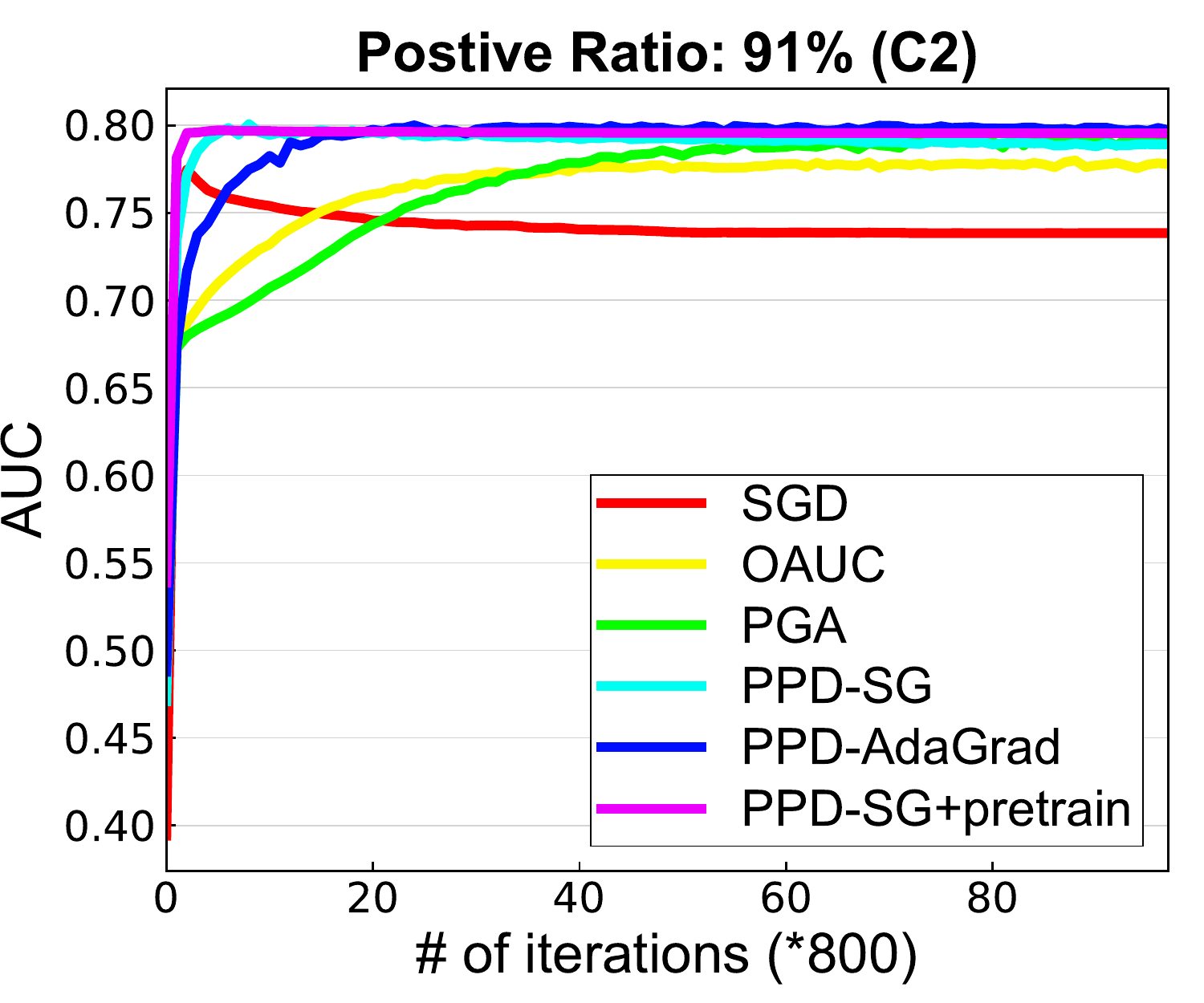}
	\includegraphics[scale=0.2]{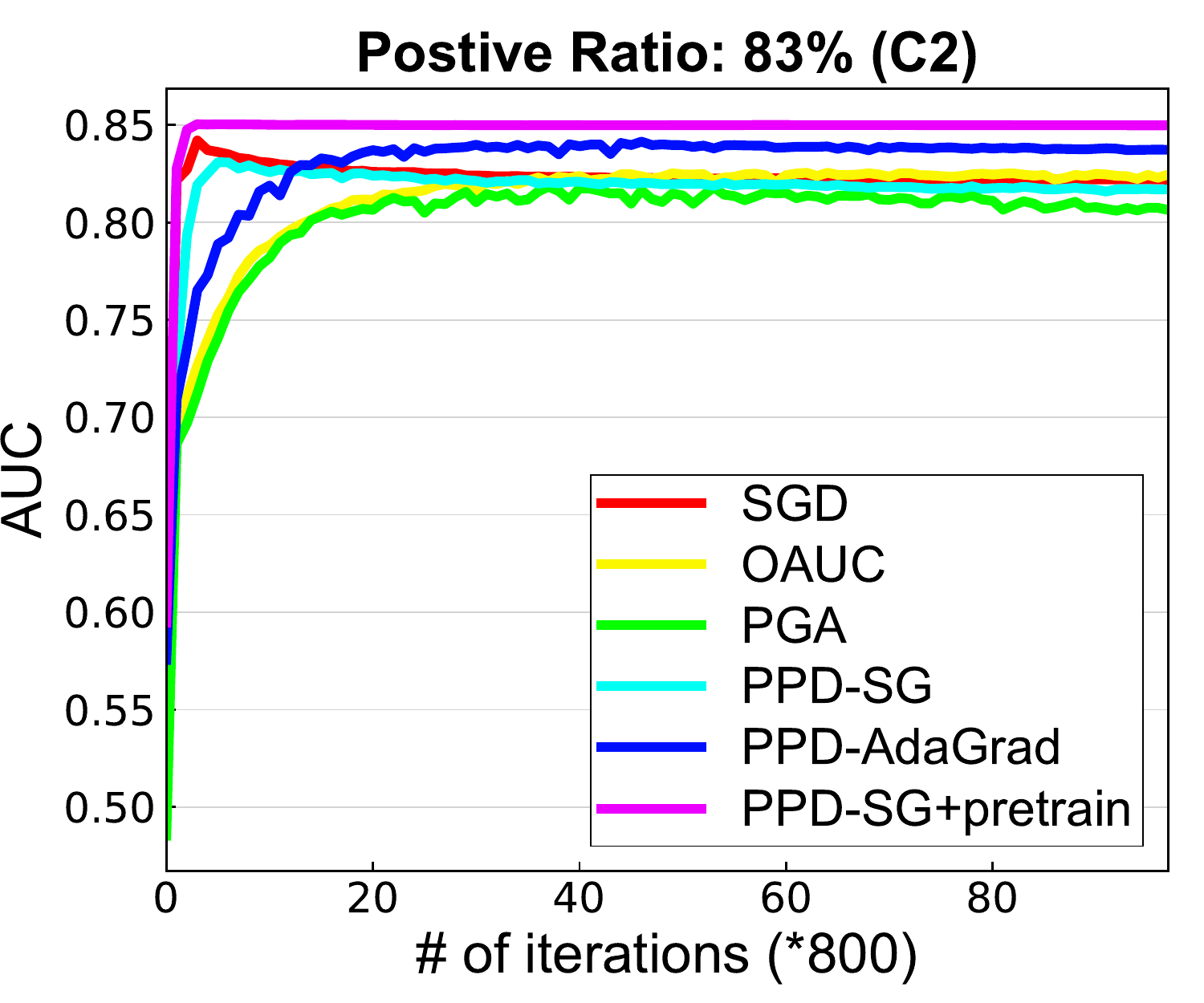}
	\includegraphics[scale=0.2]{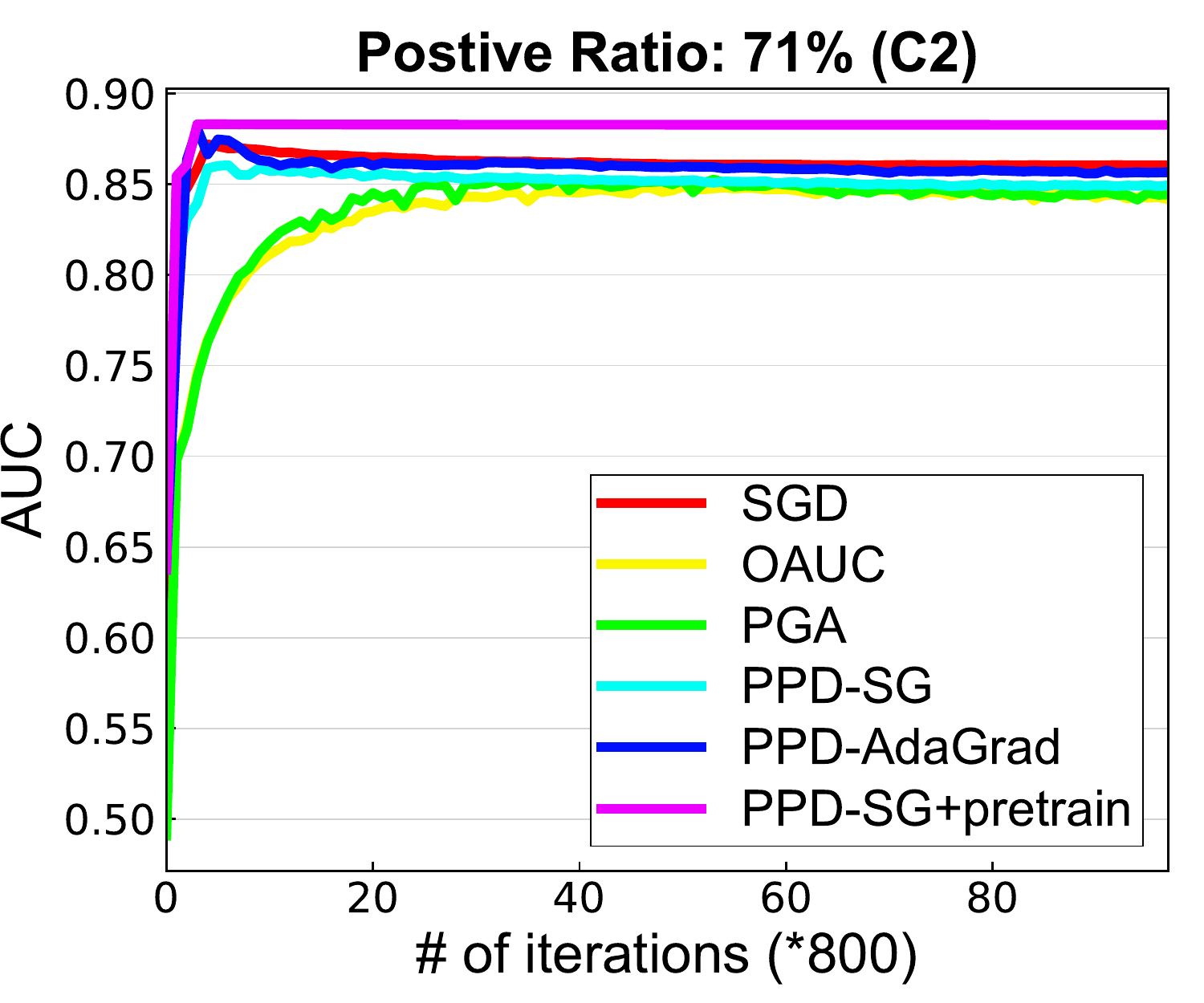}
	\includegraphics[scale=0.2]{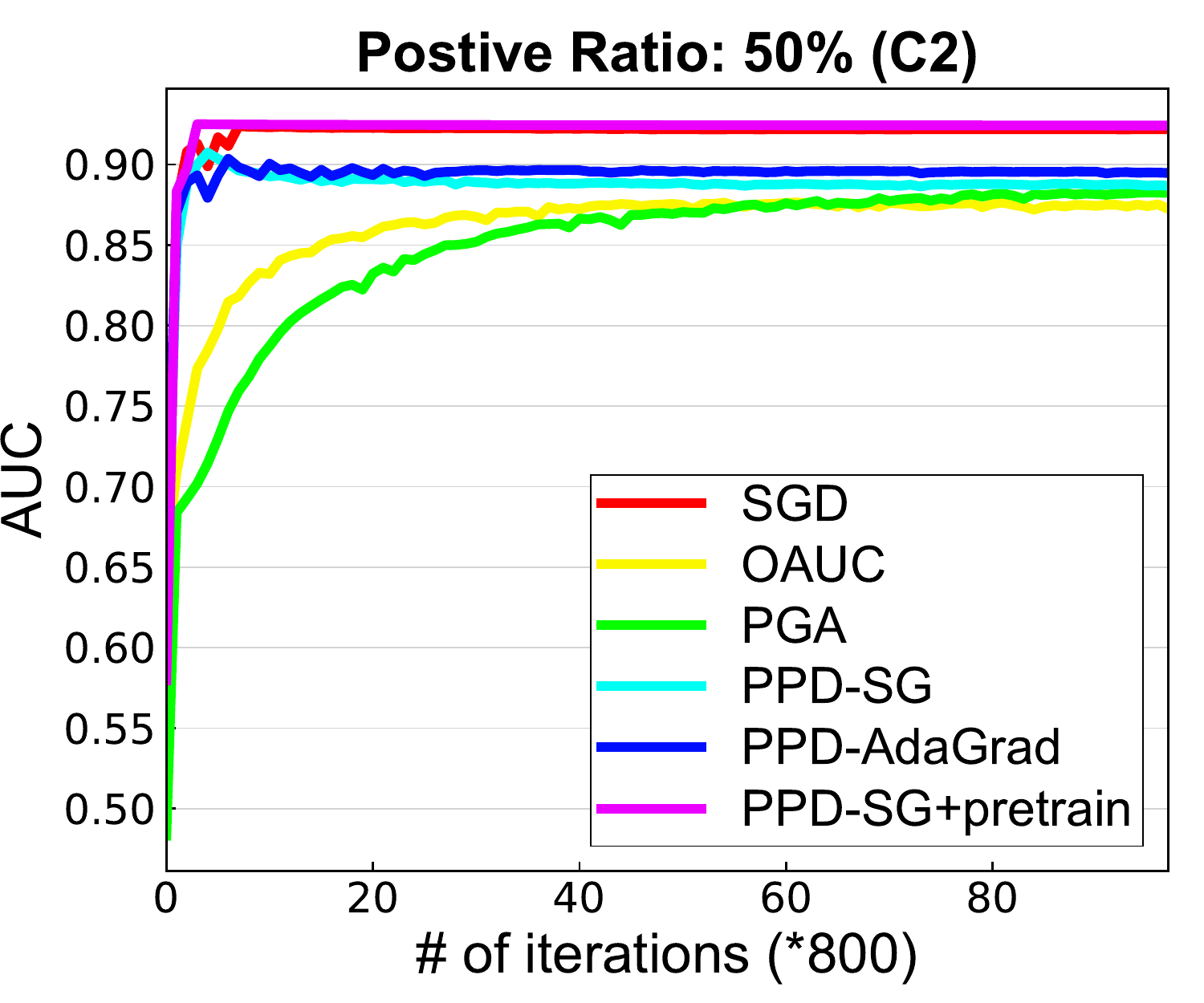}
	
	%\hspace{-10pt}
	\includegraphics[scale=0.2]{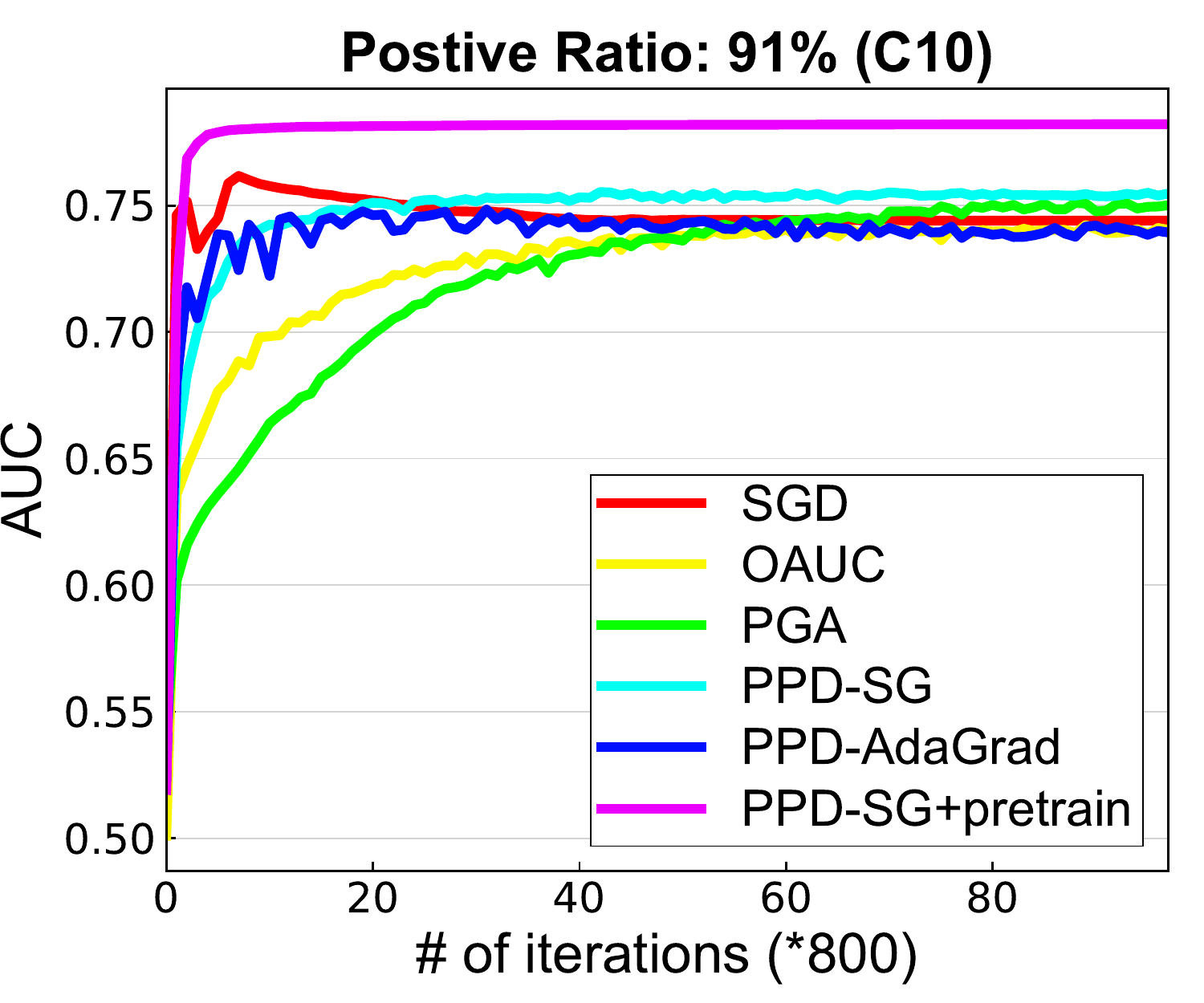}
	\includegraphics[scale=0.2]{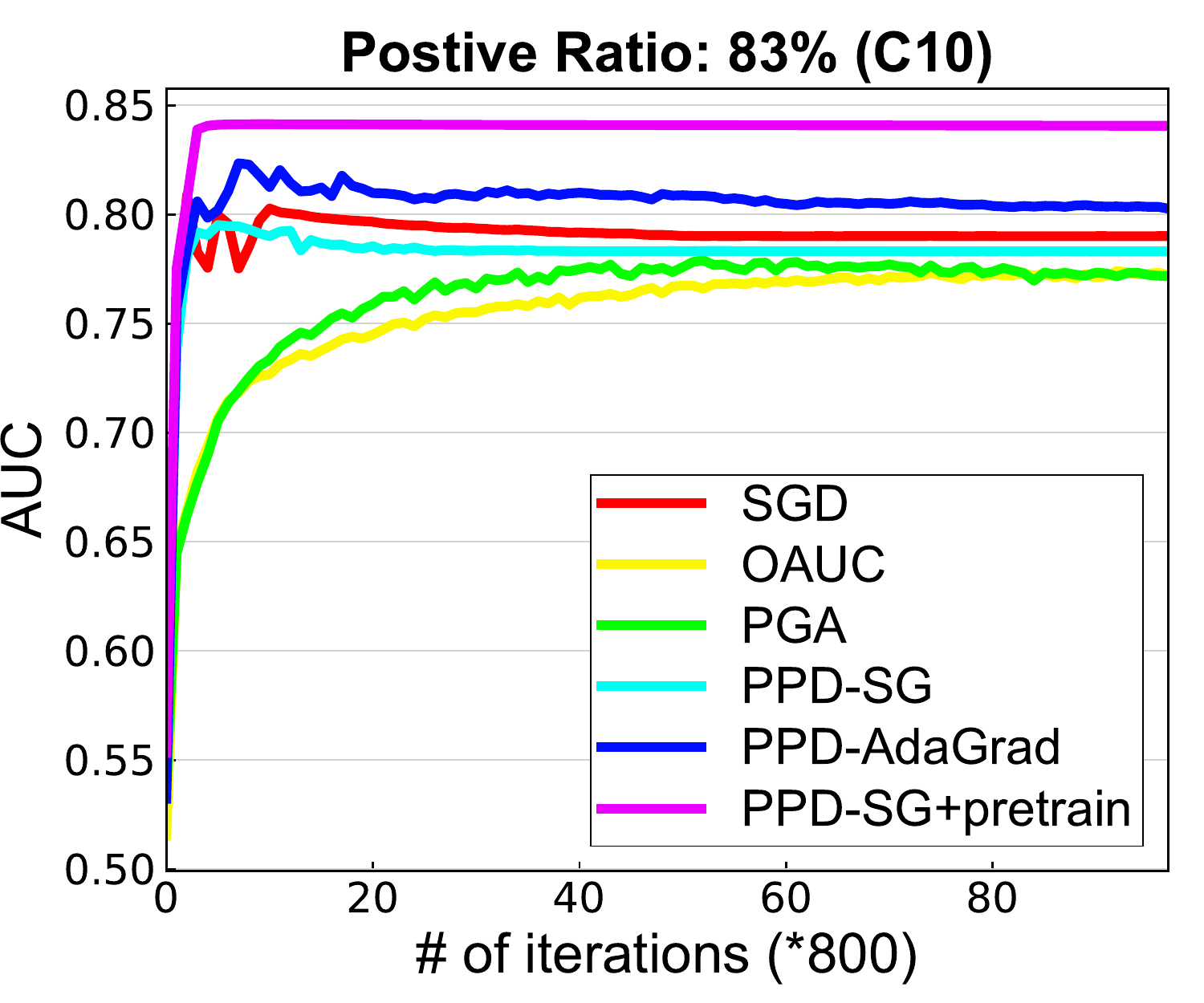}
	\includegraphics[scale=0.2]{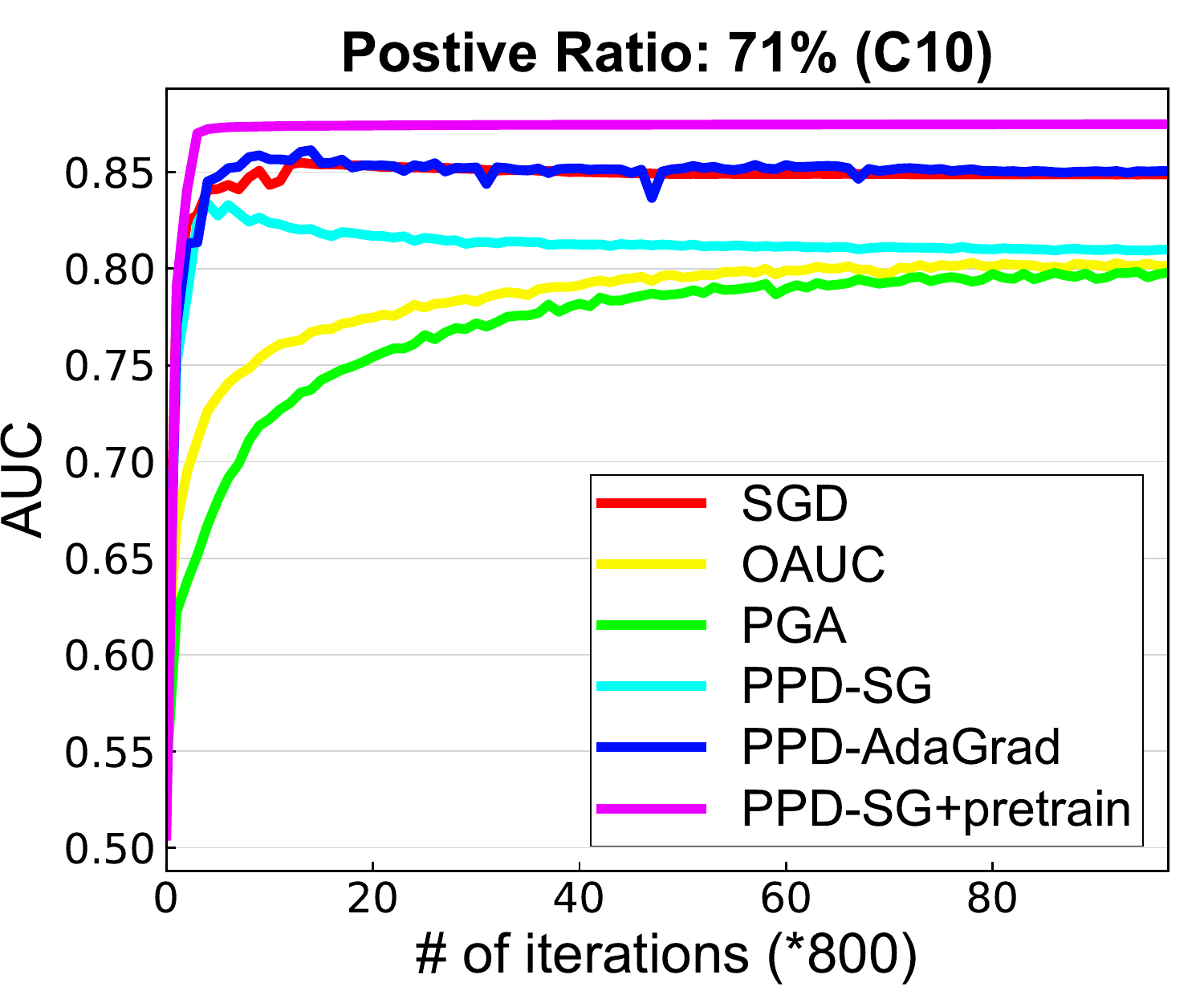}
	\includegraphics[scale=0.2]{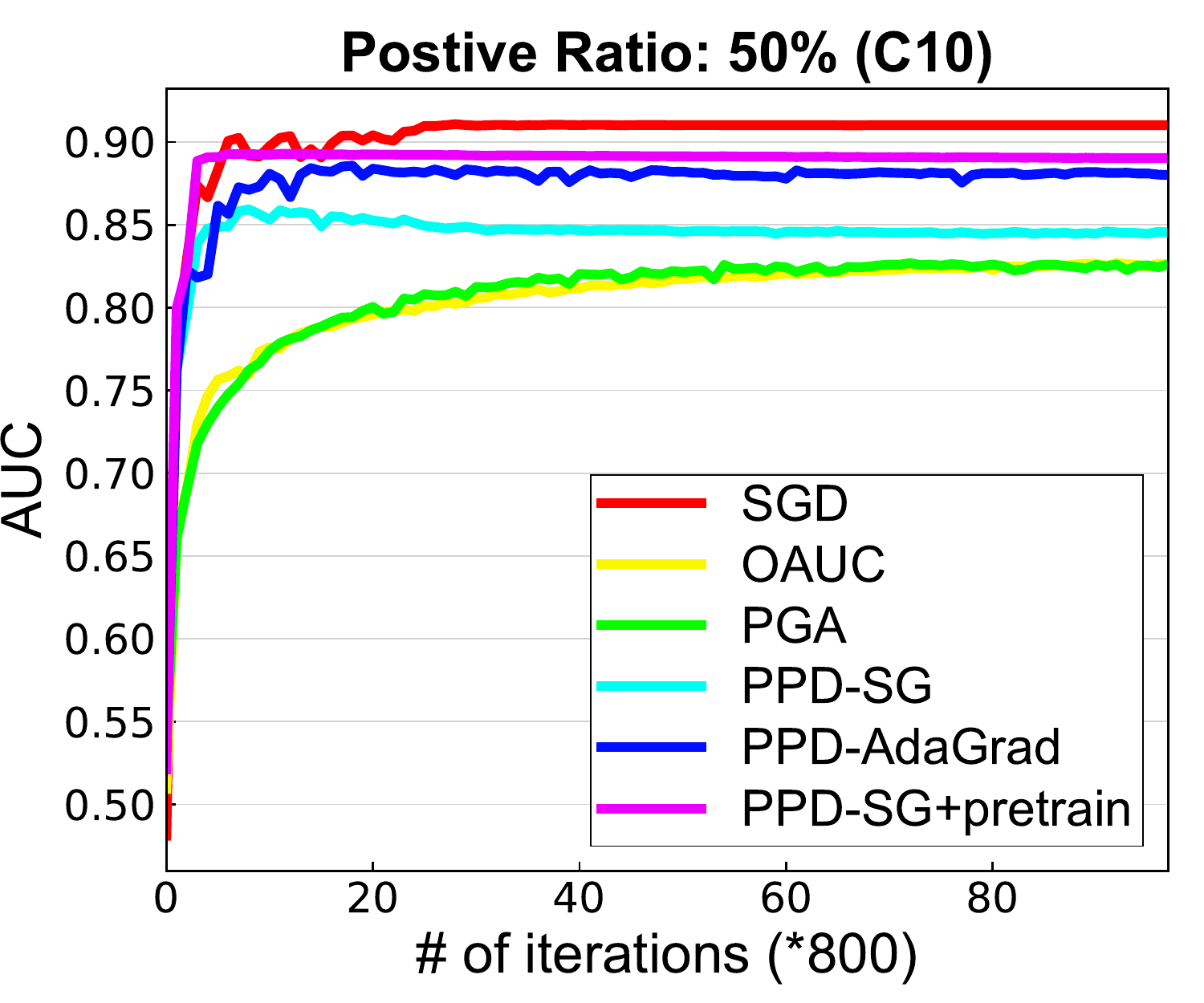}
	
	%\hspace{-10pt}
	\includegraphics[scale=0.2]{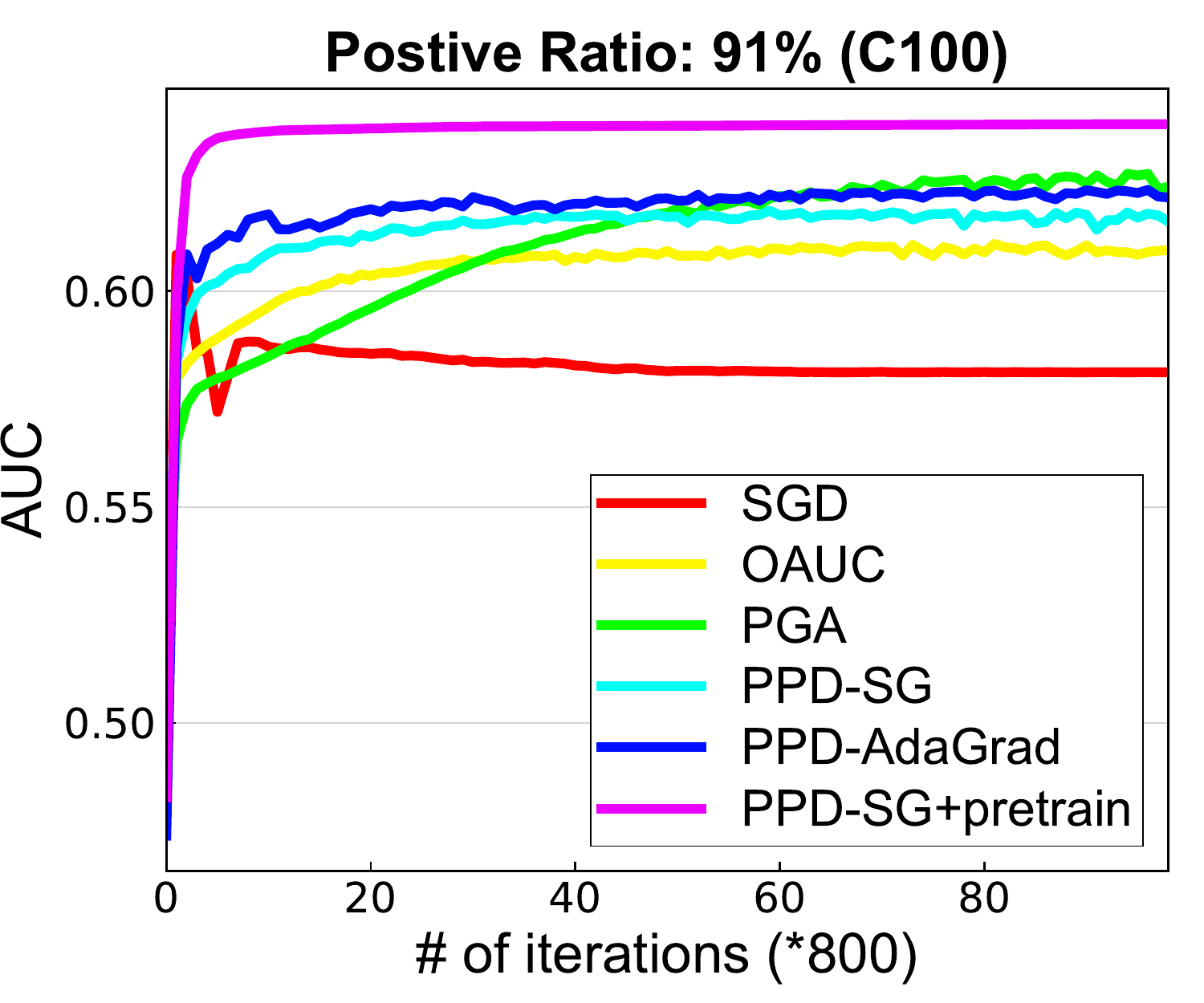}
	\includegraphics[scale=0.2]{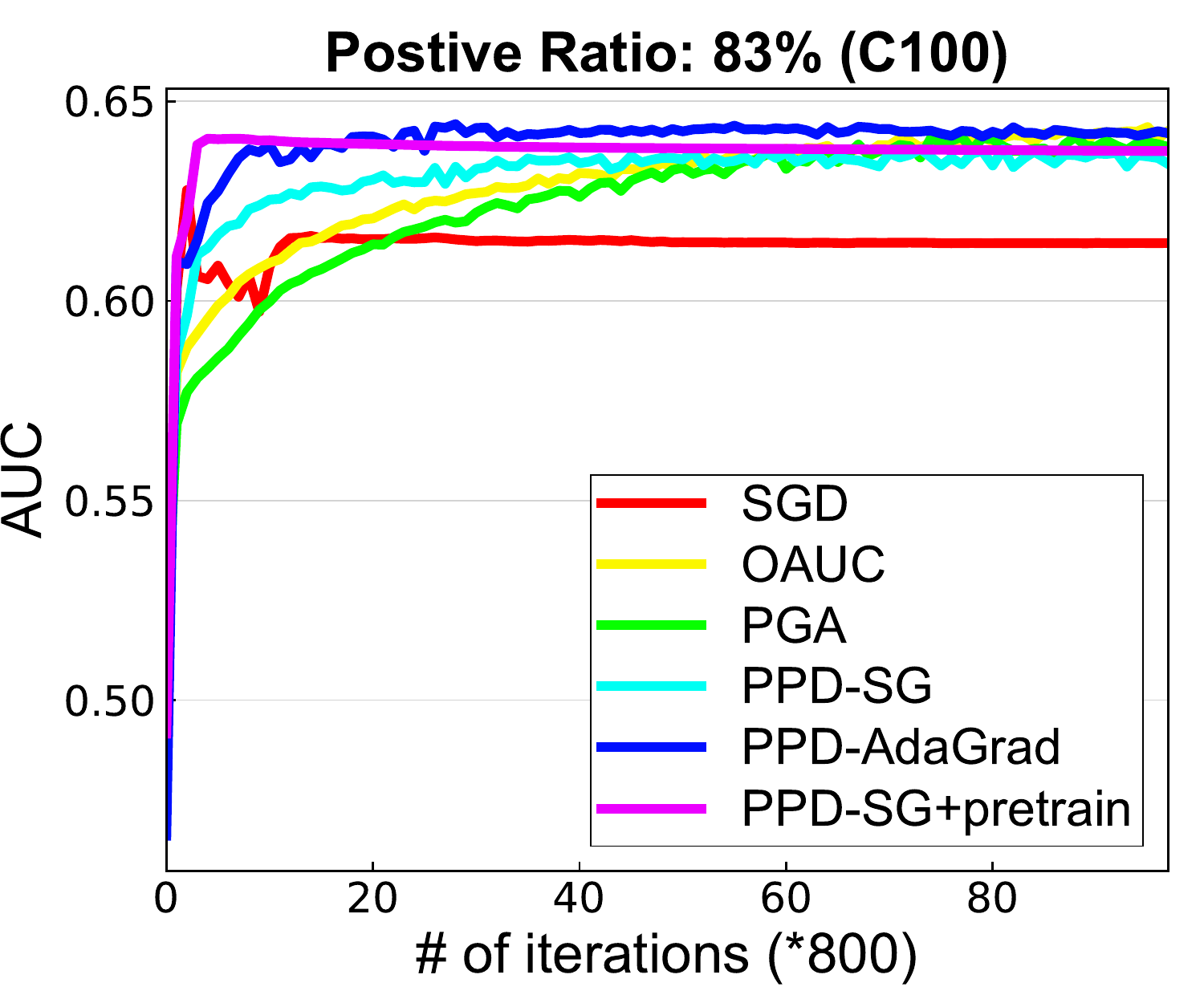}
	\includegraphics[scale=0.2]{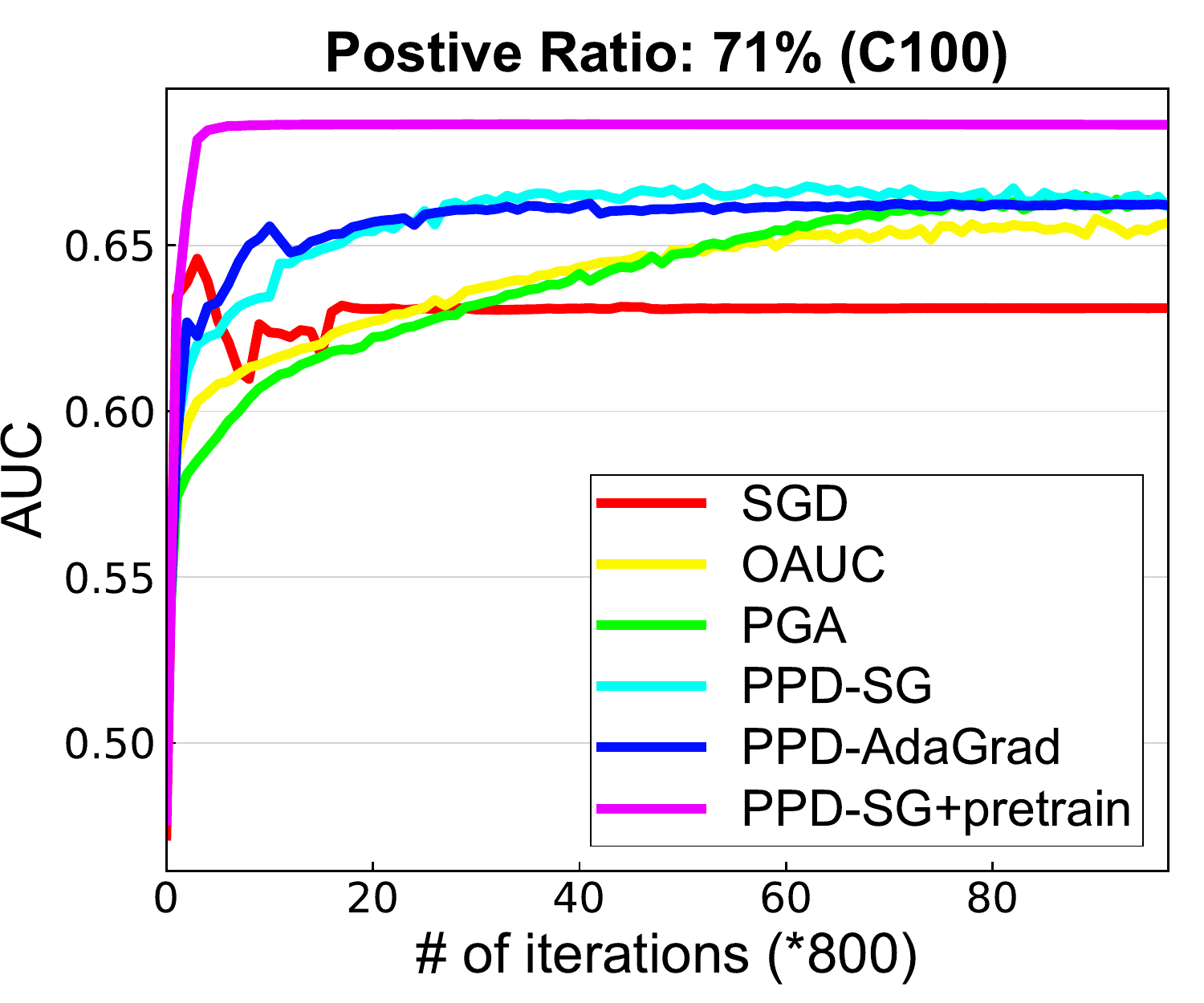}
	\includegraphics[scale=0.2]{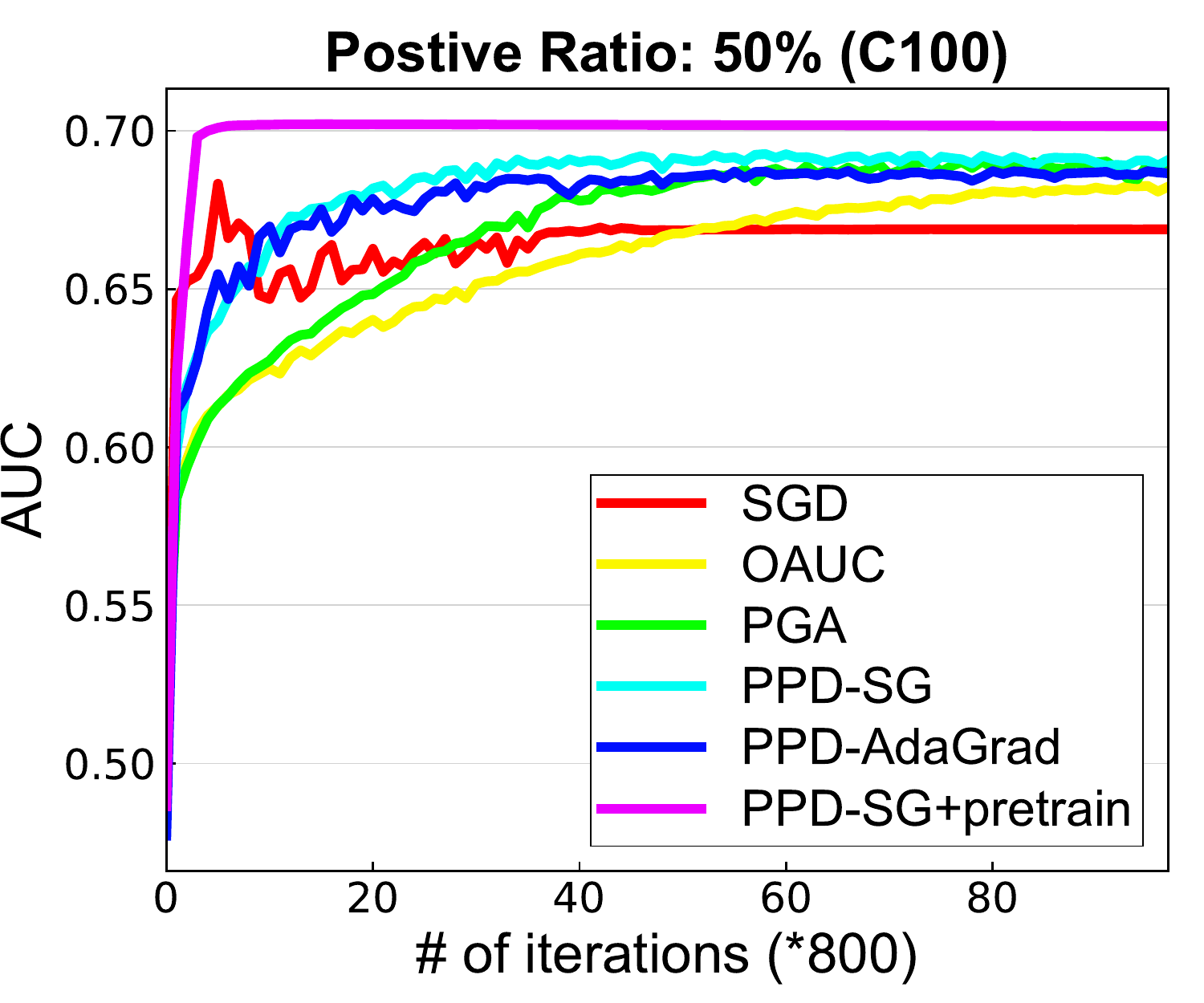}
	
	%\hspace{-10pt}
	\includegraphics[scale=0.2]{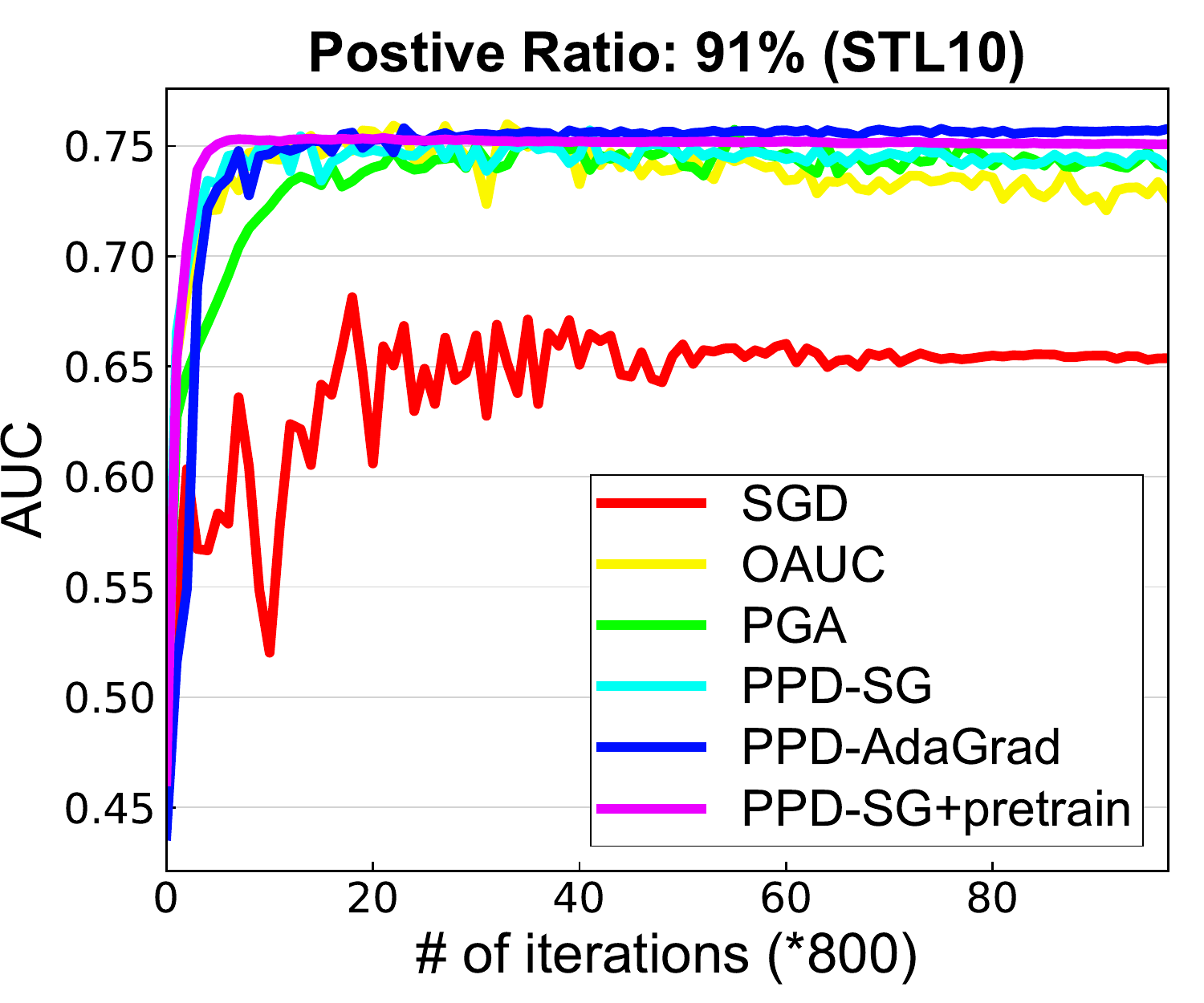}
	\includegraphics[scale=0.2]{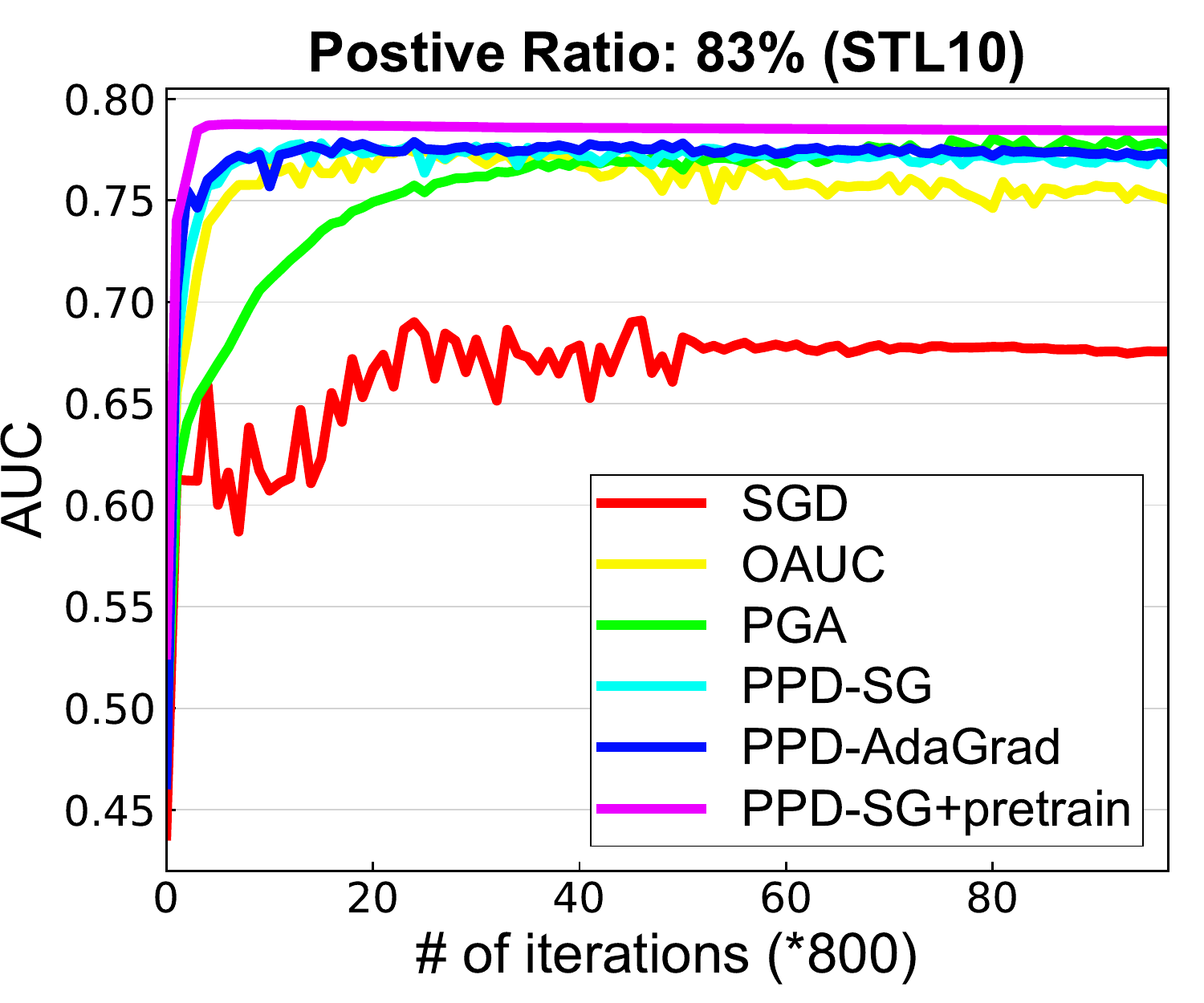}
	\includegraphics[scale=0.2]{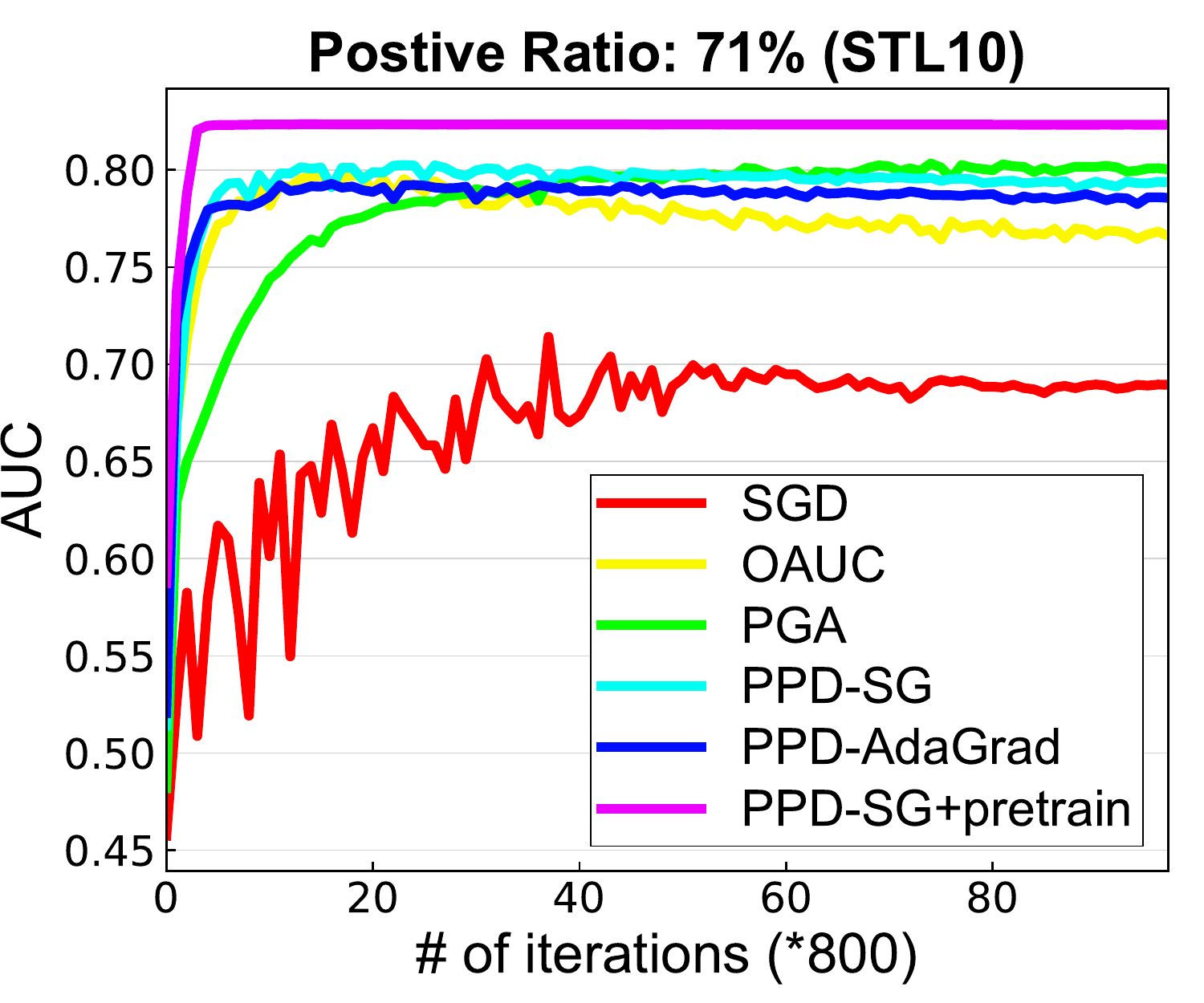}
	\includegraphics[scale=0.2]{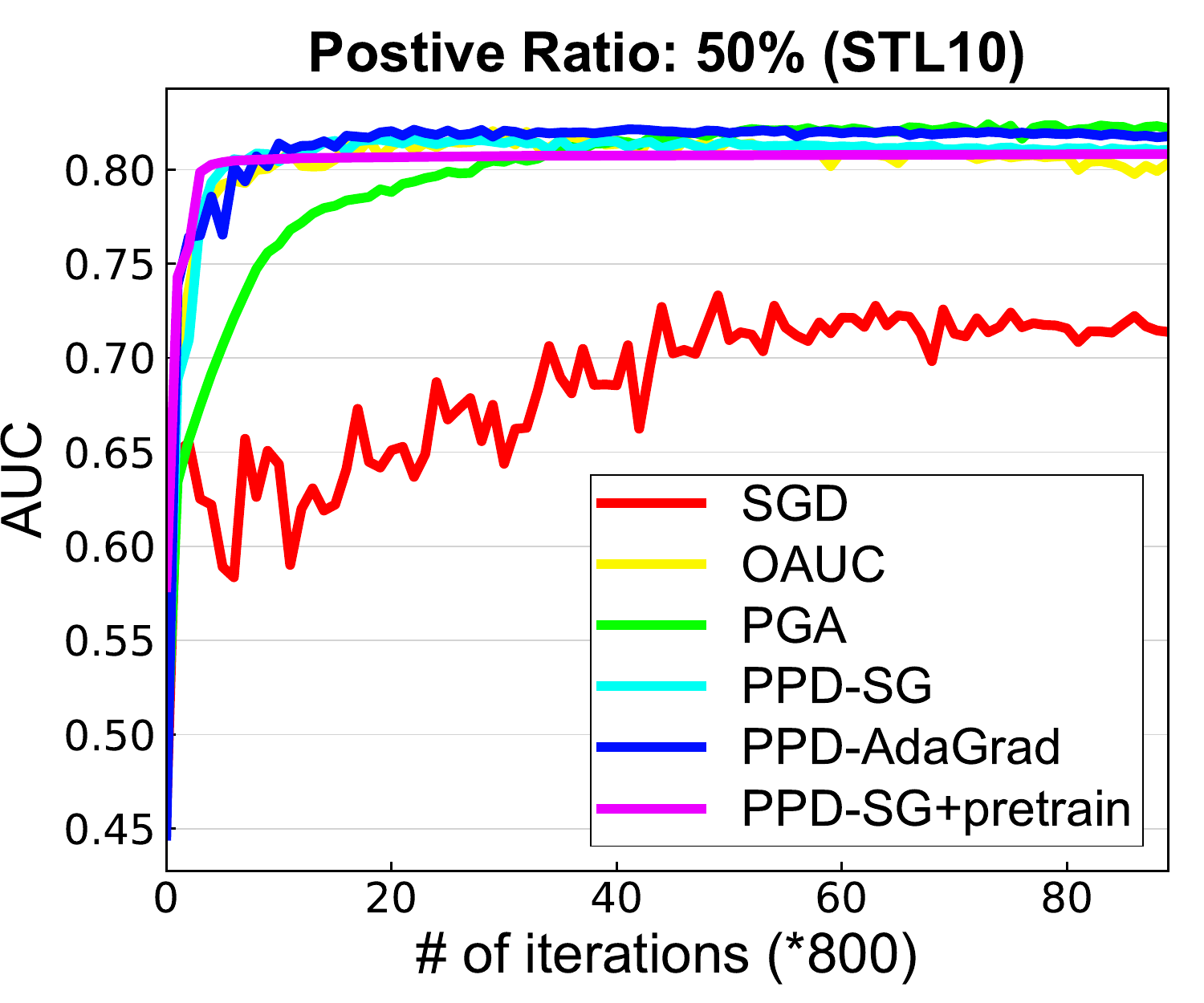}
	\vspace*{-0.08in}
	\caption{Comparison of testing AUC on  Cat\&Dog, CIFAR10, CIFAR100 and STL10.}
	\label{fig:pre_train}
	\vspace*{0.02in}
\end{figure}

Model pretraining is effective in many deep learning tasks, and thus we further evaluate the performance of the proposed methods on pretrained models. We first train the model using SGD up to 2000 iterations with an initial step size of 0.1, and then continue training using PPD-SG. We denote this method as \textit{PPD-SG+pretrain} and the results are shown in Figure~\ref{fig:pre_train}. The parameters are tuned in the same range as in Section~\ref{exp:settings}. It is observed that pretraining model helps the convergence of model and it can achieve the better performance in terms of AUC in most cases.

% \begin{figure}[ht]
% 	\centering
% 	%\hspace{-10pt}
% 	\includegraphics[scale=0.2]{{rebuttal/[-1-0.05]train_test_C10}.pdf}
% 	\includegraphics[scale=0.2]{{rebuttal/[-1-0.05]train_test_C100}.pdf}
% 	\includegraphics[scale=0.2]{{rebuttal/[-1-0.05]train_test_time_C10}.pdf}
% 	\includegraphics[scale=0.2]{{rebuttal/[-1-0.05]train_test_time_C100}.pdf}
% 	\includegraphics[scale=0.2]{{rebuttal/[-1-0.1]train_test_C2}.pdf}
% 	\includegraphics[scale=0.2]{{rebuttal/[-1-0.1]train_test_C10}.pdf}
% 	\includegraphics[scale=0.2]{{rebuttal/[-1-0.1]train_test_C100}.pdf}
% 	\includegraphics[scale=0.2]{{rebuttal/[-1-0.1]train_test_STL10}.pdf}
% 	%\includegraphics[scale=0.2]{{v2_0-0.4_train_test_C2-eps-converted-to}.pdf}
% 	%\includegraphics[scale=0.2]{{v2_0-1.0_train_test_C2-eps-converted-to}.pdf}

% 	\vspace*{-0.08in}
% 	\caption{Comparison of testing AUC on  Cat\&Dog, CIFAR10, CIFAR100 and STL10 with random labeled class.}
% 	\label{fig:add_exp}
% 	\vspace*{0.02in}
% \end{figure}
\vspace*{-0.1in}
\subsection{Additional Experiments with Different Labeling Order}
\begin{figure}[ht]
	\centering
	%\hspace{-10pt}
	\includegraphics[scale=0.2]{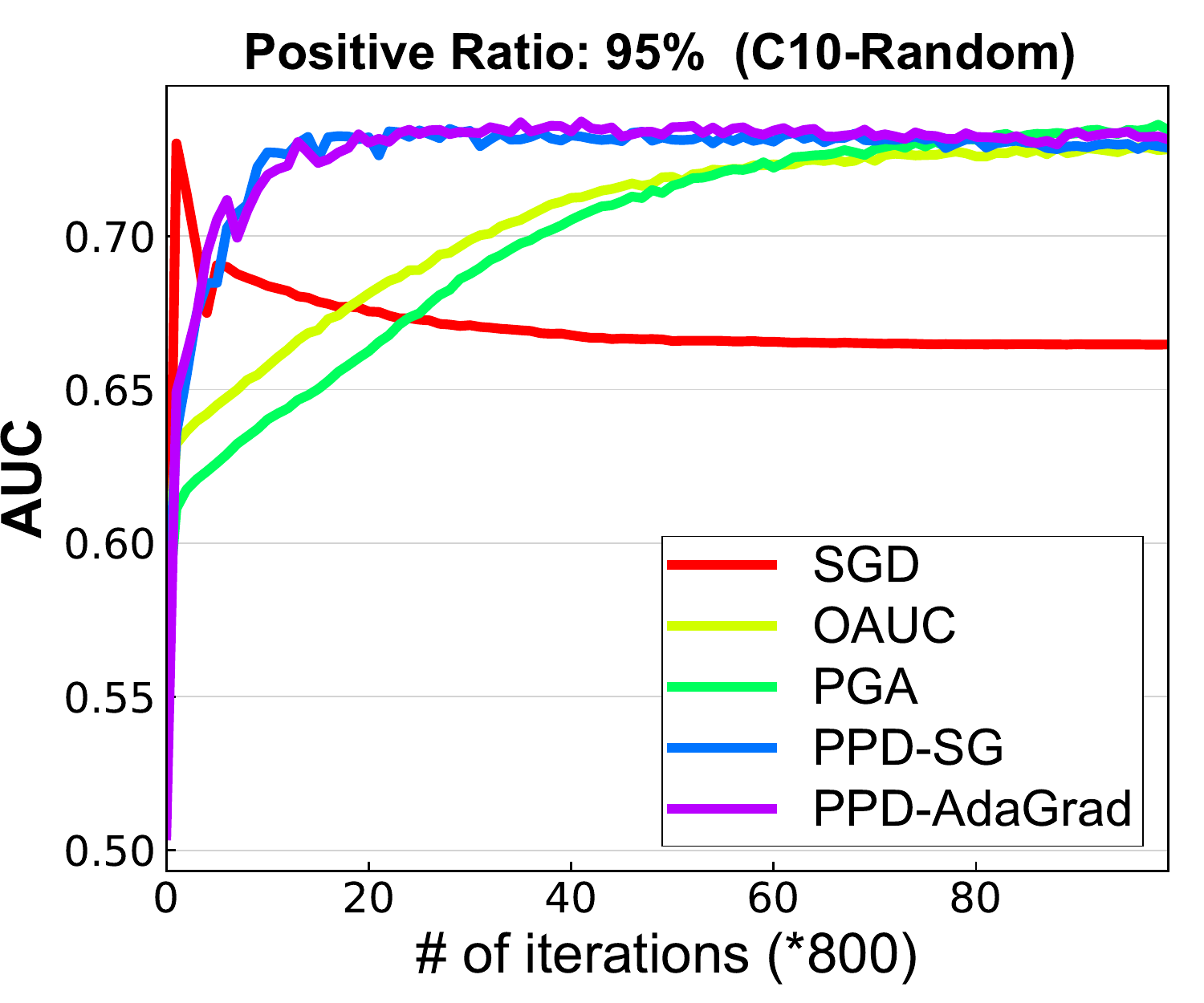}
	\includegraphics[scale=0.2]{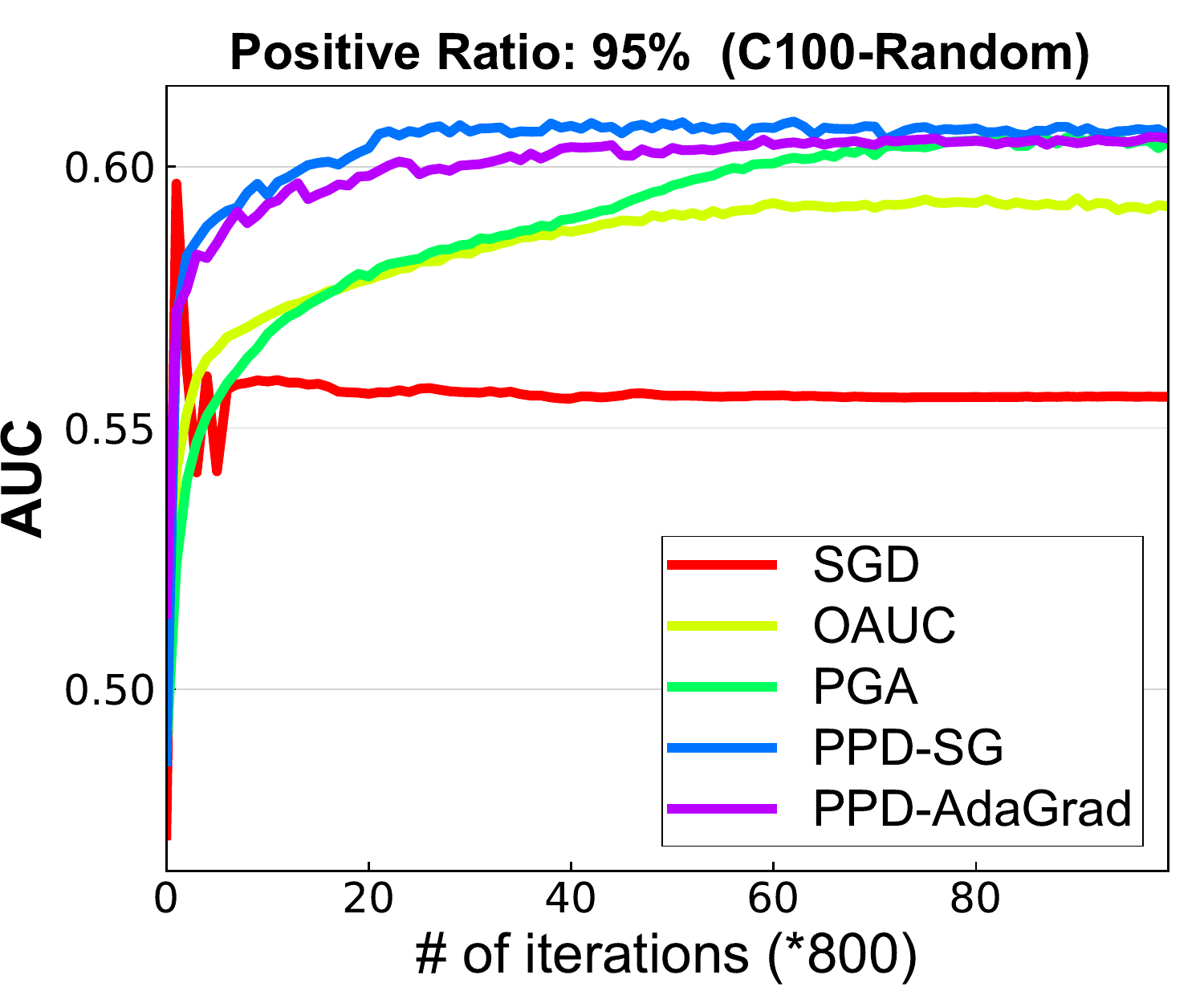}
	\includegraphics[scale=0.2]{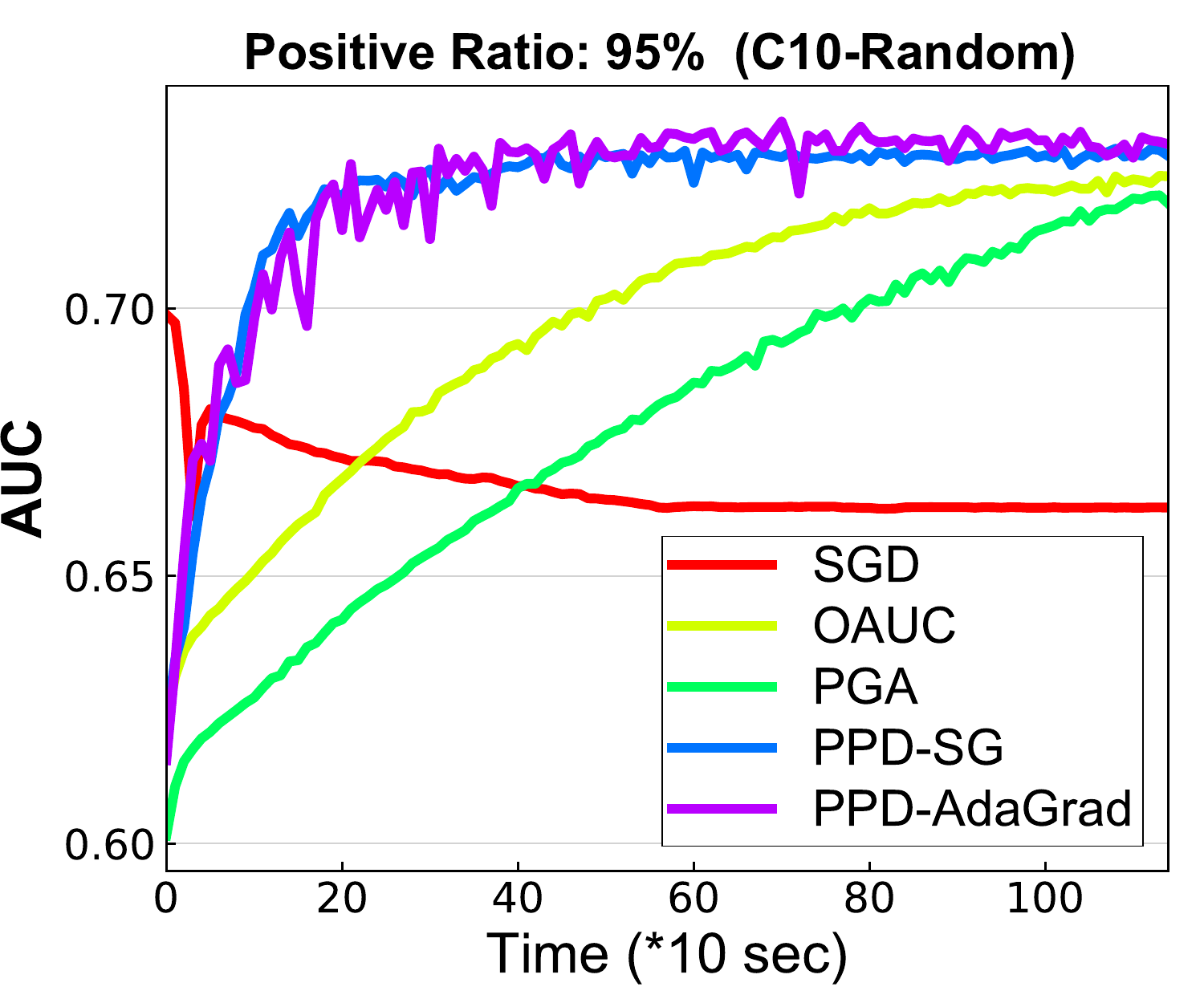}
	\includegraphics[scale=0.2]{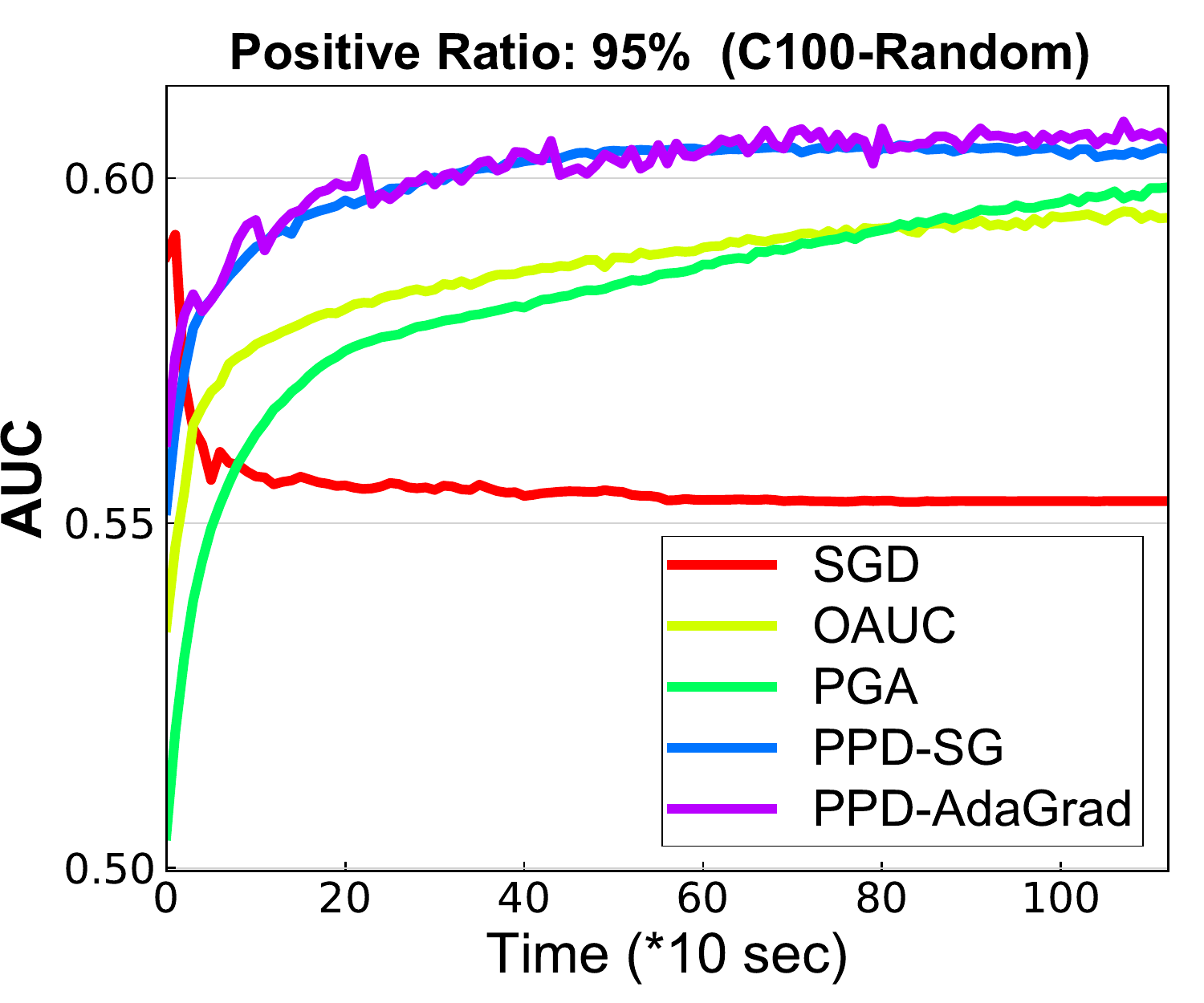}
	\includegraphics[scale=0.2]{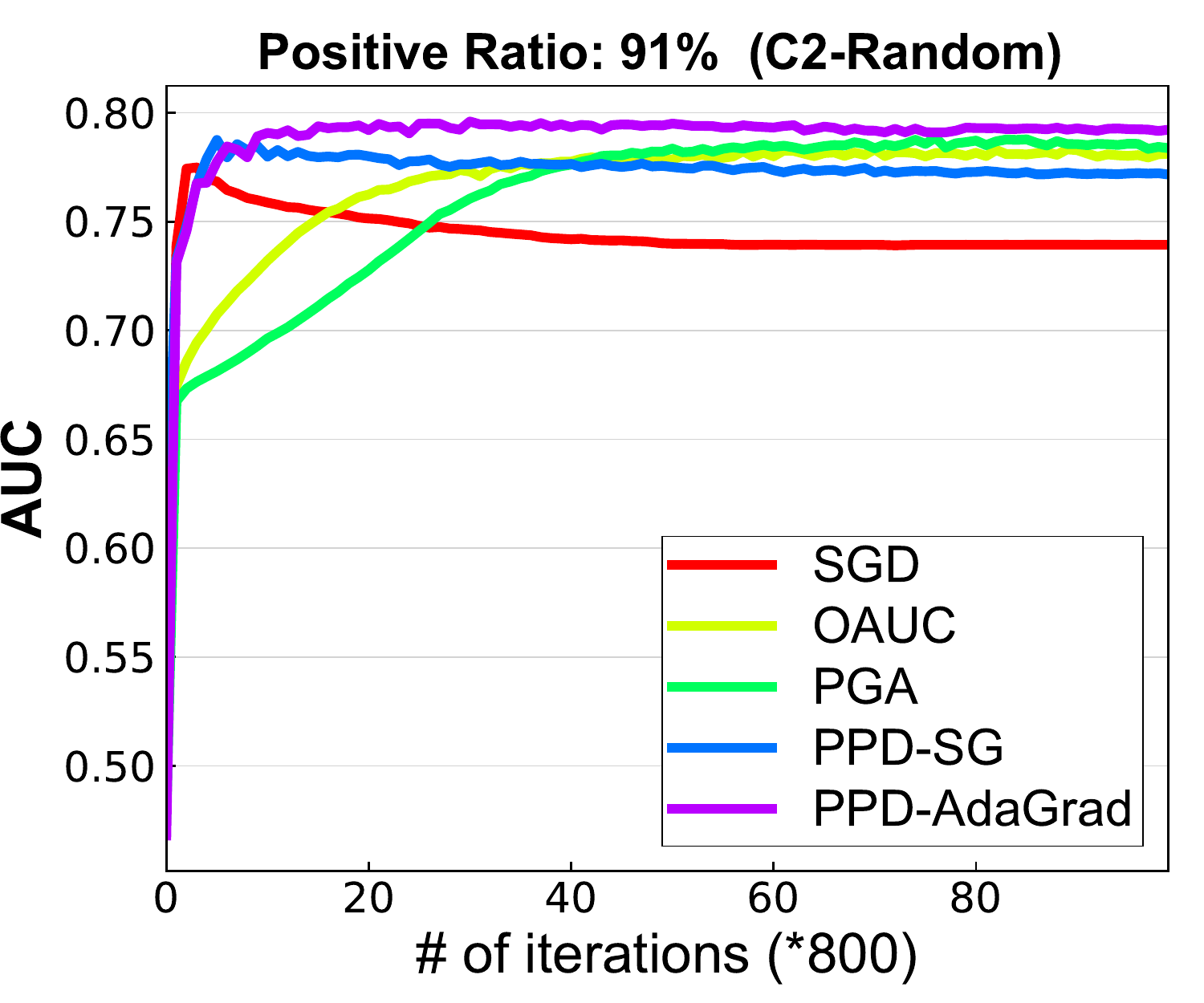}
	\includegraphics[scale=0.2]{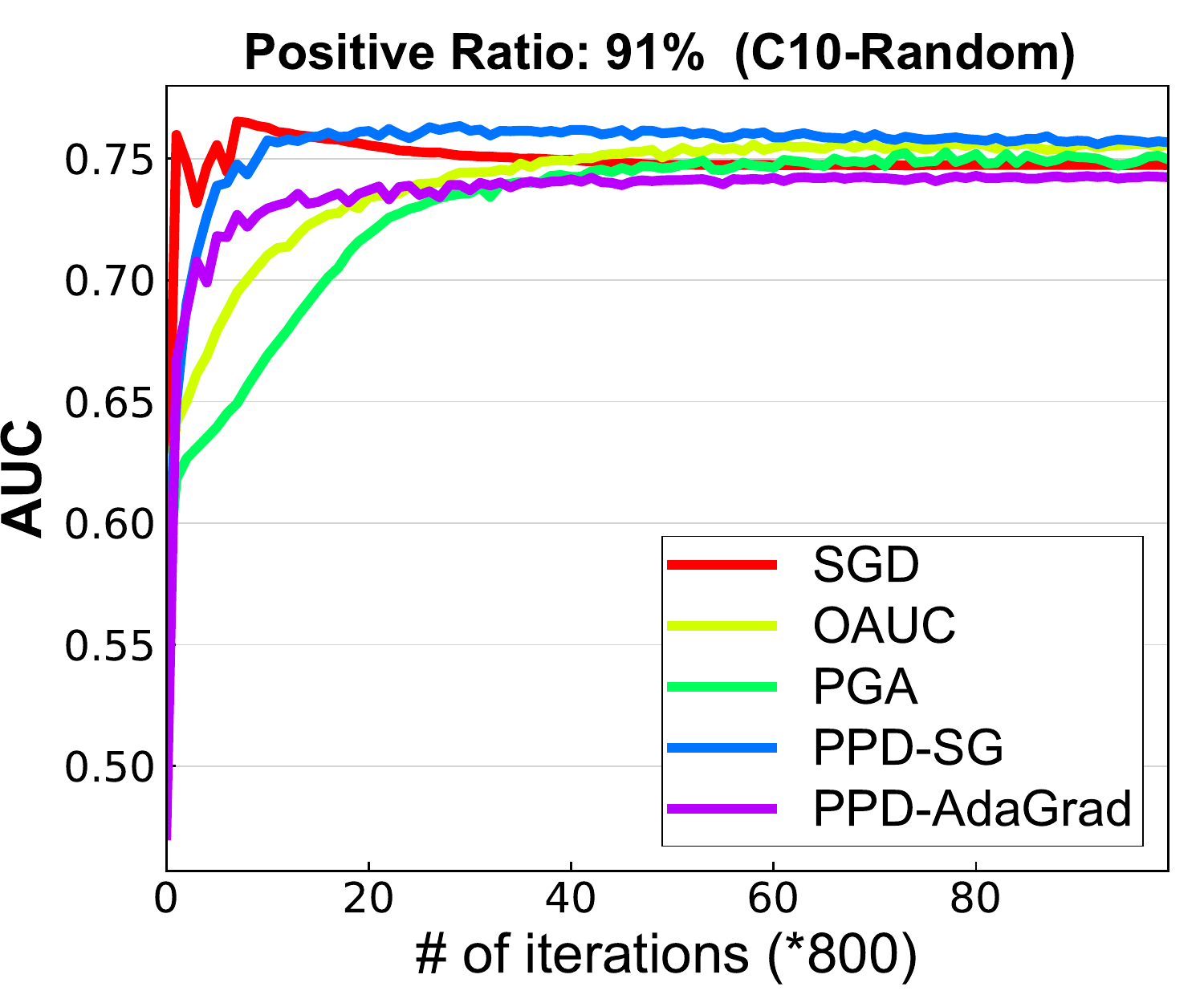}
	\includegraphics[scale=0.2]{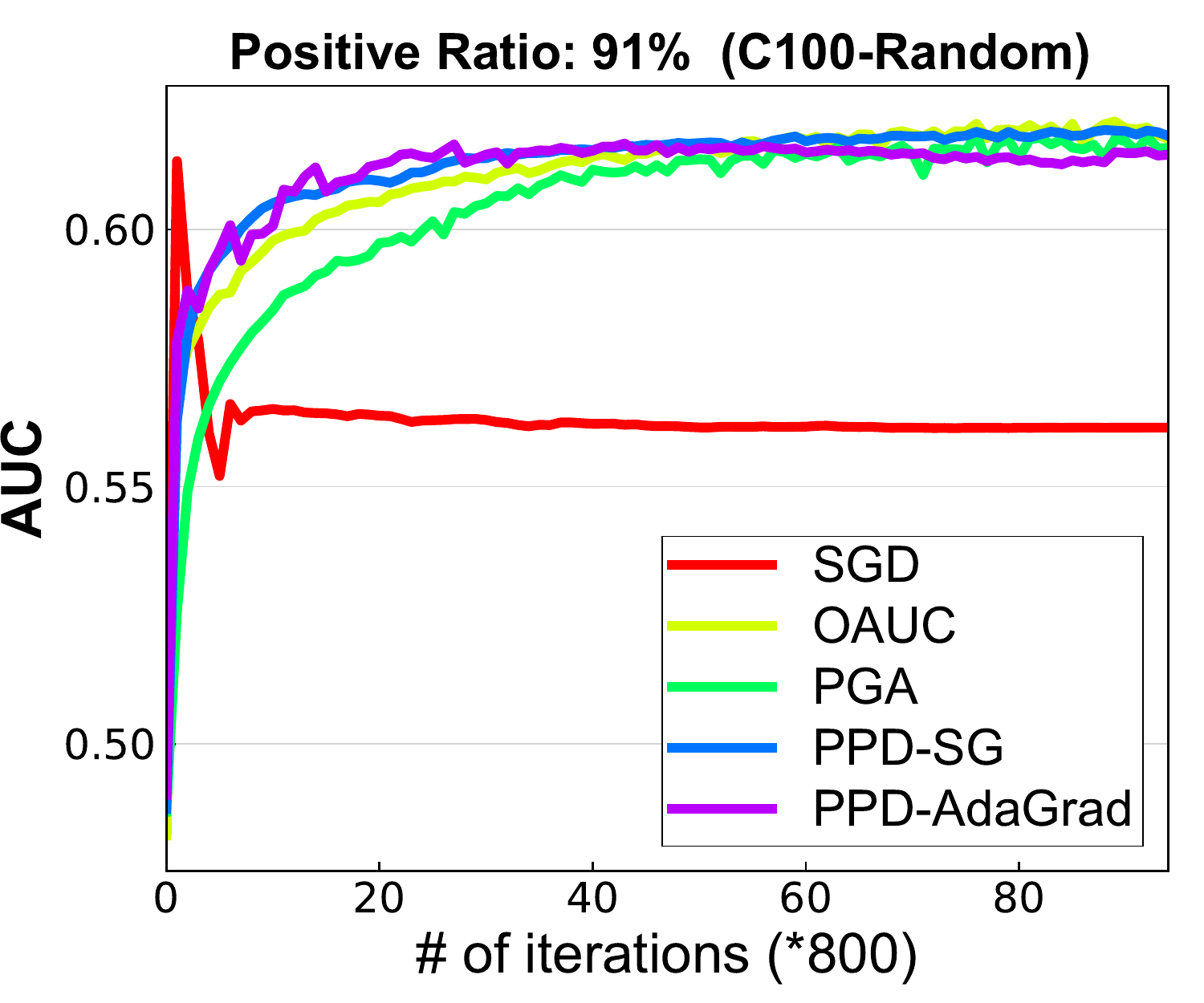}
	\includegraphics[scale=0.2]{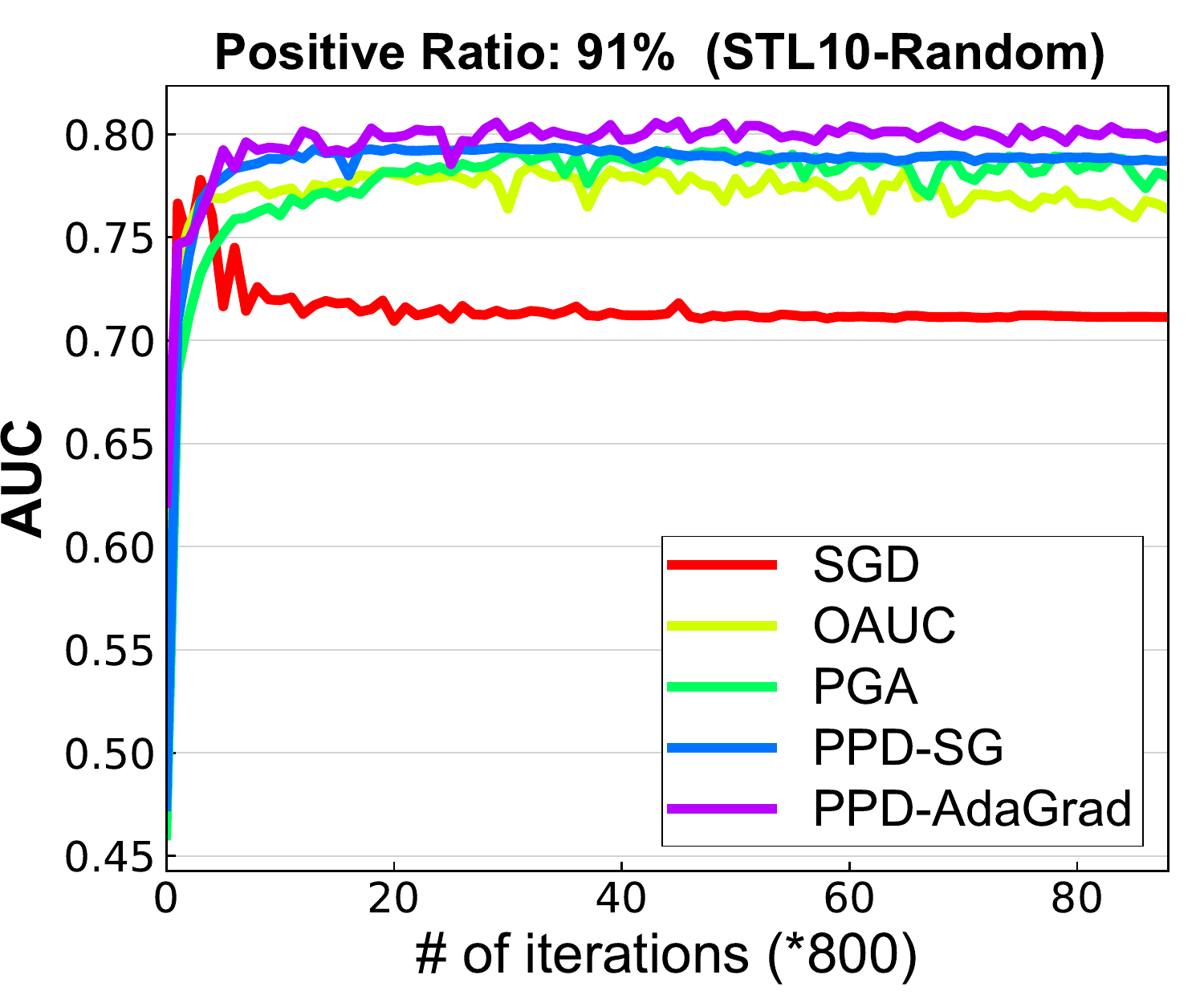}

	\vspace*{-0.08in}
	\caption{Comparison of testing AUC on  Cat\&Dog, CIFAR10, CIFAR100 and STL10. For CIFAR10 and STL10 dataset, we randomly partition the 10 classes into two labels (i.e., randomly select 5 classes as positive label and other 5 classes as negative label). For CIFAR100 dataset, we randomly partition the 100 classes into two labels (i.e., randomly select 50 classes as positive label and other 50 classes as negative label).}
	\label{fig:add_exp}
	\vspace*{0.02in}
\end{figure}

To investigate the effects of labeling order, we also attempt to randomly partition the classes as positive or negative equally. For CIFAR10 and STL10 dataset, we randomly partition the 10 classes into two labels (i.e., randomly select 5 classes as positive label and other 5 classes as negative label). For CIFAR100 dataset, we randomly partition the 100 classes into two labels (i.e., randomly select 50 classes as positive label and other 50 classes as negative label). After that we randomly remove 95\%, 90\%, from negative samples on all training data, which lead to  20:1, 10:1 ratios respectively. For testing data, we keep them unchanged. We also add AdaGrad for minimizing cross-entropy loss as a new baseline. The corresponding experimental results are included in Figure~\ref{fig:add_exp}. We can see that PPD-Adagrad and PPD-SG converge faster than other baselines. 
%\begin{figure}[ht]
%	\centering
%	%\hspace{-10pt}
%	\includegraphics[scale=0.2]{{rebuttal_adagrad/[-1-0.05]train_test_C10}.pdf}
%	\includegraphics[scale=0.2]{{rebuttal_adagrad/[-1-0.05]train_test_C100}.pdf}
%	\includegraphics[scale=0.2]{{rebuttal_adagrad/[-1-0.05]train_test_time_C10}.pdf}
%	\includegraphics[scale=0.2]{{rebuttal_adagrad/[-1-0.05]train_test_time_C100}.pdf}
%	\includegraphics[scale=0.2]{{rebuttal_adagrad/[-1-0.1]train_test_C2}.pdf}
%	\includegraphics[scale=0.2]{{rebuttal_adagrad/[-1-0.1]train_test_C10}.pdf}
%	\includegraphics[scale=0.2]{{rebuttal_adagrad/[-1-0.1]train_test_C100}.pdf}
%	\includegraphics[scale=0.2]{{rebuttal_adagrad/[-1-0.1]train_test_STL10}.pdf}
%	%\includegraphics[scale=0.2]{{v2_0-0.4_train_test_C2-eps-converted-to}.pdf}
%	%\includegraphics[scale=0.2]{{v2_0-1.0_train_test_C2-eps-converted-to}.pdf}
%
%	\vspace*{-0.08in}
%	\caption{Comparison of testing AUC on  Cat\&Dog, CIFAR10, CIFAR100 and STL10. For CIFAR10 and STL10 dataset, we randomly partition the 10 classes into two labels (i.e., randomly select 5 classes as positive label and other 5 classes as negative label). For CIFAR100 dataset, we randomly partition the 100 classes into two labels (i.e., randomly select 50 classes as positive label and other 50 classes as negative label).}
%	\label{fig:add_exp}
%	\vspace*{0.02in}
%\end{figure}
\end{document}